\documentclass{article}

\usepackage{microtype}
\usepackage{graphicx}
\usepackage{subcaption}
\usepackage{booktabs} 
\usepackage{comment}
\usepackage{adjustbox}
\usepackage{float}

\usepackage{hyperref}

\usepackage[accepted]{icml2023}

\usepackage{amsmath}
\usepackage{amssymb}
\usepackage{mathtools}
\usepackage{amsthm}

\usepackage[capitalize,noabbrev]{cleveref}

\theoremstyle{plain}

\theoremstyle{definition}

\theoremstyle{remark}

\usepackage[textsize=tiny]{todonotes}

\icmltitlerunning{SAP-sLDA: An Interpretable Interface for Exploring Unstructured Text}

\begin{document}

\twocolumn[
\icmltitle{SAP-sLDA: An Interpretable Interface for Exploring Unstructured Text}



\icmlsetsymbol{equal}{*}

\begin{icmlauthorlist}
\icmlauthor{Charumathi Badrinath}{harvard}
\icmlauthor{Weiwei Pan}{harvard}
\icmlauthor{Finale Doshi-Velez}{harvard}
\end{icmlauthorlist}

\icmlaffiliation{harvard}{Harvard University}

\icmlcorrespondingauthor{Charumathi Badrinath}{charumathibadrinath@college.harvard.edu}

\icmlkeywords{Machine Learning, ICML}

\vskip 0.3in
]



\printAffiliationsAndNotice{}  

\begin{abstract}
A common way to explore text corpora is through low-dimensional projections of the documents, where one hopes that thematically similar documents will be clustered together in the projected space. However, popular algorithms for dimensionality reduction of text corpora, like Latent Dirichlet Allocation (LDA), often produce projections that do not capture human notions of document similarity. We propose 
a semi-supervised human-in-the-loop LDA-based method for learning topics that preserve semantically meaningful relationships between documents in low-dimensional projections. On synthetic corpora, our method yields more interpretable projections than baseline methods with only a fraction of labels provided. On a real corpus, we obtain qualitatively similar results.
\end{abstract}

\section{Introduction}
Navigation of unstructured text data is important in many applications. In this paper, we focus on enabling community members of \textit{Dharma Seed}, a non-profit organization for the preservation and sharing of Buddhist teachings, to explore the website's large repository of Buddhist talks. Currently, users must rely on keyword searches to navigate the corpus, limiting their ability to browse talks based on topical interests. Our goal is to improve the user experience by providing an intuitive visualization of the corpus that allows users to browse talks based on thematic similarity.

\begin{figure}
 \centering
 \begin{subfigure}[b]{0.49\linewidth}
     \centering
     \includegraphics[width=\textwidth]{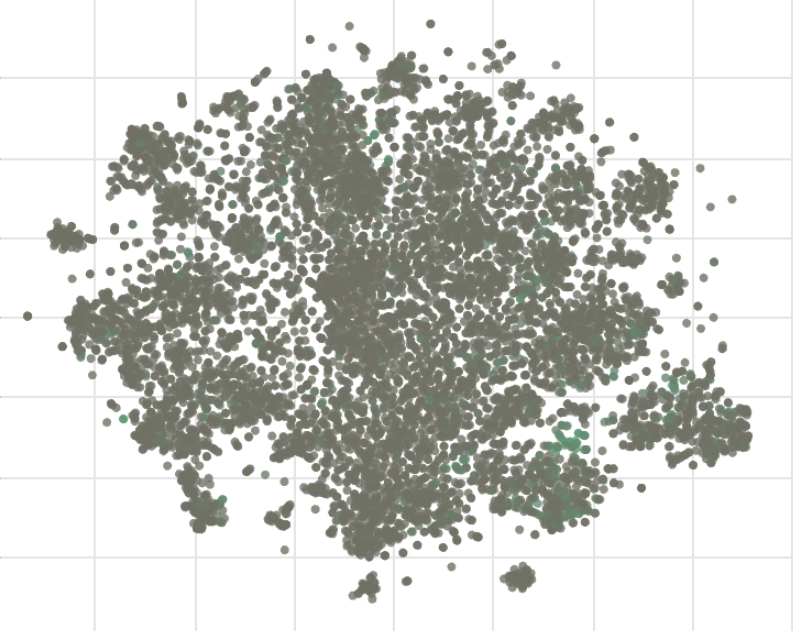}
     \caption{Projection of Dharma Seed corpus colored by expression level of Topic 17.}
 \end{subfigure}
 \hfill
 \begin{subfigure}[b]{0.49\linewidth}
     \centering
     \includegraphics[width=\textwidth]{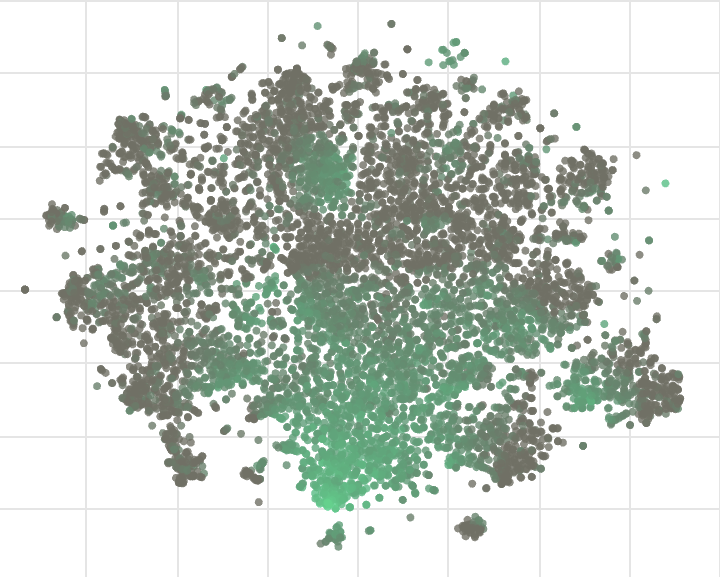}
     \caption{Projection of Dharma Seed corpus colored by expression level of Topic 36.}
 \end{subfigure}
 \caption{2-dimensional projection of documents in the Dharma Seed corpus obtained by applying LDA with 50 topics followed by t-SNE. The figure on the left colors documents in the projection on a scale from grey to green proportional to a document's expression of \textit{Topic 17} and the figure on the right does the same for \textit{Topic 36}. The 5 words with highest probability mass in Topic 17 are \textit{mindfulness, wisdom, develop, quality} and \textit{concentration}. The 5 words with highest probability mass in Topic 36 are \textit{get, see, let, want} and \textit{know}. Distances in the projected space do \textit{not} seem to capture human-relevant notions of similarity.}
\label{fig:Dharma Seed_tsne}
\end{figure}

A common approach for creating such visualizations is through low-dimensional projections of the corpus. Specifically, one can use Latent Dirichlet Allocation (LDA) \cite{lda} to model documents in the corpus as a combination of $K$ number of topics (i.e. a $K$-dimensional vector), then use t-SNE to project each topic combination into a 2-dimensional space \cite{tsne-lda}. However, applying this method to the Dharma Seed corpus results in projections that do not align with human notions of document similarity. For example, in Figure \ref{fig:Dharma Seed_tsne}, we see that while Topic 17 seems interpretable (looking at the top words for Topic 17, we infer that the topic is \textit{practices of mindfulness}), documents primarily associated with Topic 17 show no clustering in the low-dimensional visualization. In contrast, documents predominantly associated the comparatively non-interpretable Topic 36 are clustered close together.

In this work, we formalize our objectives for interpretable low-dimensional representations of a text corpus as follows:
\begin{enumerate}
    \item \textbf{(Semantic Alignment)} The distance between any two documents in the projection should align with human notions of document similarity.
    \item \textbf{(Robustness)} Relative distance between documents in the projection should remain reasonably stable across random restarts and choices of hyper-parameters.
\end{enumerate}

Naive applications of LDA for dimensionality reduction often satisfy neither criteria. This stems from \textit{non-identifiability} of the LDA model (i.e. there are many \textit{locally optimal} topic models for a given corpus that produce non-comparable low-dimensional clusterings of the documents; these clusterings are not necessarily interpretable to humans).
Existing literature on LDA focuses on mitigating non-identifiability \cite{anchor-word, hierarchical}, improving stability \cite{stability-constraints, ldade}, or guiding model inference towards more useful topics (e.g. topics that are predictive of a downstream task \cite{pf-slda}). However, to our knowledge, there are no works on ensuring that low-dimensional projections of corpora based on LDA preserve semantically meaningful notions of document similarity. 

In this work, we propose an LDA-based method for learning topic models with human-in-the-loop feedback, which produce semantically meaningful low-dimensional projections of corpora. We call our method \emph{Semantically-Aligned-Projection (focused) supervised LDA (SAP-sLDA)}. SAP-sLDA is a semi-supervised method that regularizes the LDA training objective with a ``interpretability" term based on a small set of human labeled documents. The regularizer penalizes the model for producing low-dimensional projections wherein documents with the same label are spread far apart and those with different labels are clustered together. Through active learning, we acquire the set of document labels by querying human experts.

On synthetic corpora, with access to ground-truth document labels, we demonstrate that SAP-sLDA yields projections that satisfy our interpretability criteria, while baselines struggle. On the Dharma Seed corpus, we explore manual labeling under two active learning regimes, and show that a semantically aligned low-dimensional projection of the corpus is achievable with as few as 15\% of documents labeled. Finally, preliminary comparisons on the Dharma Seed corpus shows that our method produces more human-interpretable low-dimensional projections.

\section{Related Works}


\textbf{Metric Learning on Text Data.} Our goal of producing low-dimensional projections of text corpora wherein distances in the projection space are semantically meaningful can be recast as metric learning -- the task of learning a distance function over a set of objects. 

Since early works on metric learning for text data which improved upon the Euclidean distance metric for characterizing similarity \cite{labanon}, several works have explored using \textit{additional information} to optimize a target distance function class subject to constraints. 

In fully-supervised methods, each document in the corpus is associated with a label -- documents with the same label are constrained to be similar while those with different labels are constrained to be dissimilar. However, accurate, computationally efficient implementations often require that the \textit{entire} corpus be labelled \cite{fully-supervised} which is infeasible for the Dharma Seed corpus where labels are not available by default.

In semi-supervised methods, document similarity constraints are explicitly specified \cite{xing, kumar}. The number of constraints is \textit{combinatorial} in the number of documents annotated, rendering these methods impractical in our use-case. SAP-sLDA is able to achieve an interpretable clustering of documents with \textit{one} label per document for a \textit{fraction} of the corpus.

Furthermore, none of these works explicitly utilize the learned metric to project documents into a 2-dimensional space in a way that empowers people to explore the documents, which is the goal of SAP-sLDA.

\textbf{Human in the Loop Topic Modelling.} For improving the interpretability of topics learned by LDA, there is a body of work that incorporates human feedback during model training, human in the loop (HiTL) topic modelling.

A great deal of research has focused on HiTL interventions on LDA to improve topic interpretability. In a vast majority of these works, humans intervene on the topics themselves performing operations like \textit{topic merging}, \textit{topic splitting}, \textit{word removal} and \textit{word addition} \cite{closing-the-loop, convisit, itm}. The success of these methods, however, depends on the learned topics being reasonably interpretable to begin with, which does not hold for the Dharma Seed corpus due to non-identifiability. In SAP-sLDA human feedback is in the form of providing labels for documents, which is independent of the quality of the topic model itself.

\section{Background}
We briefly review Latent Dirichlet Allocation (LDA) on which SAP-sLDA is based \cite{lda}.


LDA operates on a \textit{corpus}, or a collection of documents, where each \textit{document} is a vector of \textit{word} frequencies. The generative process for each document $d \in \{1, \cdots, D\}$ is as follows. First, we draw a distribution over $K$ topics $\theta_d \sim \text{Dir}(\alpha)$. To generate a word, we first draw a topic $z_{n, d} \sim \text{Cat}(\theta_d)$ and subsequently draw a word from the chosen topic's distribution over words $w_{n, d} \sim \text{Cat}(\beta_{z_{n, d}})$. The objective of LDA is to maximize the evidence lower bound (ELBO). The full form of the LDA ELBO is provided in Appendix A.1.



\section{Semantically-Aligned-Projection (Focused) Supervised LDA (SAP-sLDA)}
\subsection{The SAP-sLDA Objective}
SAP-sLDA modifies the LDA objective to explicitly achieve the desired interpretability criteria. Recall that SAP-sLDA is a semi-supervised method, and thus has access to human-provided \textit{labels} for a small subset of documents. To simplify the formulation, let there be $L$ possible labels of which exactly one must be assigned to each document in the subset of documents we have chosen to label. For each label $\ell \in L$, let $S_\ell$ denote the set of documents that were assigned that label. Let $f : \mathbb{R}^T \rightarrow \mathbb{R}^2$ be a dimensionality reduction algorithm. 

SAP-sLDA adds the following \textbf{regularization term} to the LDA ELBO (the full form of the LDA ELBO can be found in Appendix A.1):
\begin{multline}
    \sum_{\ell = 1}^L \sum_{\ell' = 1}^L \mathbb{I}(\ell \neq \ell') \cdot \lambda_1 \mathrm{dist} (S_\ell, S_{\ell'}, \lambda_2) \\
    -\mathbb{I}(\ell = \ell') \cdot \lambda_3\mathrm{dist}(S_\ell, S_{\ell'}, \lambda_4)
\end{multline}
where \texttt{dist} refers to the cumulative $p$-norm distance:
\begin{multline}
     \mathrm{dist}(S_\ell, S_{\ell'}, \lambda) = \sum_{x \in S_\ell}\sum_{x' \in S_{\ell'}} ||f(x) - f(x')||_{\lambda}
\end{multline}

In this paper we let $f$ represent applying PCA which is re-fit every iteration on the current MLE of $\theta$. 

Intuitively, our training objective aims to \textit{maximize} the ELBO (i.e. finding a topic model that explain the corpus well), while simultaneously: (A) maximizing the first line of equation (1), which prefers \textbf{documents with different labels being far apart} in the low-dimensional projection, and (B) minimizing the second line of equation (1), which prefers \textbf{documents with the same label being close together}. $\lambda_1$ and $\lambda_3$ are hyperparameters that control the strength of ``interpretability" regularization, while $\lambda_2$ and $\lambda_4$ affect the distance metric itself. Note that documents whose labels are not provided do not contribute to the regularization term.

\subsection{Training Pipeline}
Recall that labels are not available by default for the Dharma Seed corpus. Thus, SAP-sLDA comprises a two-step algorithm to incrementally label the corpus until an interpretable projection is achieved. Iteration $i$ of the algorithm proceeds as follows.


\textbf{Active Learning for More Labels.} The first step is to use active learning to (1) \textbf{choose a set $\mathcal{D}_i$ of unlabelled documents to label} and (2) \textbf{query the labels for $\mathcal{D}_i$ from human experts}. There are several ways to determine the set $\mathcal{D}_i$ for (1) with the desideratum being to choose the set of documents whose corresponding $\{\theta_d\}_{d \in \mathcal{D}_i}$ the optimizer is least ``certain" about. We propose one such active labelling scheme in Section 4.3. For (2), we note that there are numerous \textit{classes} of labels which could be assigned to documents, which may provide varying levels of signal about $\{\theta_d\}_{d \in \mathcal{D}_i}$. For example, one way of labelling could be to have humans decide what thematic keyword best aligns with the content of a document. Alternatively, humans could label documents based on the type of piece (anecdote, poem, non-fiction, etc.). It is not immediately clear what label class provides the \textit{most} signal, and it is possible that using signal from multiple label classes in tandem might yield the most interpretable projections. We discuss active learning and label classes in more detail in Section 4.3.

\textbf{Optimization.} The second step in the pipeline is to \textbf{update the regularizer} of the SAP-sLDA objective to include all labelled documents $\mathcal{D}_{0\cdots i}$, and \textbf{retrain the model with $R > 1$ number of random restarts}. For each restart, we project documents into 2-dimensions based on the corresponding $\hat{\theta}$. 

The \textbf{termination criterion for SAP-sLDA is stability of projections across the $R$ restarts}. This can either be determined by visual inspection or by calculating the pairwise distances in the projected space between all documents and checking that the cumulative variance in pairwise distances across the $R$ restarts is less than $\epsilon > 0$. 

\subsection{Instantiating the Training Pipeline}
While our method presents a general framework, for the purposes of this paper we instantiate our training pipeline by deciding upon label classes and active learning schemes. We first perform a feasibility analysis on a  synthetic corpus whose construction is described in Section 5.1 followed by experiments on a small random sample of documents from Dharma Seed corpus. 

\textbf{Active Learning for More Labels.} On the Dharma Seed corpus, we \textit{do not} use active learning to \textbf{choose a set $\mathcal{D}_i$ of unlabelled documents to label}, opting to assess the ability of our regularizer to produce interpretable projections in the ideal scenario where all documents are labelled. On the synthetic corpus, in addition to the setting where all labels are provided, we assess the efficacy of two ways of deciding which documents to label. 

\begin{enumerate}
    \item 
    \textbf{(Baseline) Labelling random documents.} As a naive baseline, we choose 5\% of documents at random to add to the set of labelled documents at every iteration. 
    \item 
    \textbf{Labelling documents with the highest variance in relative position.} For each random restart in an iteration, we calculate the cumulative distance from every document to every other document in the projected space. We then calculate the variances of these pairwise distances across the three runs and sum them up, labelling the 5\% of documents for which this quantity is largest.
\end{enumerate}

\textbf{Label Classes.} On the synthetic corpus labels are generated along with documents in a Bayesian fashion (see Section 5.1). On the Dharma Seed corpus we test three label classes that provide varying amounts of signal about the document-topic distribution $\theta$.

\begin{enumerate}
    \item \textbf{Labelling documents randomly} from 0 to 9 provides \textit{no signal} about $\theta$.
    \item \textbf{Labelling documents by their author} provides \textit{minimal signal} about $\theta$ since some authors might consistently give talks on the same topic while others may give talks on a variety of topics.
    \item
    \textbf{Labelling documents by broad theme} 
     -- whether they fall under \textit{Buddhism in Practice} or \textit{Reflections on the World} -- provide \textit{low to moderate signal} about $\theta$ since these themes relate to documents' topics, but are extremely broad. We used Chat-GPT to generate these labels, which are provided for only 50\% of the corpus to avoid over-regularization (see Appendix A.3 for details). 
\end{enumerate}

\textbf{Termination Criterion.} We assess the convergence of our method visually by determining whether projections look similar across random restarts. It is also possible for our algorithm to \textit{fail to converge} (projections vary significantly across restarts). In this case, we visually determine the \textit{degree} of failure depending on the level of perceived instability.

\section{Empirical Evaluation of SAP-sLDA on Synthetic Corpus}
We compare SAP-sLDA to baseline methods on a range of synthetic corpora as a feasibility exploration.

\subsection{Generation of Synthetic Corpora}
On synthetic corpora where we have access to the ground-truth data generating models, we answer the following:
\begin{enumerate}
    \item Does SAP-sLDA consistently learn representations that yield qualitatively similar low-dimensional projections as the ground-truth data generating models?
    \item Do the projections corresponding to SAP-sLDA satisfy our two interpretability criteria? If so, how many documents need to be labelled in order for this to be the case?
    \item Is SAP-sLDA able to recover the parameters of the ground-truth data generating model? 
\end{enumerate}
For all synthetic corpora the number of ground-truth topics is fixed at $K = 4$. By design, the first three topics are \textit{highly predictive} of the document label while the fourth is not -- formally, we mean that 

\begin{align*}
    y_d \sim \text{Cat}\left(\text{Softmax}\left(
    \begin{bmatrix}
        10 & 0 & 0 & 0\\
        0 & 10 & 0 & 0\\
        0 & 0 & 10 & 0\\
        0 & 0 & 0 & 0
        \end{bmatrix} \cdot \theta_d^T \right)\right)
\end{align*}
where $y_d$ is the label for document $d$ and $\theta_d$ is document $d$'s topic distribution.

We test three different qualitatively different scenarios for the ground-truth per document topic-distribution, $\theta$.

\textbf{Setting 1 (Single-topic documents):} Documents are predominantly about one of the four ground truth topics -- $\theta \sim \text{Dir}(\alpha)$ where $\alpha_i = 0.001$ for $i \in \{1, \cdots 4\}$. Since only three of the topics are predictive of the document's label, the projection of $\theta$ reveals three single-label clusters (corresponding to documents predominantly about Topics 1, 2 and 3) and one multi-label cluster (corresponding to documents predominantly about Topic 4). 

This setting is chosen to test whether SAP-sLDA achieves interpretable clusterings \textit{without} destroying features of the original data, since over-regularization would manifest in the multi-label cluster failing to appear in the projection of the learned $\hat{\theta}$. 

\textbf{Setting 2 (Mixed-topic documents):} Documents are predominantly about a mix of ground truth topics --  $\theta \sim \text{Dir}(\alpha)$ where $\alpha_i = 1$ for $i \in \{1, \cdots 4\}$. From the projection of $\theta$ we see that the edges are predominantly single-label, while the center is mixed. 

This setting is difficult when 
the word-topic distribution $\beta$ is non-identifiable, since the mixed-topic nature of documents means that they may comprise of highly-prevalent words from \textit{all} topics yielding more locally optimal $\hat{\theta}$.

\textbf{Setting 3 (Predictive-topic documents with Garbage Words):} Documents are about a random mixture of less than 0.5 of \textit{one of} Topics 1, 2 and 3 with greater than 0.5 of Topic 4. The projection of $\theta$ shows three single-label clusters. This setting is intended to mirror real-life corpora where documents tend to be focused on one main topic but may contain portions pertaining to irrelevant side-topics.

This setting is difficult when the word-topic distribution $\beta$ is non-identifiable, since there are fewer words on Topics 1, 2 and 3 in each document, complicating inference for the main topic per document.

\textbf{Varying identifiability of $\beta$.} For each of the settings above we test cases where $\beta$ is \textit{identifiable} (each topic has disjoint support over a quarter of the vocabulary), and \textit{non-identifiable} (the first three topics have support over the first three quarters of the vocabulary, with Topic 1 putting twice as much probability mass on the first quarter as the second half, and so on; Topic 4 has support over the last quarter of the vocabulary).

To assess the interpretability of projections corresponding to each method and to test the ability of each method to recover ground truth parameters, we compare our method with two baselines: LDA and prediction-focused supervised LDA (pf-sLDA) \cite{pf-slda}. pf-sLDA is a semi-supervised method that aims to learn topics that are \textit{predictive} of a document's label, and has a tuneable hyperparameter $p$ that controls how likely a word is to be predictive. we run LDA, pf-sLDA and each iteration of SAP-sLDA with 4 topics for 200 iterations on a corpus of 1000 documents and a vocabulary of 100 words. 
For pf-sLDA we set $p = 0.25$. For SAP-sLDA we set $\lambda_2 = 4$ and $\lambda_4 = 1$. When the word-topic distribution $\beta$ is identifiable, we set $\lambda_1 = 0.5$ and $\lambda_3 = 1$; when it is non-identifiable, we set $\lambda_1 = 5$ and $\lambda_3 = 10$. The values of $\lambda_{1\cdots 4}$ are decided through trial-and-error and are not rigorously tuned.

To evaluate how many documents are required for SAP-sLDA to produce interpretable projections, we begin by running SAP-sLDA with 0 labels given (equivalent to regular LDA) and iteratively add labels for 5\% of the documents either at random, or for the documents which had the highest variance in relative position across three random restarts (see Section 4.3). For all experiments we create projections using the TSNE class from \texttt{sklearn.manifold} with perplexity 20.

\subsection{Results}
\begin{figure*}[hbt!]\centering
\subfloat[Ground Truth $\theta$]{\label{fig.a}
\begin{tabular}[b]{@{}c@{}}
\includegraphics[width=.13\linewidth]{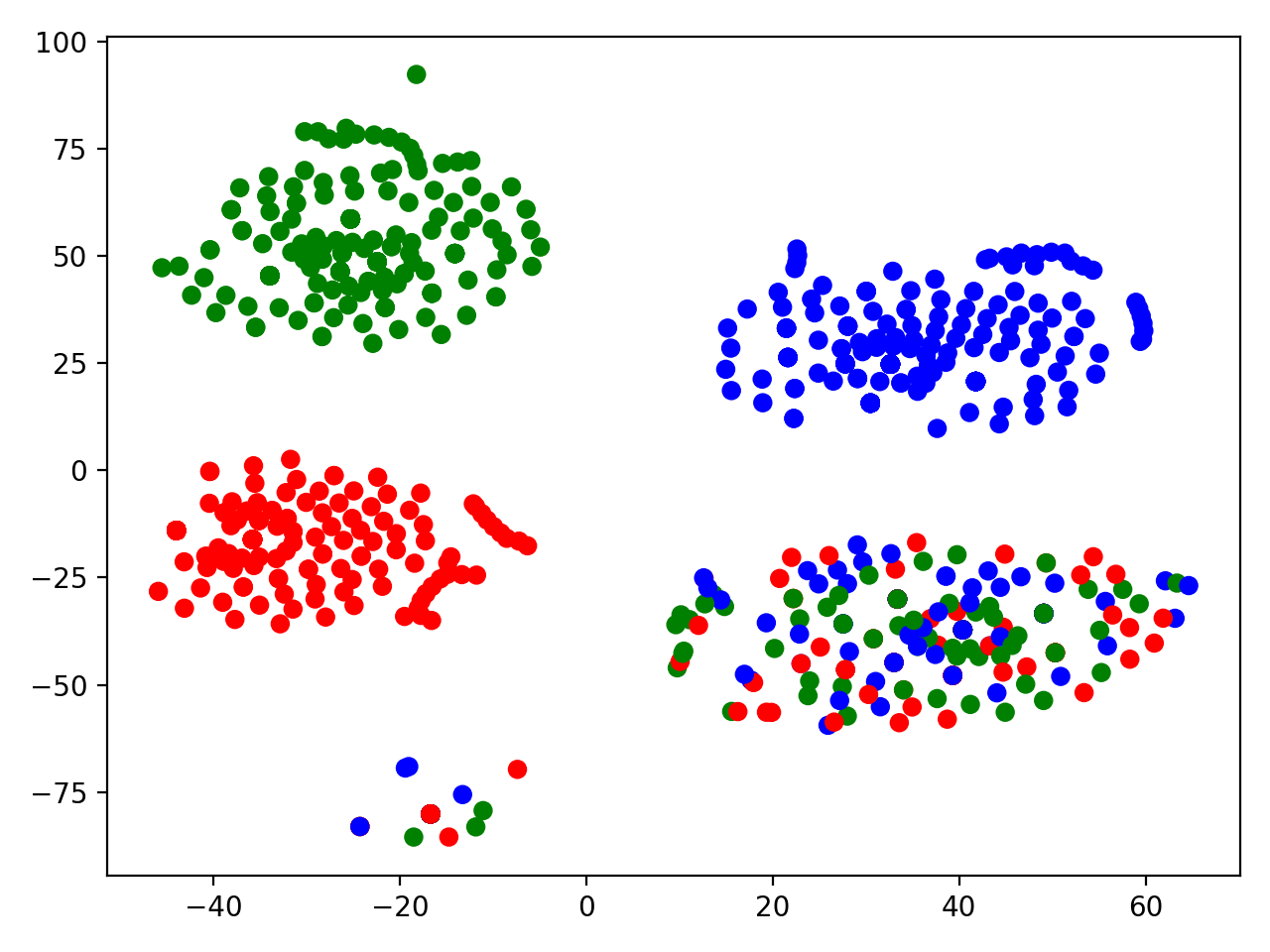}\\[-3pt]
\includegraphics[width=.13\linewidth]{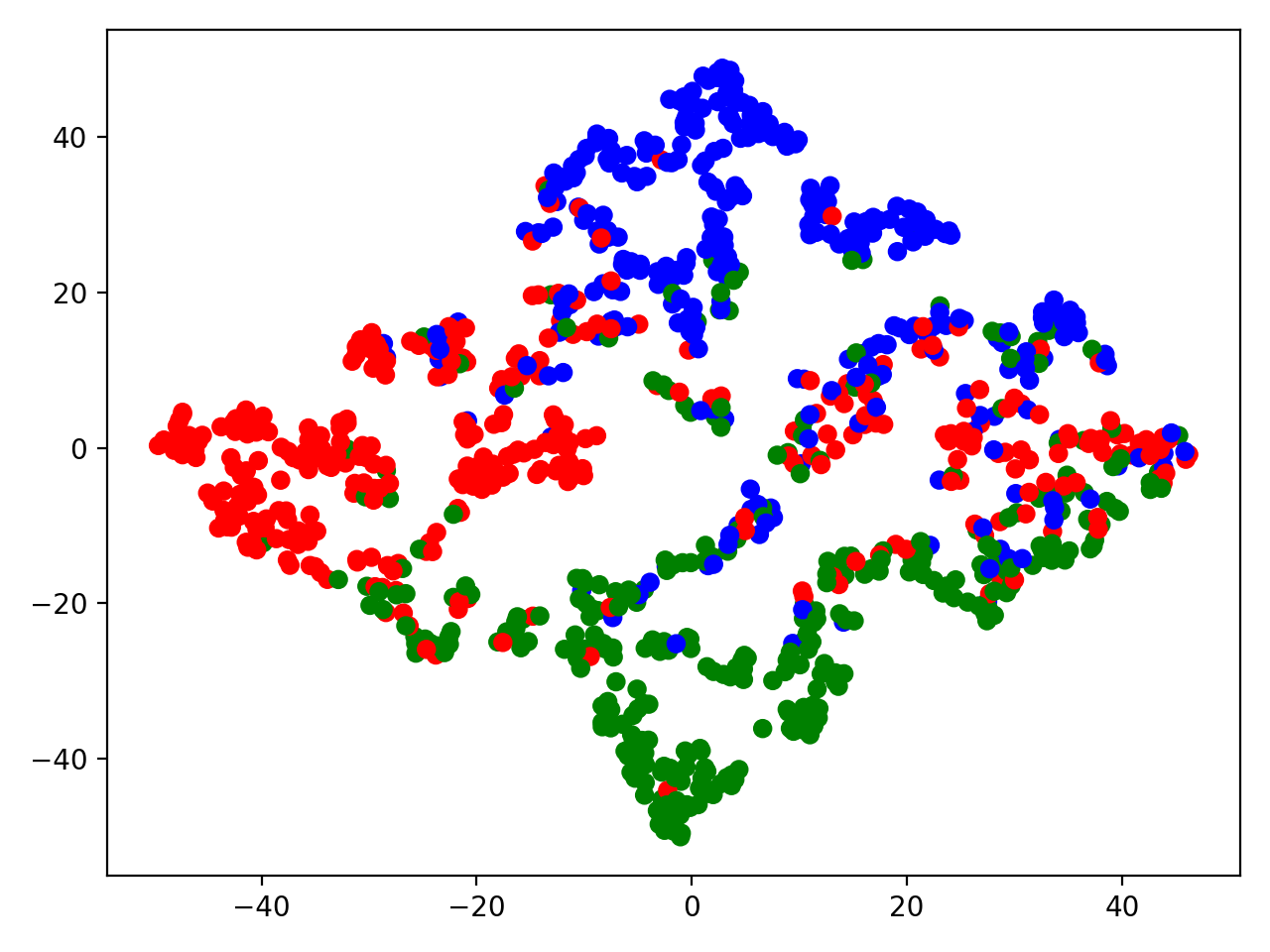}\\[-3pt]
\includegraphics[width=.13\linewidth]{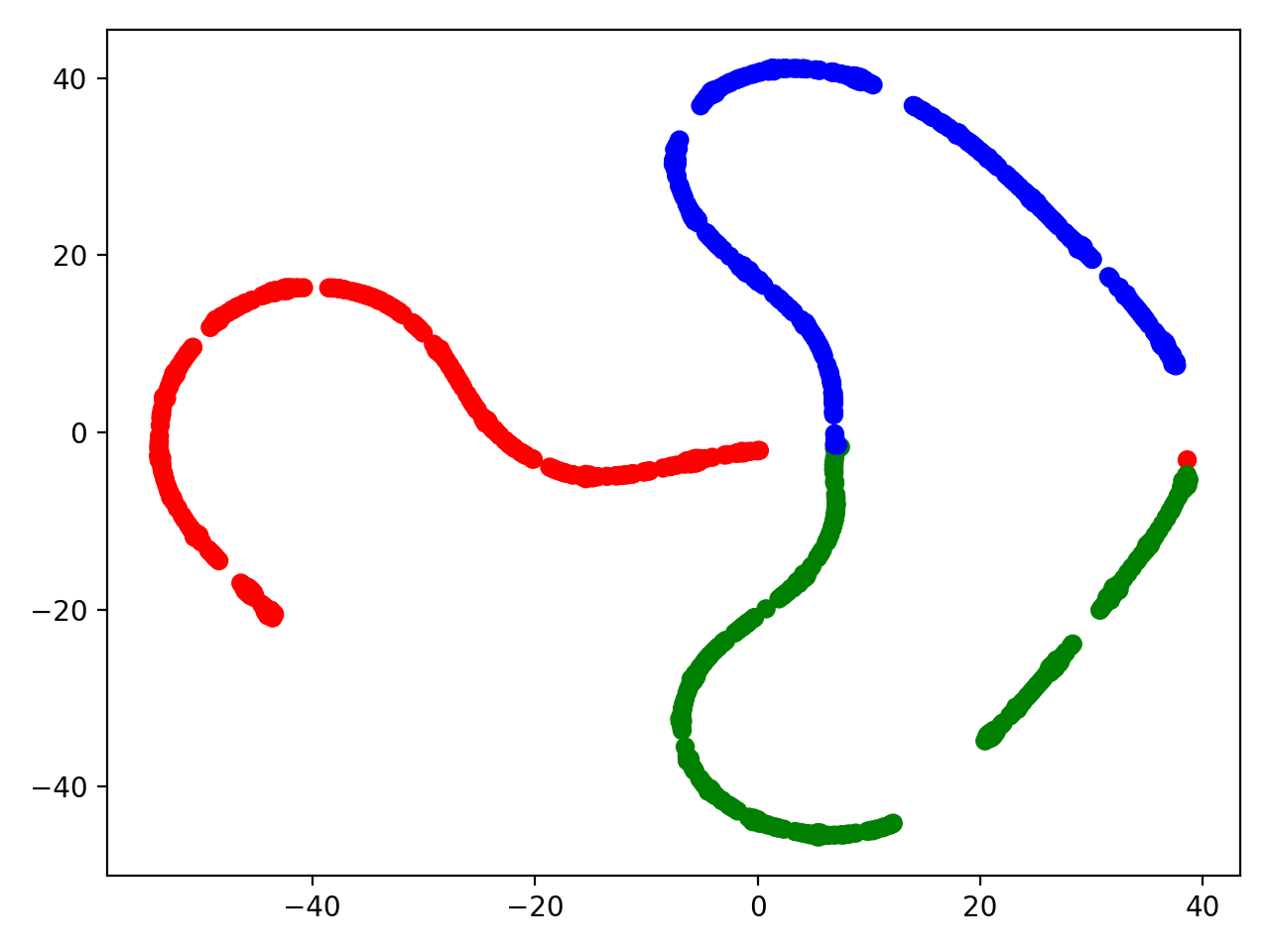}%
\end{tabular}
}
\hfill
\subfloat[LDA]{\label{fig.a}
\begin{tabular}[b]{@{}c@{}}
\includegraphics[width=.13\linewidth]{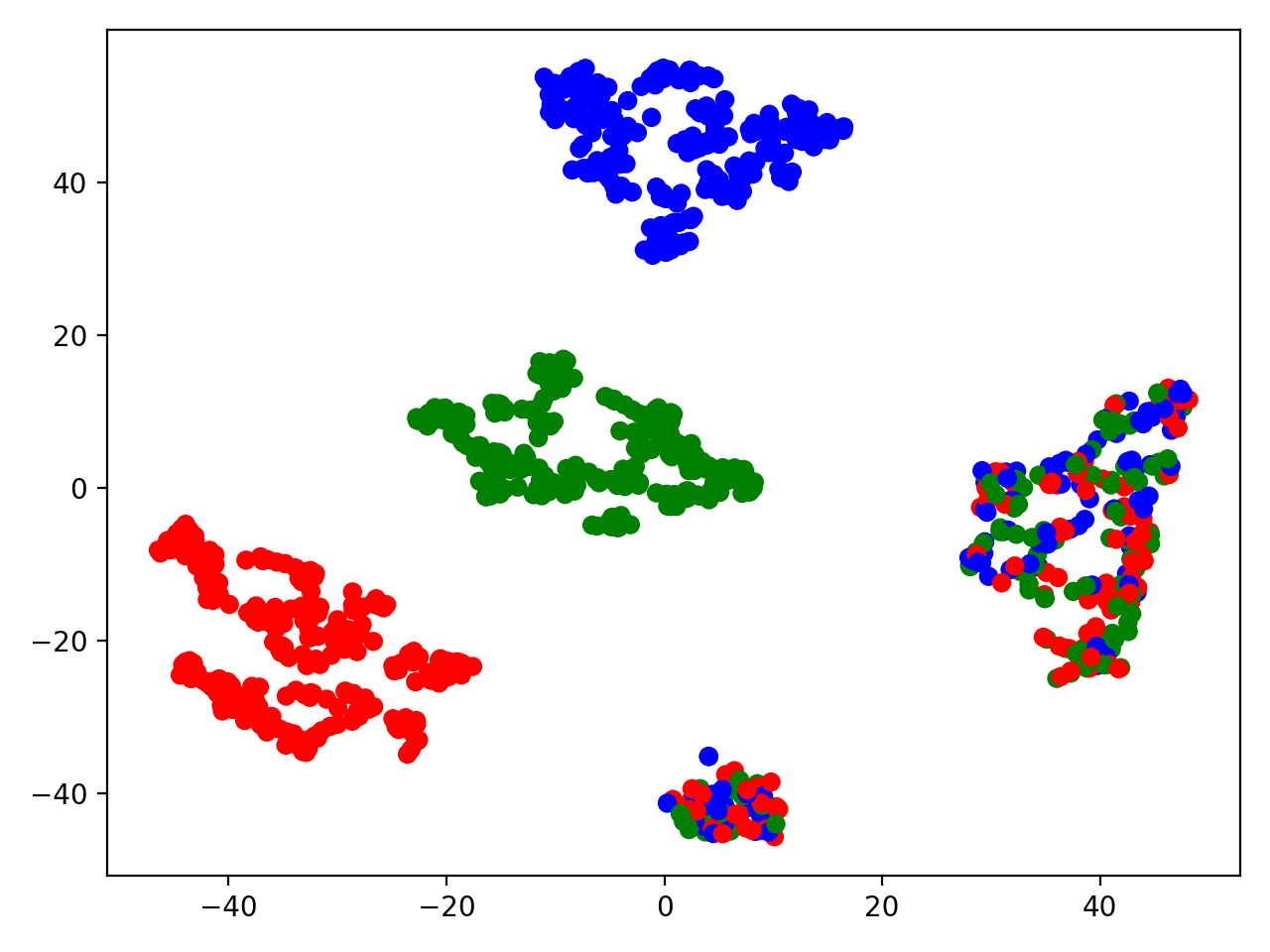}%
\includegraphics[width=.13\linewidth]{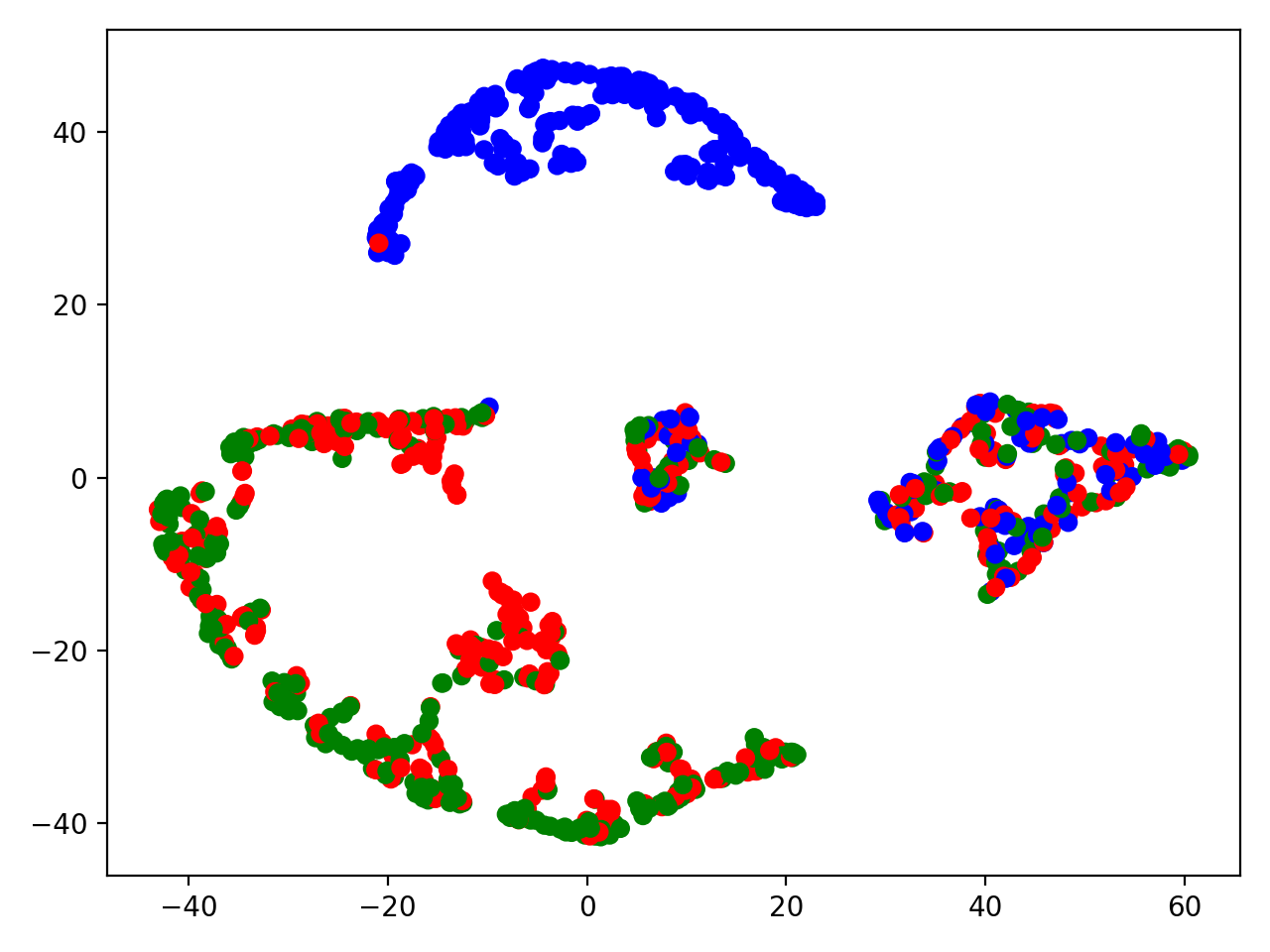}\\[-3pt]
\includegraphics[width=.13\linewidth]{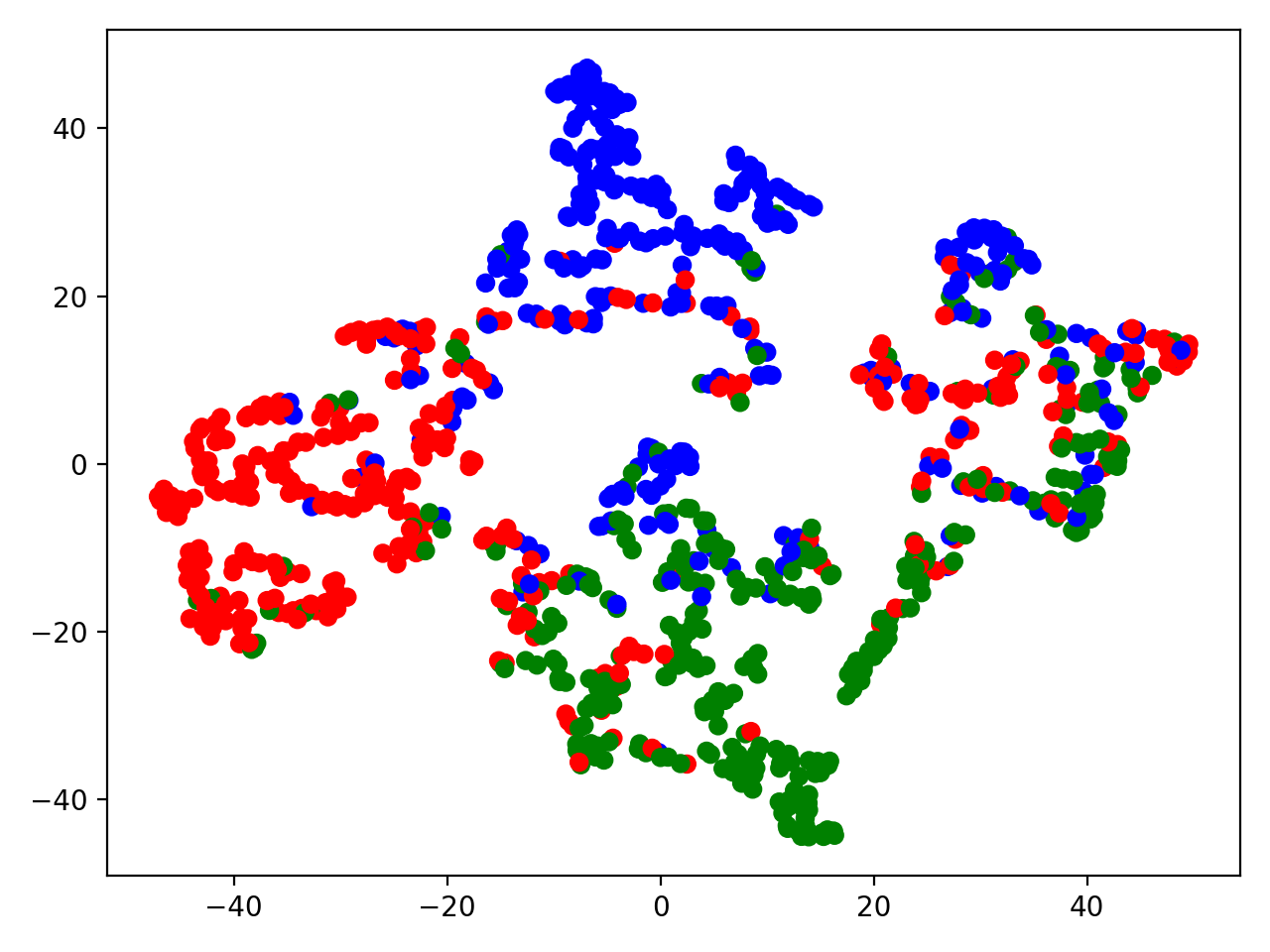}%
\includegraphics[width=.13\linewidth]{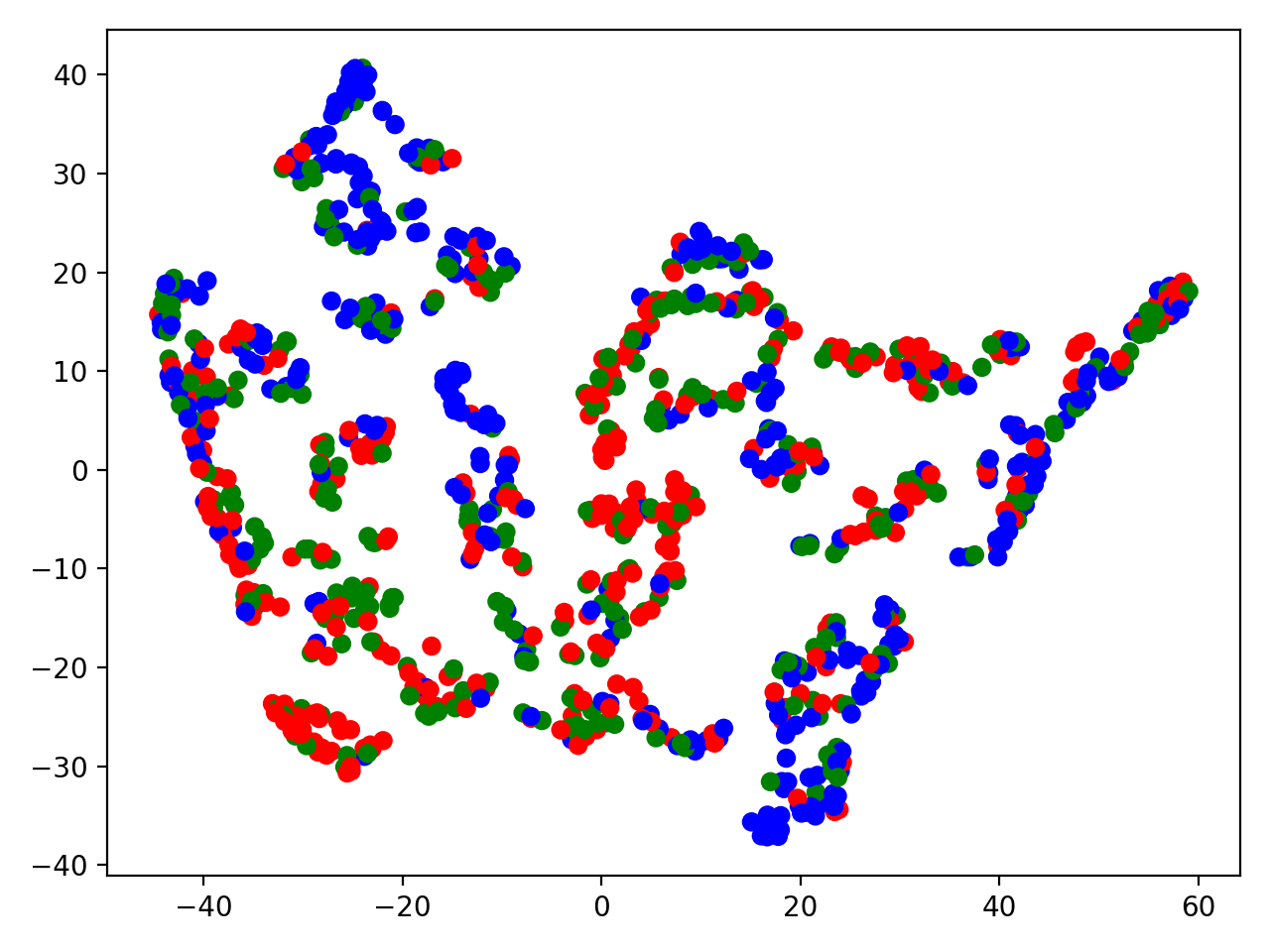}\\[-3pt]
\includegraphics[width=.13\linewidth]{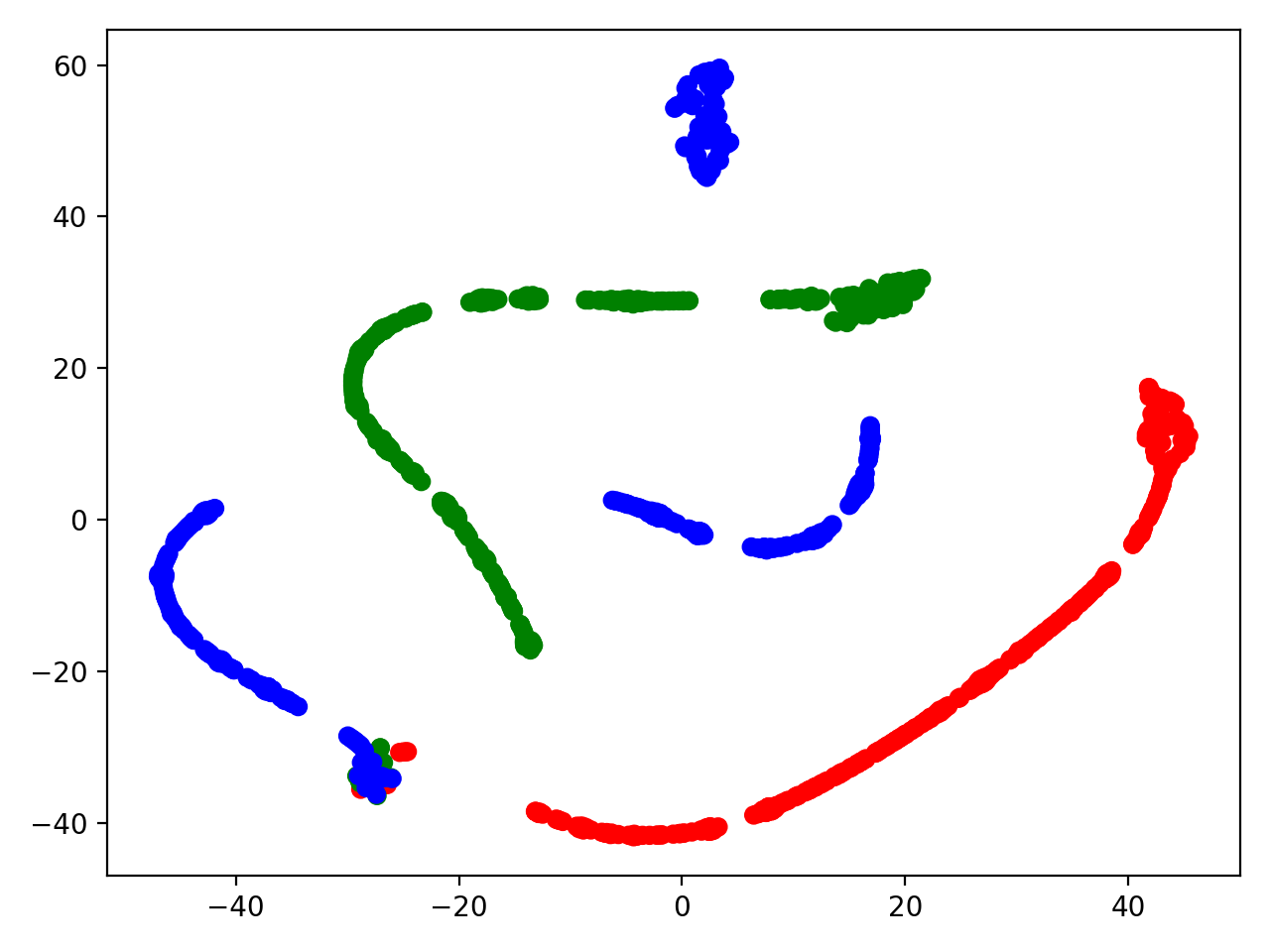}%
\includegraphics[width=.13\linewidth]{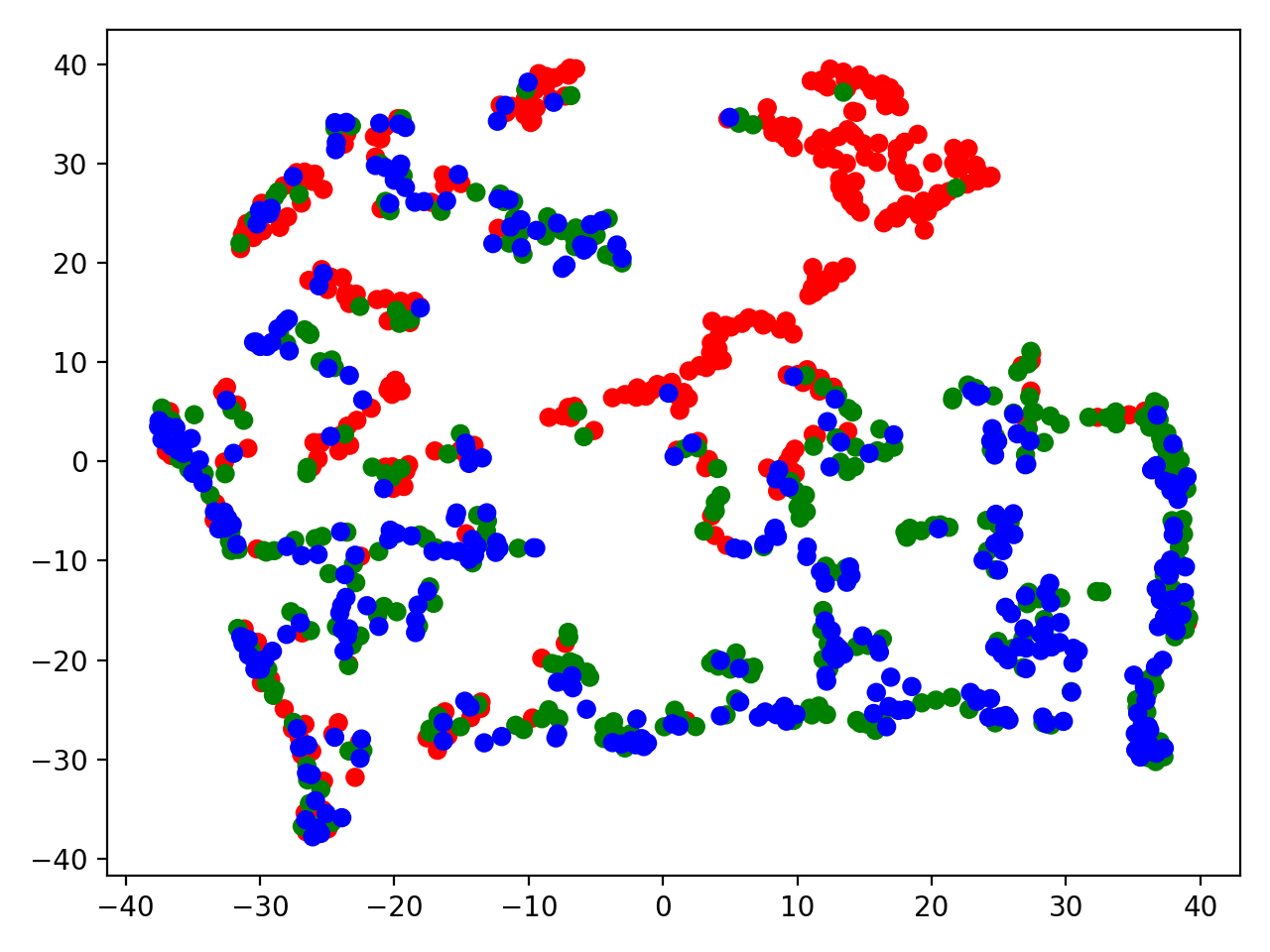}
\end{tabular}
}
\hfill
\subfloat[pf-sLDA]{\label{fig.b}
\begin{tabular}[b]{@{}c@{}}
\includegraphics[width=.13\linewidth]{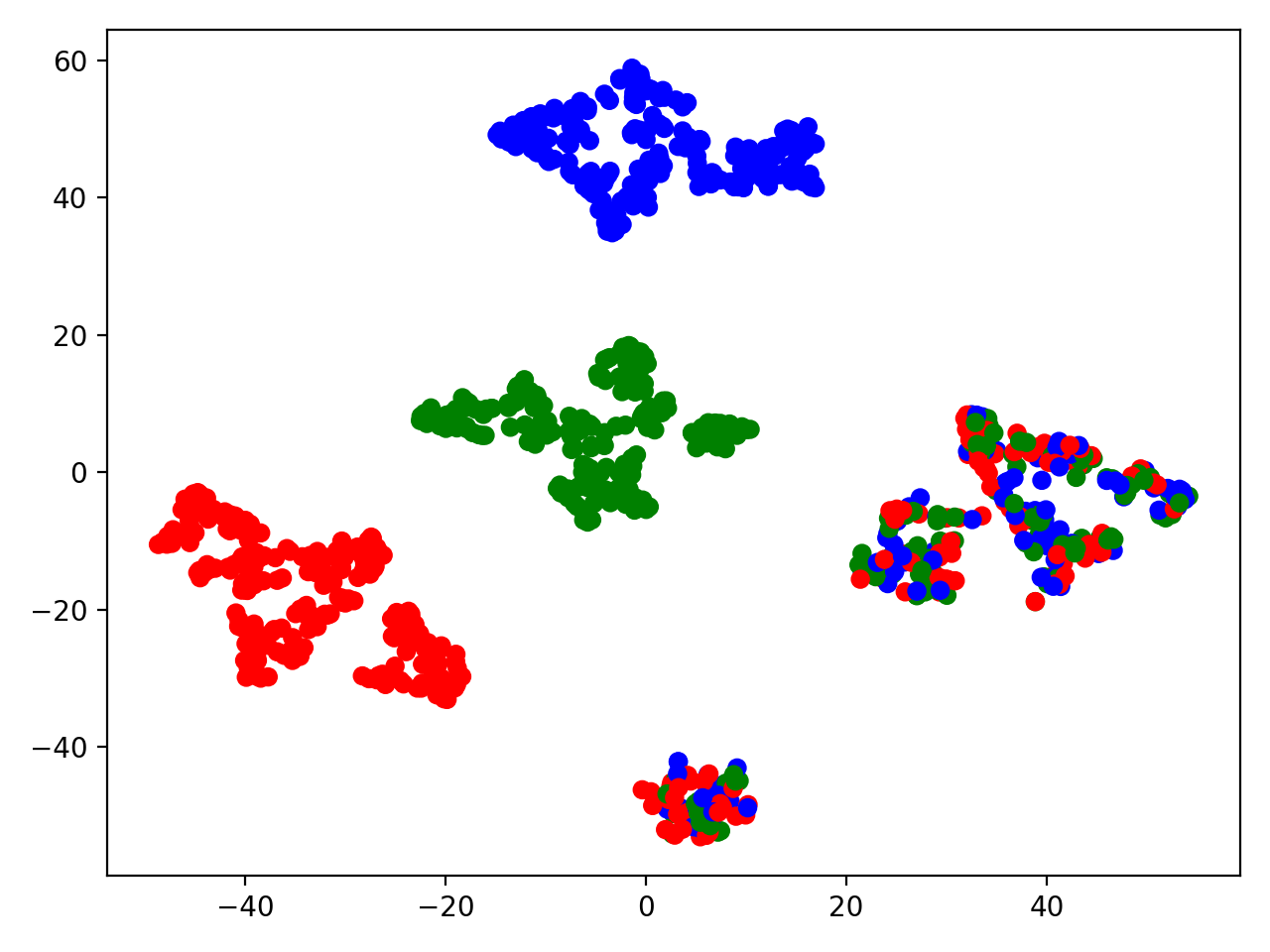}%
\includegraphics[width=.13\linewidth]{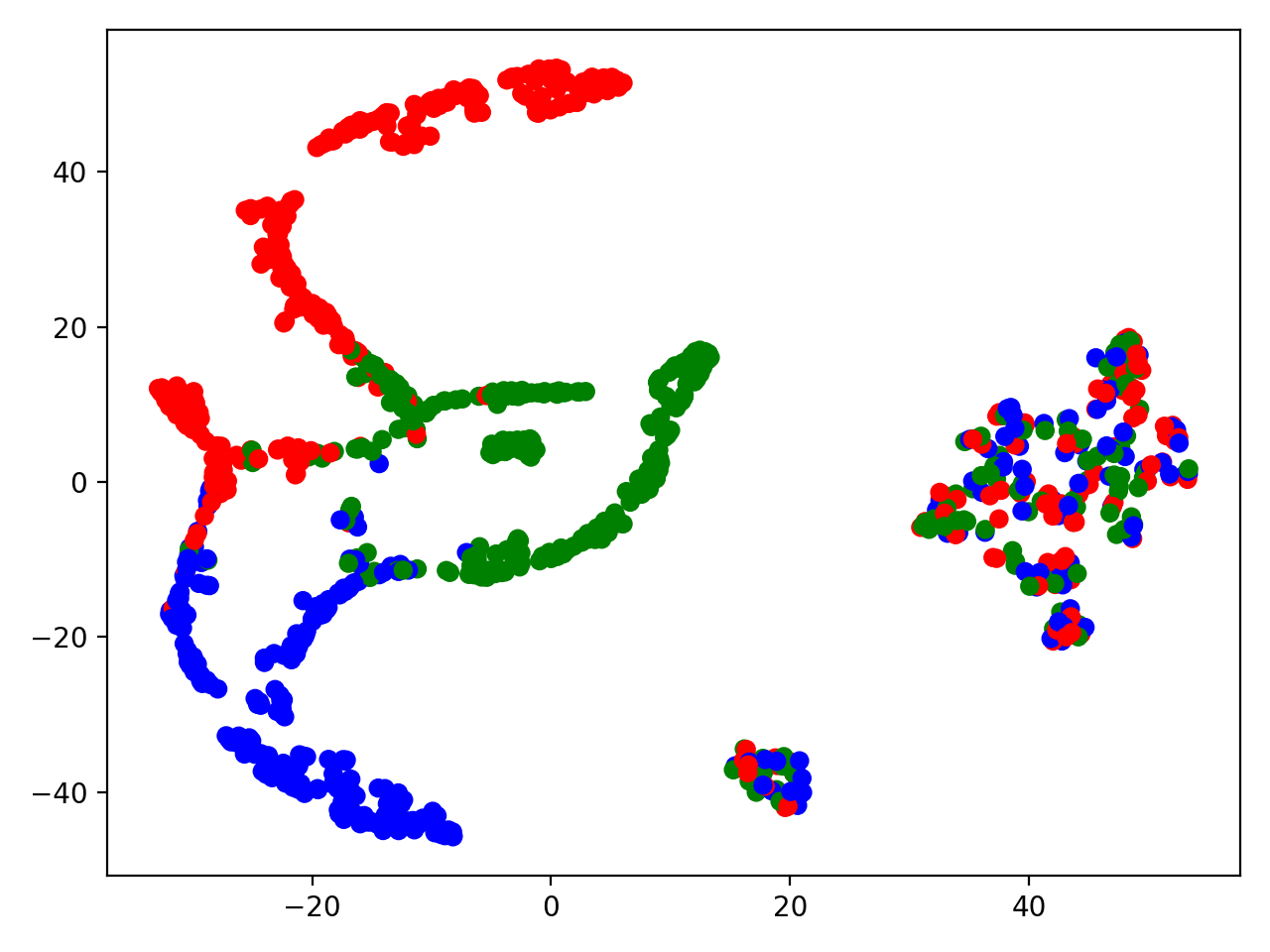}\\[-3pt]
\includegraphics[width=.13\linewidth]{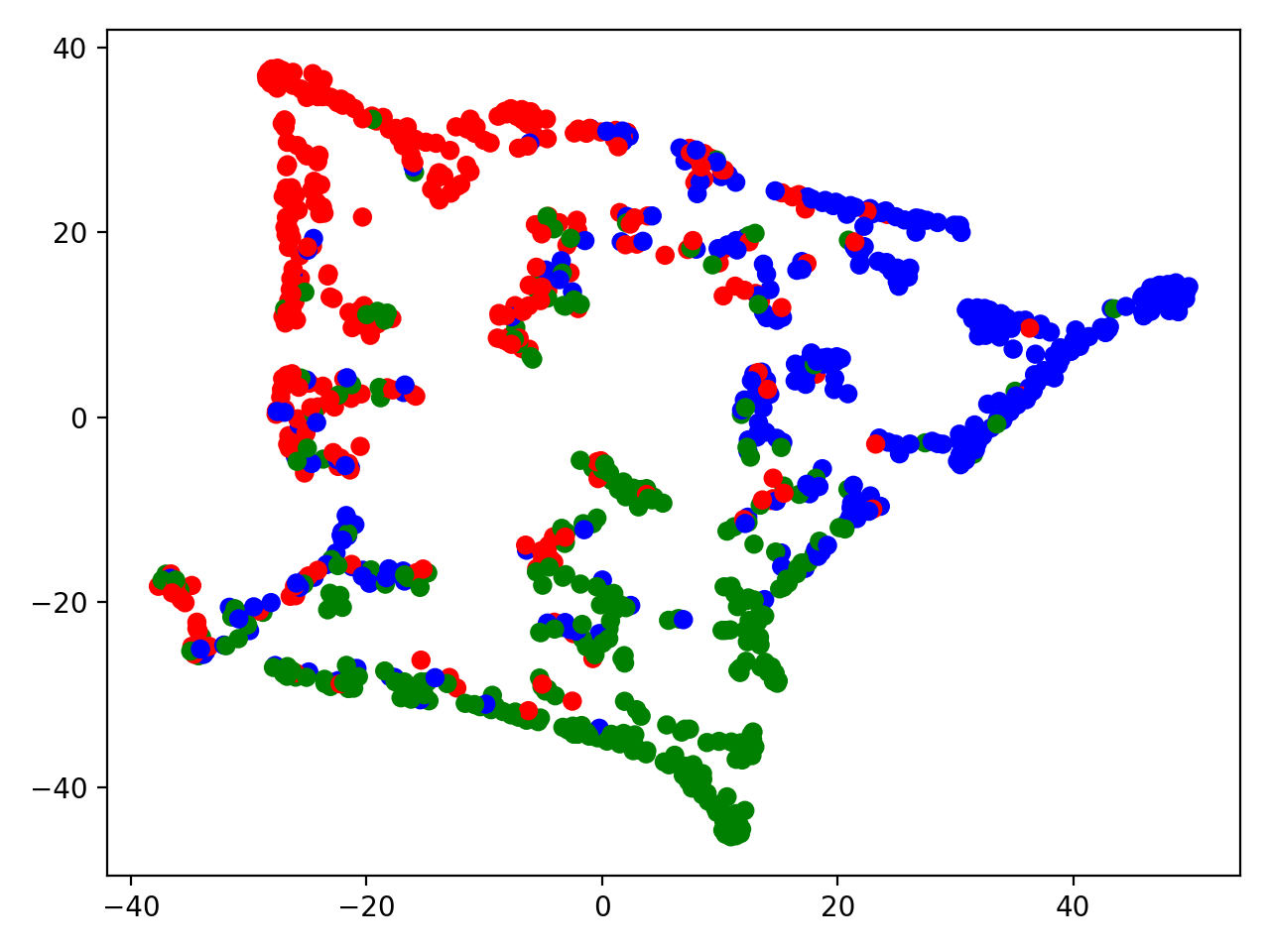}%
\includegraphics[width=.13\linewidth]{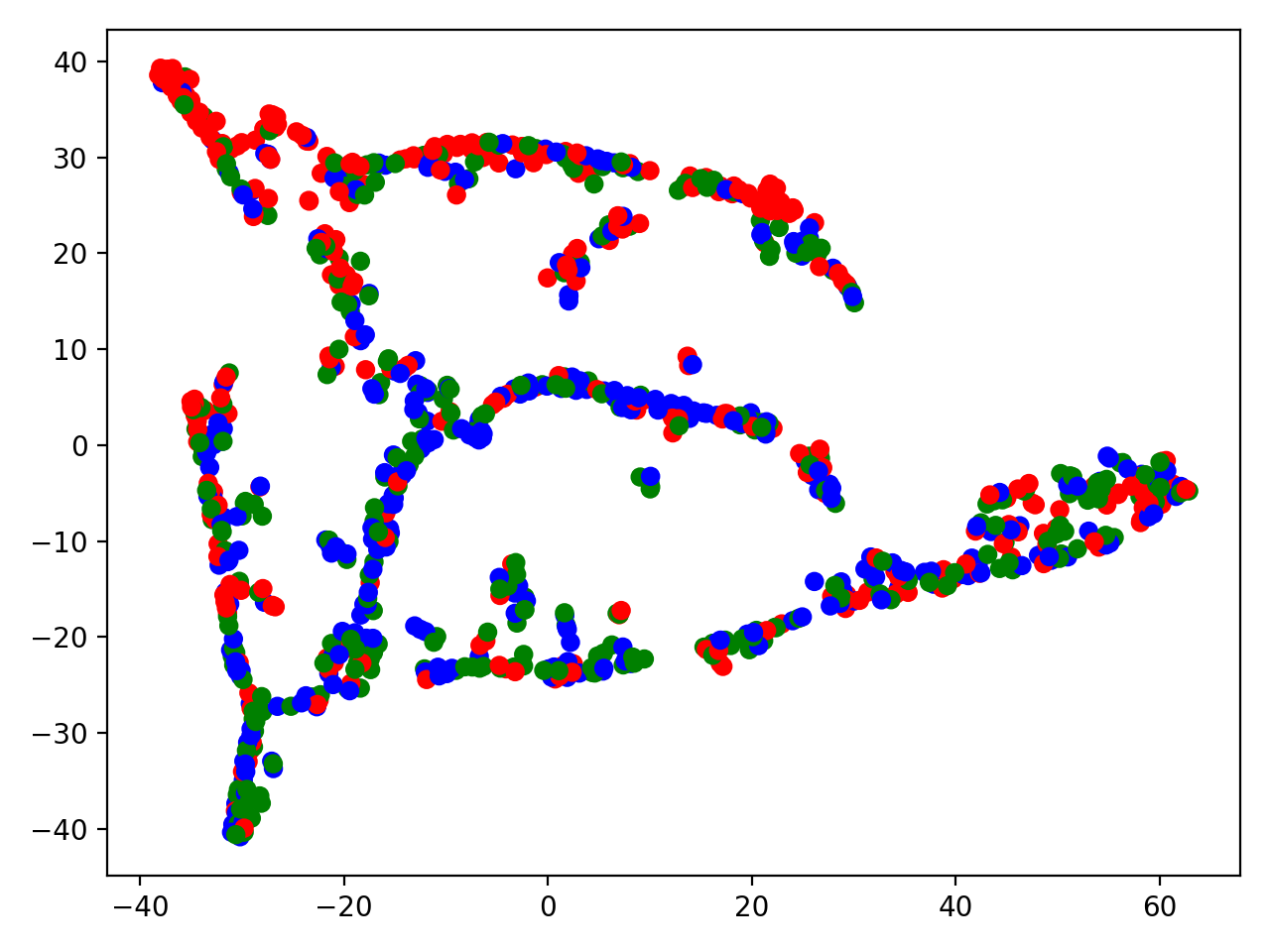}\\[-3pt]
\includegraphics[width=.13\linewidth]{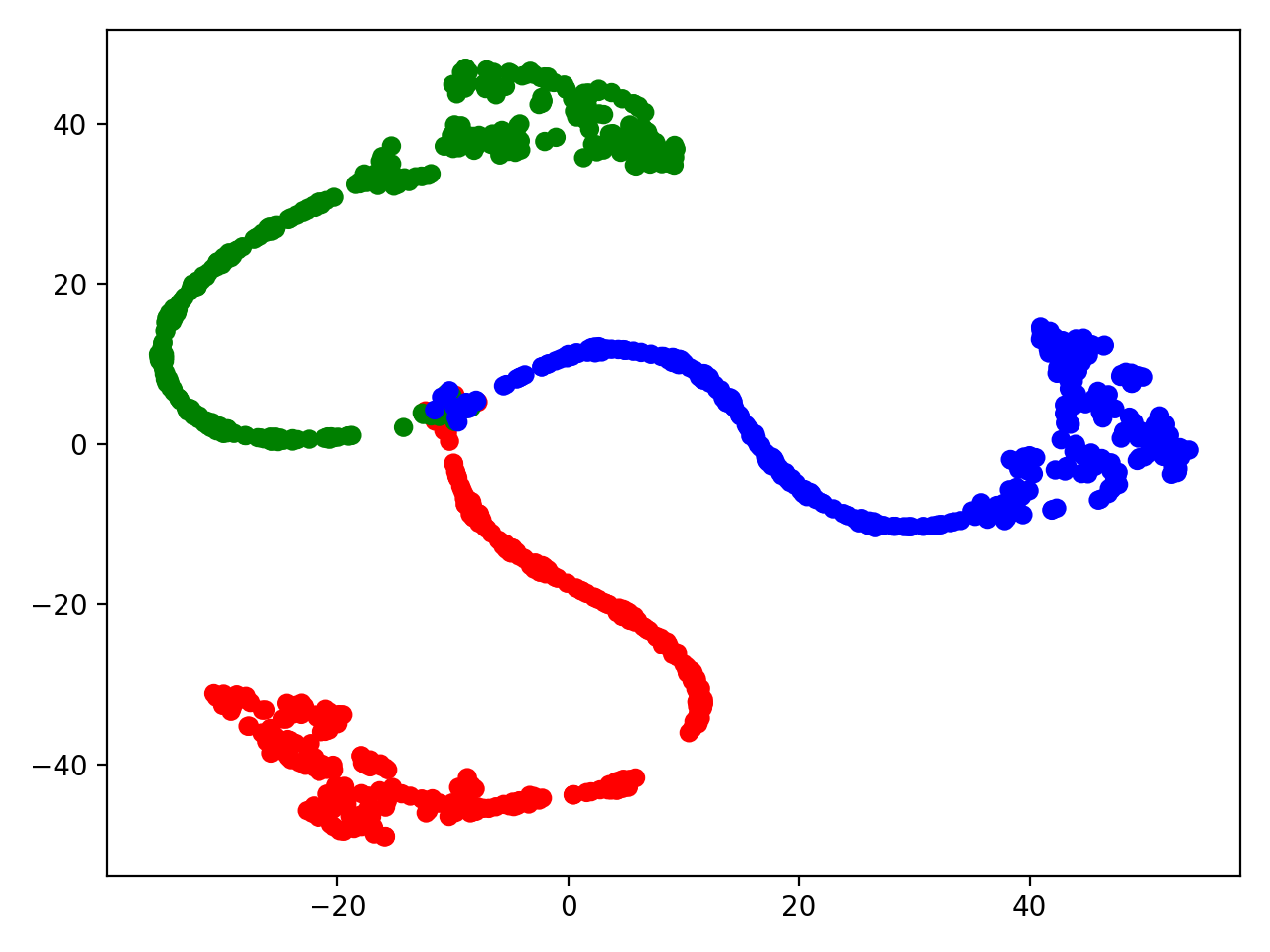}%
\includegraphics[width=.13\linewidth]{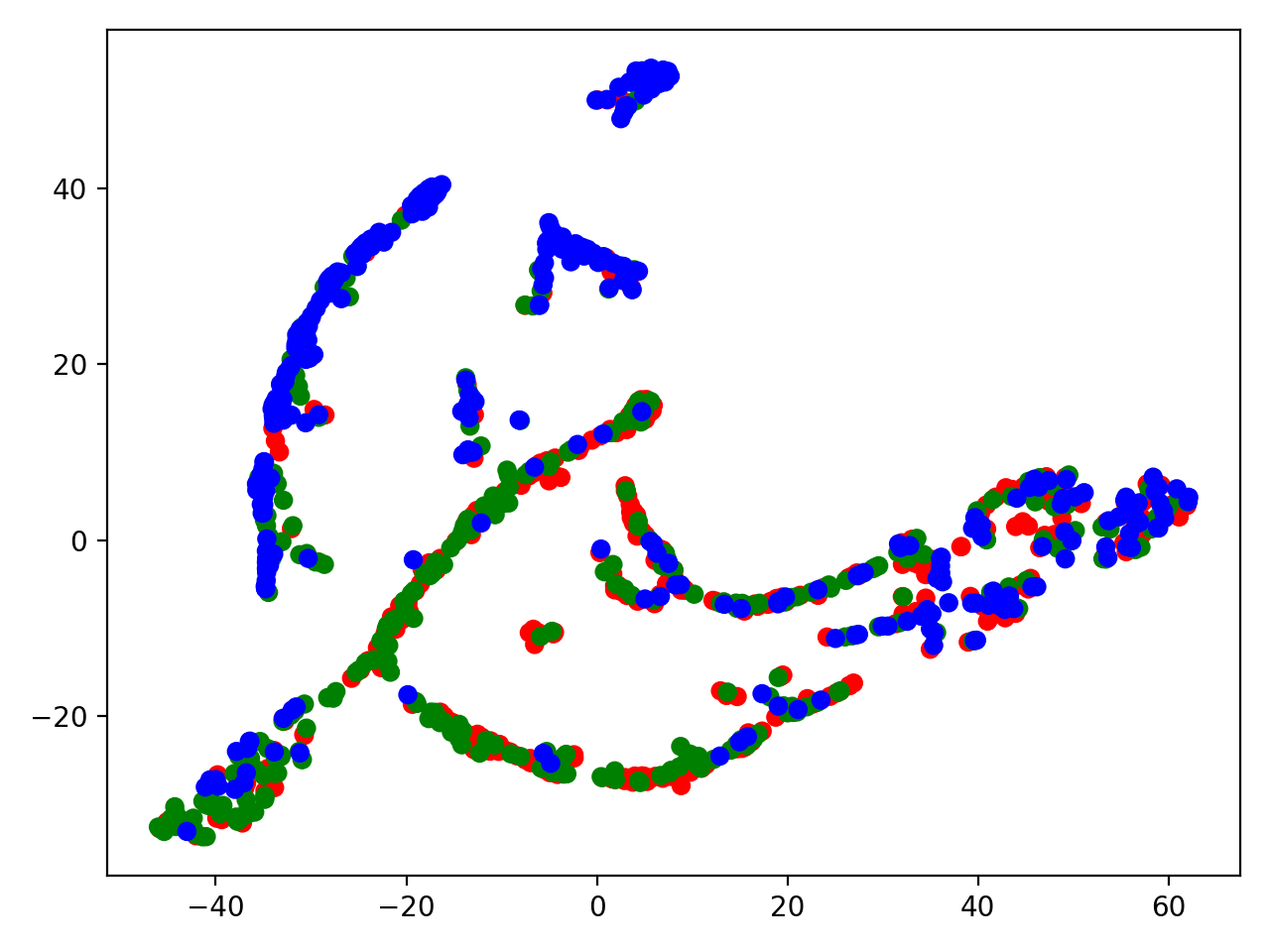}
\end{tabular}
}
\hfill
\subfloat[SAP-sLDA]{\label{fig.c}
\begin{tabular}[b]{@{}c@{}}
\includegraphics[width=.13\linewidth]{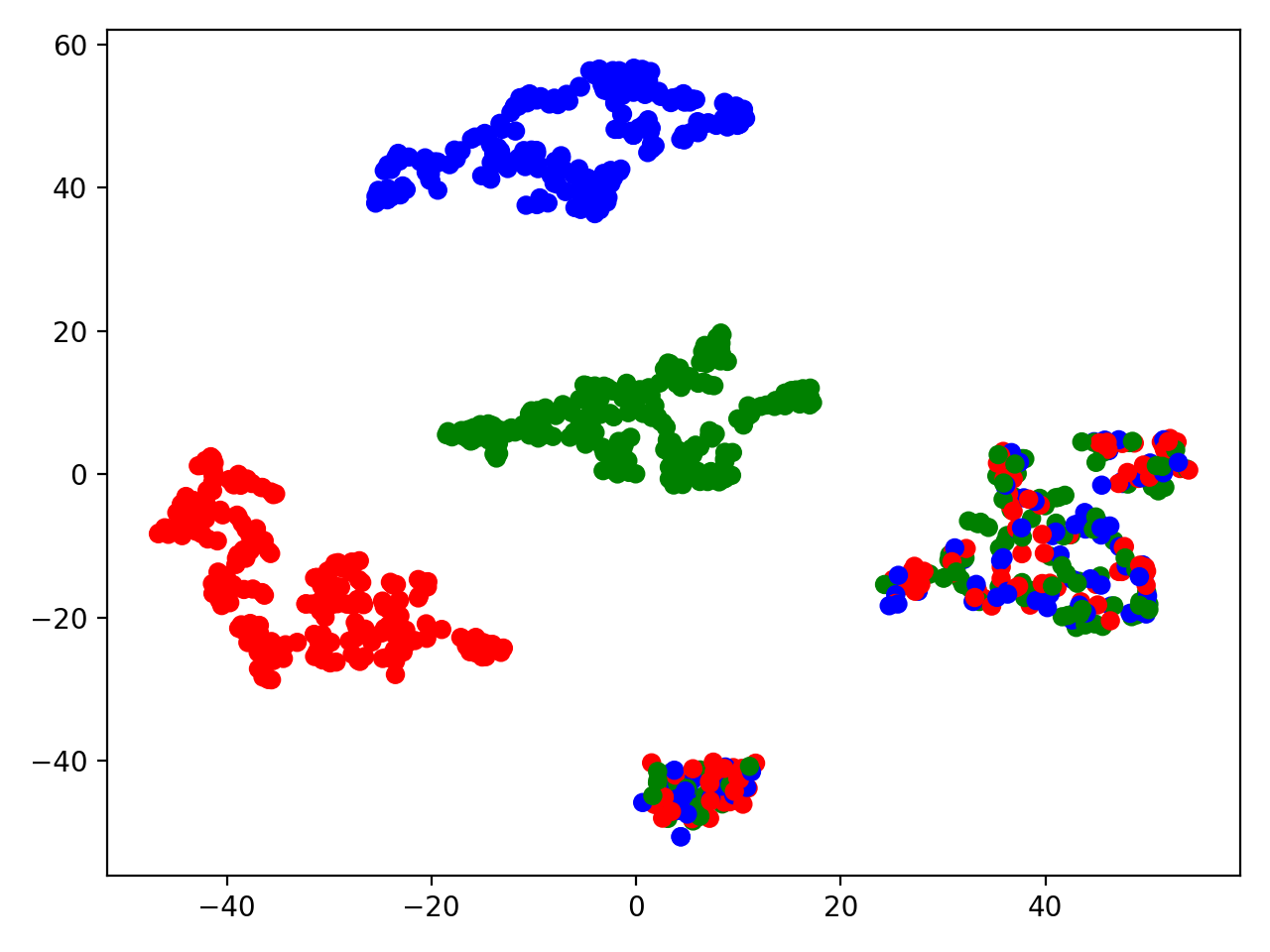}%
\includegraphics[width=.13\linewidth]{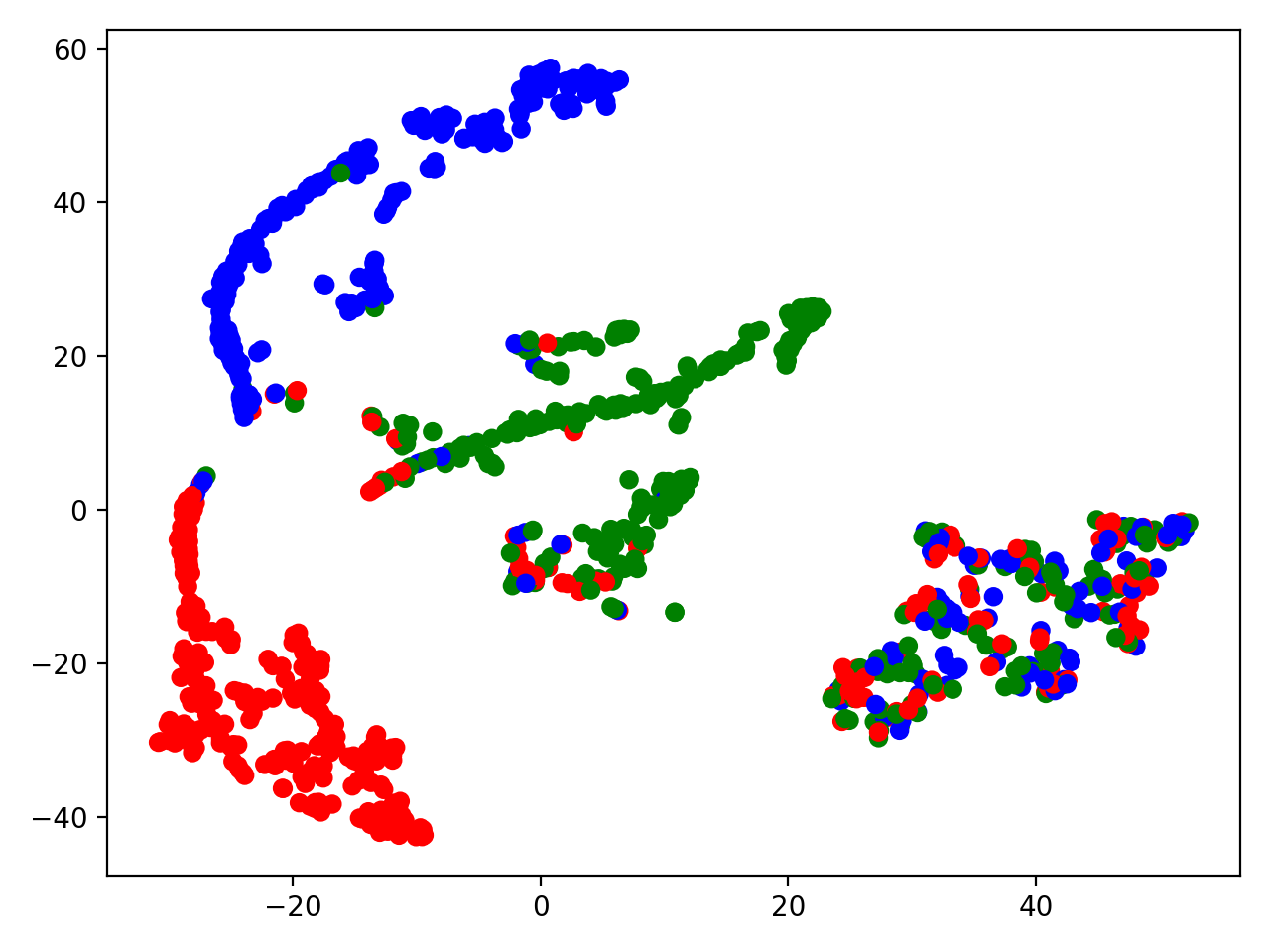}\\[-3pt]
\includegraphics[width=.13\linewidth]{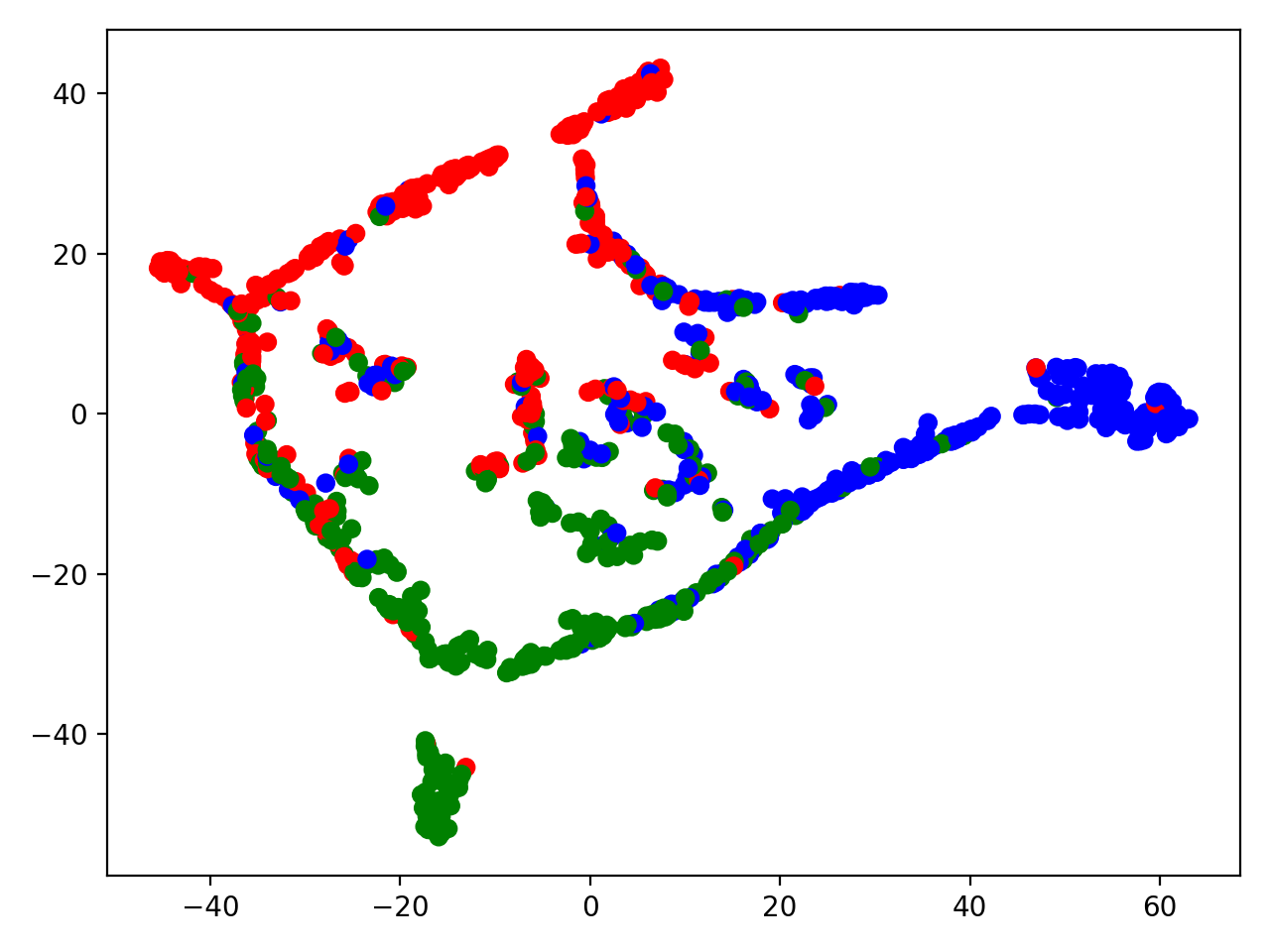}%
\includegraphics[width=.13\linewidth]{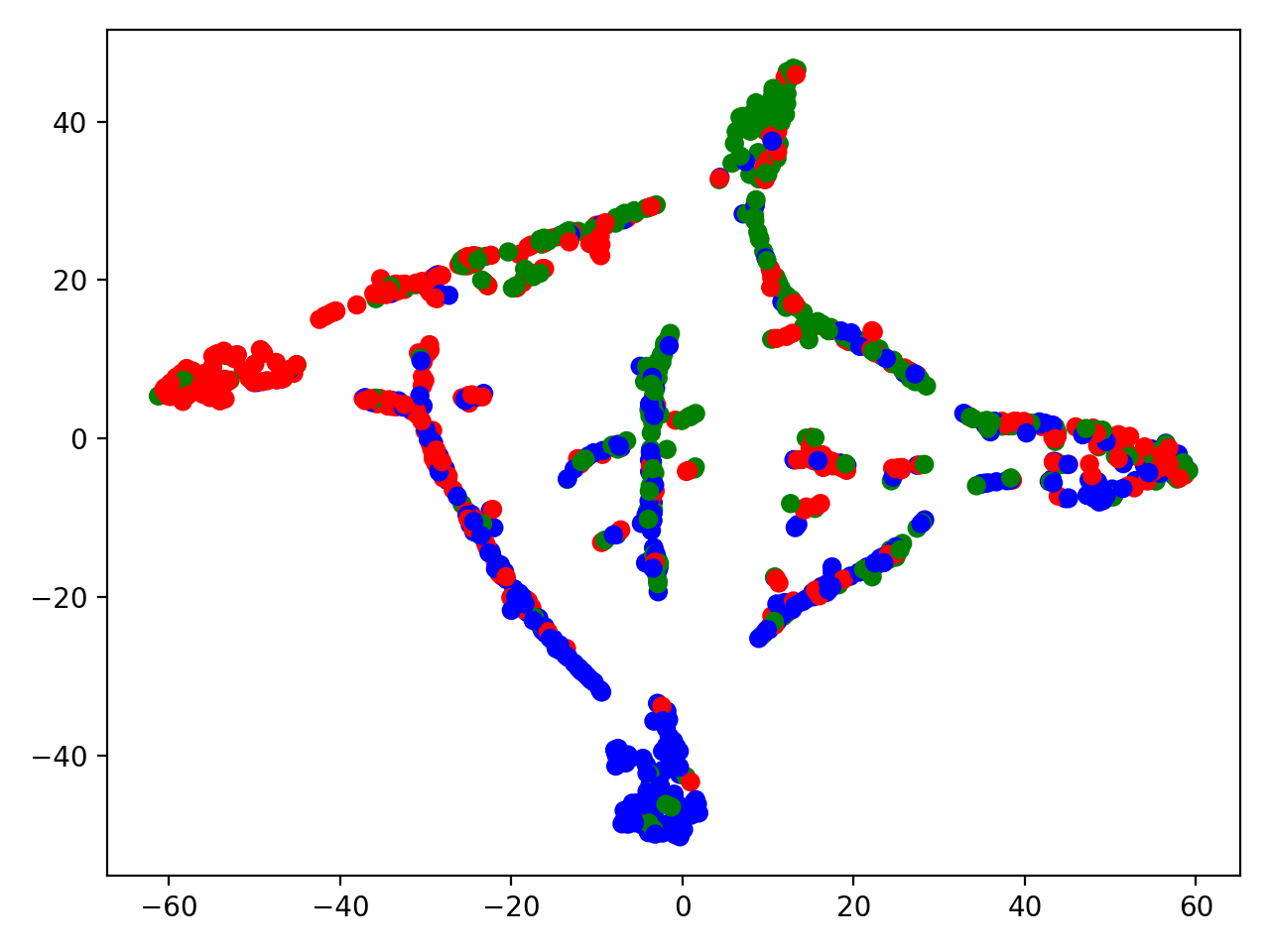}\\[-3pt]
\includegraphics[width=.13\linewidth]{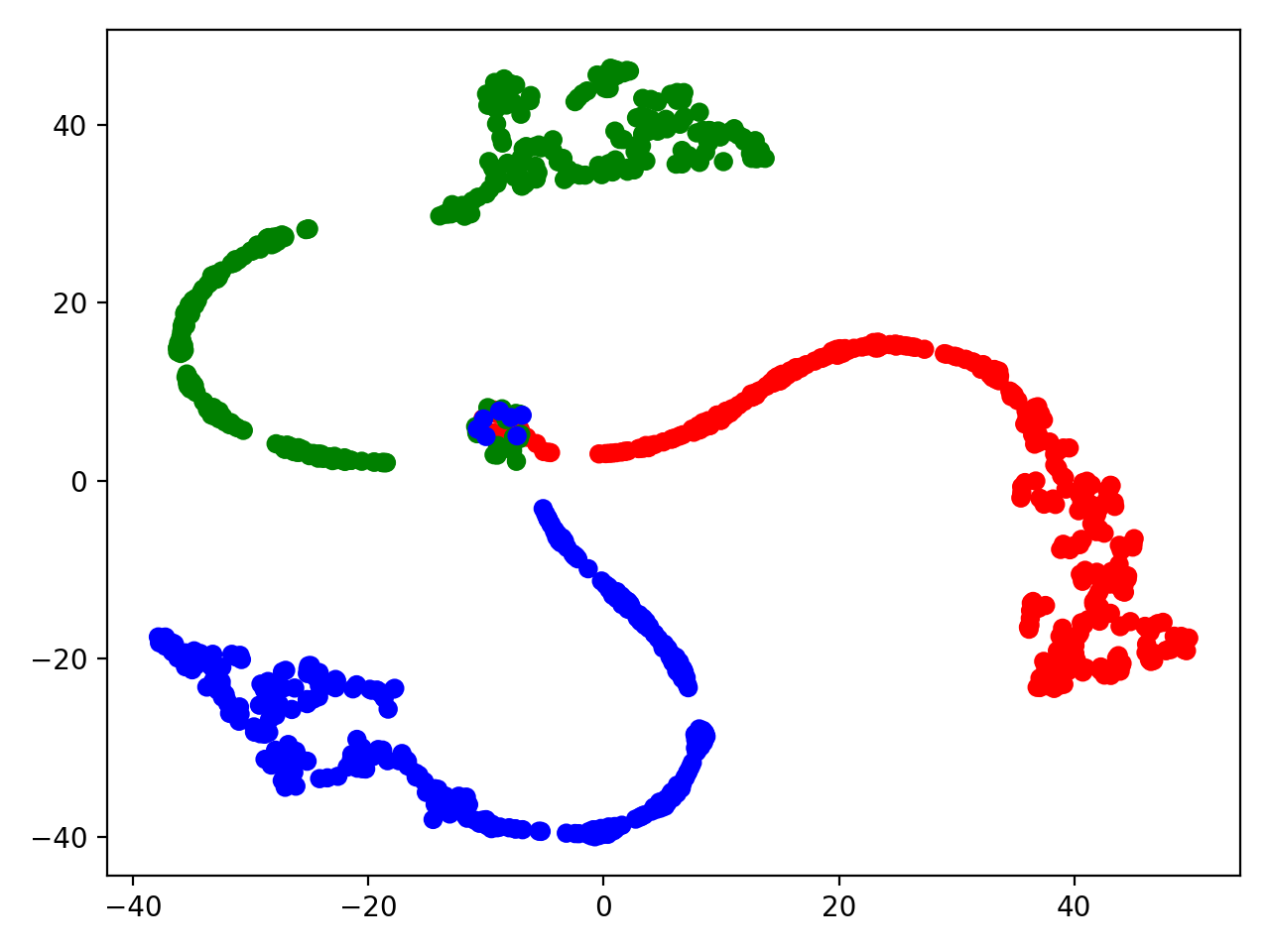}%
\includegraphics[width=.13\linewidth]{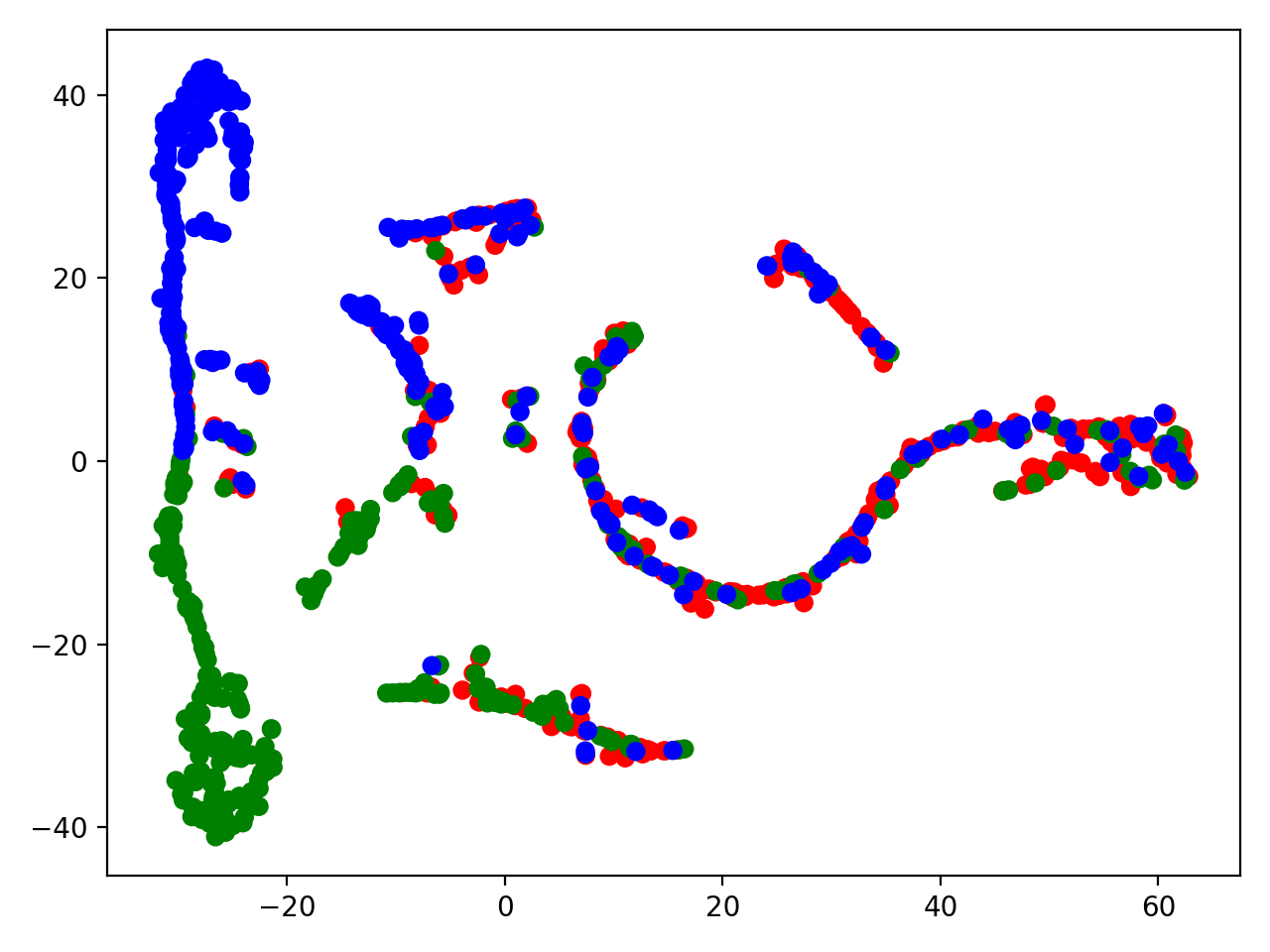}
\end{tabular}%
}
 \caption{Projections of $\hat{\theta}$ learned by LDA, pf-sLDA and SAP-sLDA where documents are colored by label. Each row corresponds to a different setting of $\theta$ -- settings 1 (single-topic documents), 2 (mixed-topic documents) and 3 (predictive-topic documents with garbage words) from top to bottom. Each column corresponds to a different setting of the word-topic distribution $\beta$  -- identifiable and non-identifiable from left to right. SAP-sLDA is able to recover ground truth clusters (matching number of clusters and distribution of documents of each color in each cluster) for more settings of $\theta$ than LDA and pf-sLDA only failing when for setting 3 when $\beta$ is non-identifiable.}
 \label{fig:toy_projection}
\end{figure*}

\begin{figure}[h!]\centering
\subfloat[LDA]{\label{fig.a}
\begin{tabular}[b]{@{}c@{}}
\includegraphics[width=0.49\linewidth]{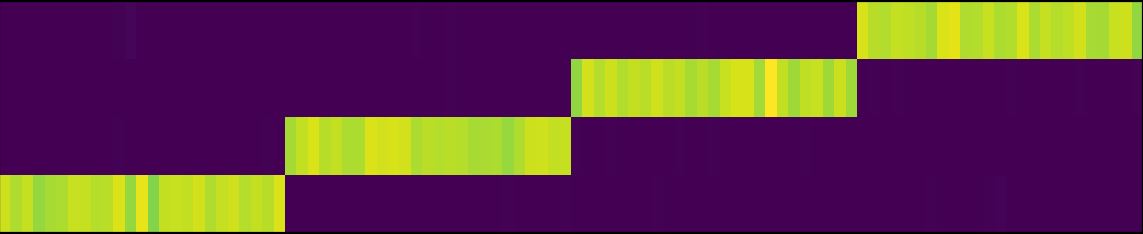}
\includegraphics[width=0.49\linewidth]{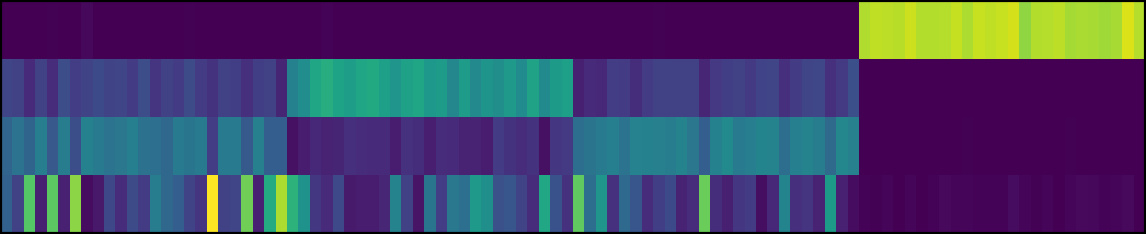}
\end{tabular}
}
\hfill
\subfloat[pf-sLDA]{\label{fig.b}
\begin{tabular}[b]{@{}c@{}}
\includegraphics[width=.49\linewidth]{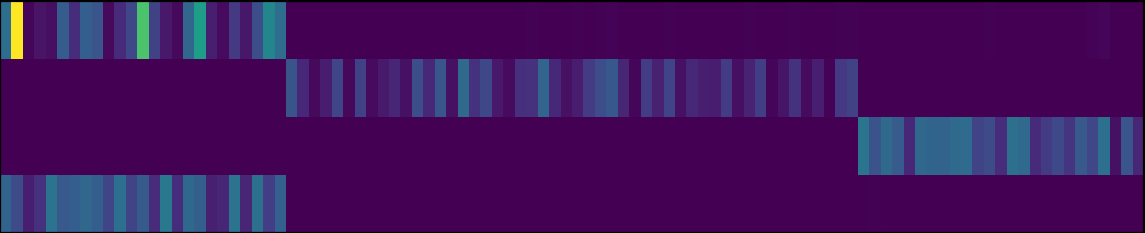}
\includegraphics[width=.49\linewidth]{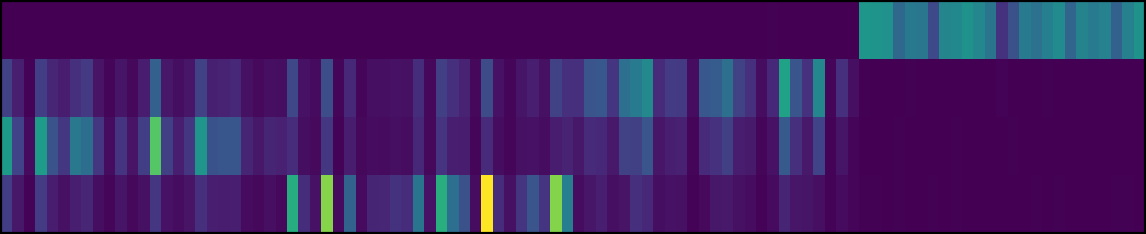}
\end{tabular}
}
\hfill
\subfloat[SAP-sLDA]{\label{fig.c}
\begin{tabular}[b]{@{}c@{}}
\includegraphics[width=.49\linewidth]{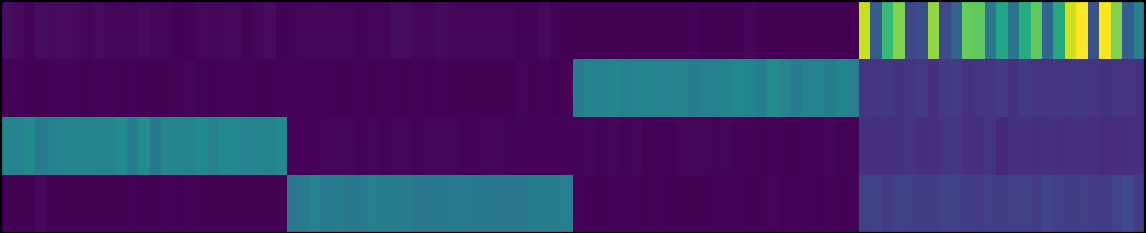}
\includegraphics[width=.49\linewidth]{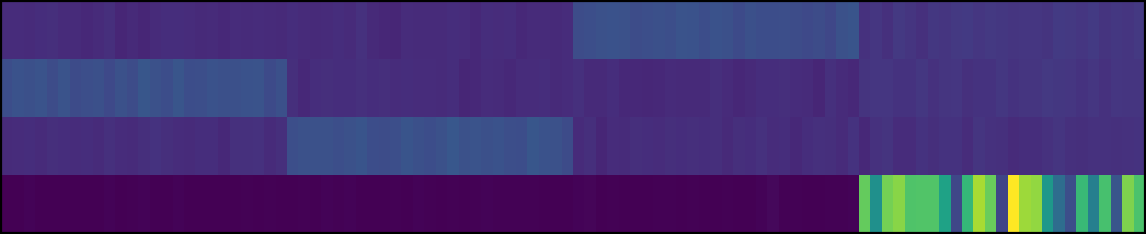}
\end{tabular}%
}
\caption{Learned word-topic distributions $\hat{\beta}$ for setting 1 (single topic documents). In each figure, rows represent topics, and columns represent words. Yellow indicates higher probability mass on the corresponding word, and purple the opposite. Left figures correspond to identifiable $\beta$; right figures correspond to non-identifiable $\beta$. SAP-sLDA better recovers the ground truth $\beta$ than LDA when $\beta$ is non-identifiable} \vspace{-0.1cm}
\label{fig:toy_beta}
\end{figure}

From Figure \ref{fig:toy_projection} we see that \textbf{when $\mathbf{\beta}$ is identifiable, LDA and pf-sLDA are able to recover ground truth clusters}, with pf-sLDA having less projection variance across random restarts than LDA (Figures \ref{fig:lda_restarts} and \ref{fig:pf-slda-restarts} in Appendix A.4). However, \textbf{when $\beta$ is non-identifiable, LDA is unable to separate documents with different labels} for all settings of $\theta$ tested. pf-sLDA is still able to reconstruct red, green, blue and mixed clusters in setting 1 but fails for settings 2 and 3. From Figure \ref{fig:toy_beta} we see that when $\beta$ is identifiable LDA learns $\hat{\beta}$ that exactly matches the ground truth; however, this is not the case when $\beta$ is non-identifiable. pf-sLDA always learns artificially sparse $\hat{\beta}$.

In contrast, \textbf{SAP-sLDA recovers ground truth clusters while preserving local features even for non-identifiable $\beta$}. In Figure \ref{fig:toy_projection} we see that we see that in 5 out of 6 experiments SAP-sLDA produces three distinct red, green and blue clusters. Notably, despite our regularization term penalizing mixed-label clusters we still recover the large mixed-label cluster in setting 1 as is present in the ground truth data. When SAP-sLDA recovers $\hat{\theta}$ with a reasonably similar projection to $\theta$, regardless of identifiability, the learned $\hat{\beta}$ is very similar to $\beta$ (Figure \ref{fig:toy_beta}). The stability of SAP-sLDA across random restarts is superior to that of pf-sLDA (Figure \ref{fig:sap-slda-restarts} in Appendix A.4).

\begin{figure}[h!]
 \centering
 \begin{subfigure}[b]{0.23\textwidth}
     \centering
     \includegraphics[width=\textwidth]{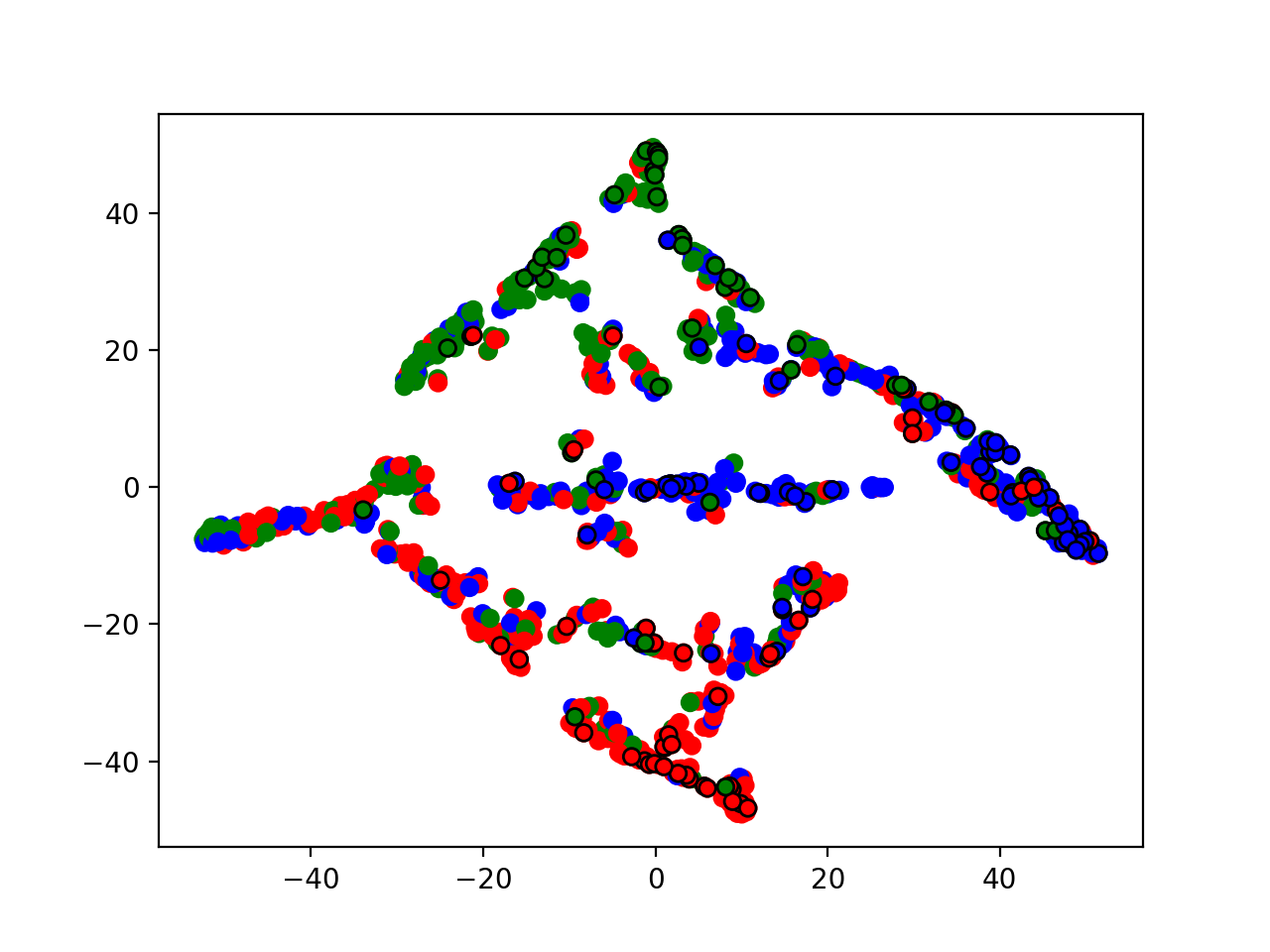}
 \end{subfigure}
 \hfill
 \begin{subfigure}[b]{0.23\textwidth}
     \centering
     \includegraphics[width=\textwidth]{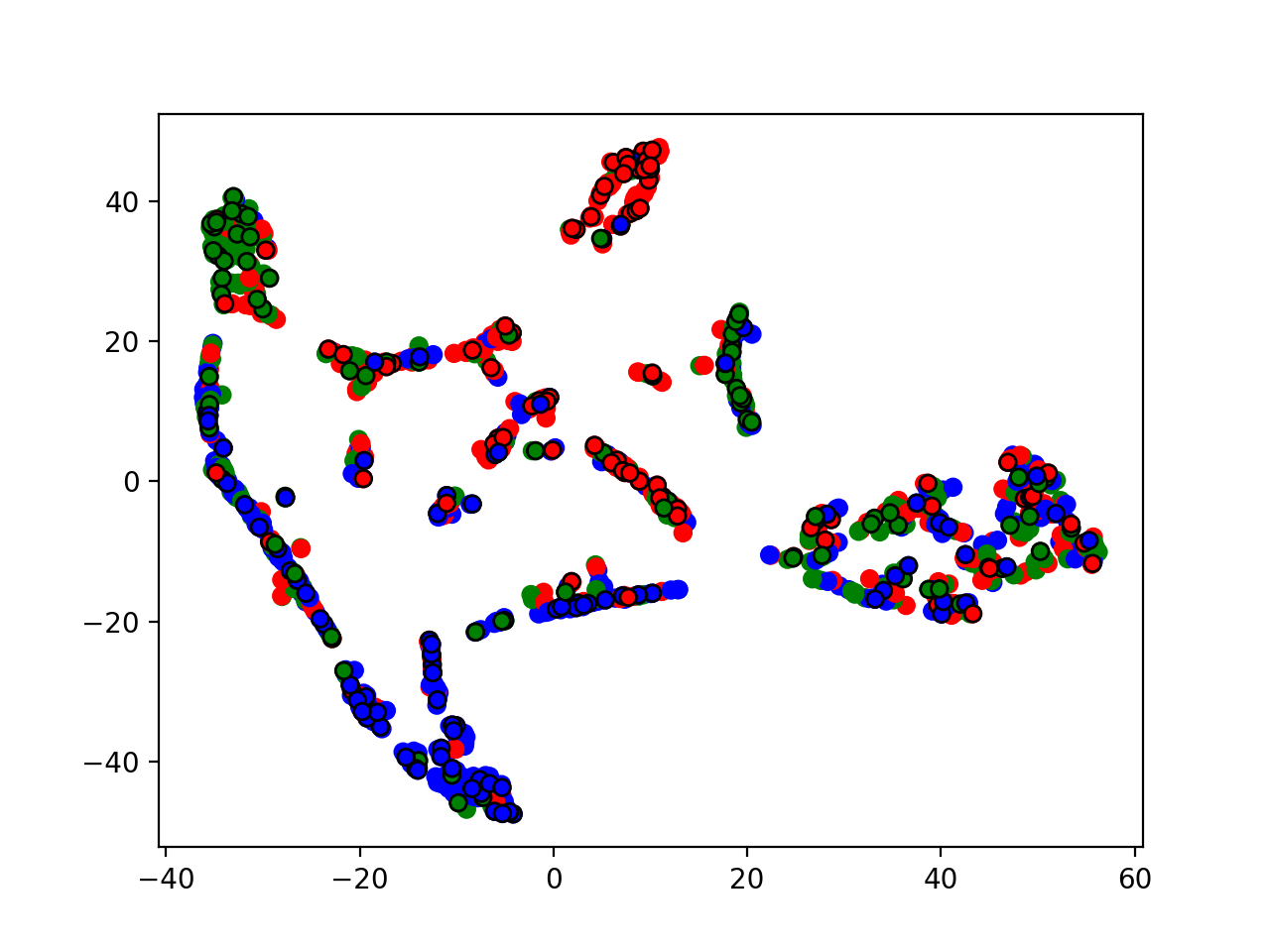}
 \end{subfigure}
 \label{active-learning}
    \caption{Projected $\hat{\theta}$ learned using SAP-sLDA in setting 2 (mixed-topic documents), non-identifiable word-topic distribution $\beta$. Variance-based active labelling separates the red, green and blue clusters with 15\% of the corpus labelled (left) whereas 25\% of documents must be labelled to achieve this with random labelling (right). Documents are outlined if label is provided.}
    \label{fig:active-learning}
\end{figure}

\textbf{Variance-based active labelling beats random labelling on toy data}. From Figure \ref{fig:active-learning} we see that the former is able to separate distinct red, green and blue clusters with only 15\% of the labels provided, while the latter requires 25\% of the corpus to be labelled to achieve this. However, we note that while random labelling produces relatively stable projections once 50\% of the corpus is labelled, variance-based labelling does not (Figures \ref{fig:random-active}, \ref{fig:variance-active} in Appendix A.5).


\begin{figure}[h!]\centering
\begin{tabular}[b]{@{}c@{}}
\includegraphics[width=.99\linewidth]{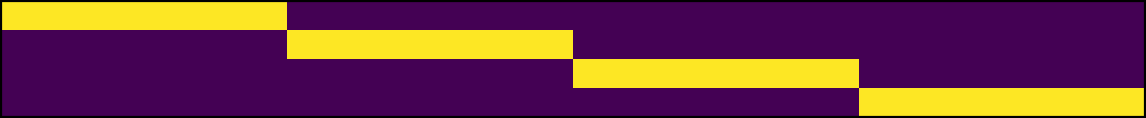}\\[3pt]
\includegraphics[width=.99\linewidth]{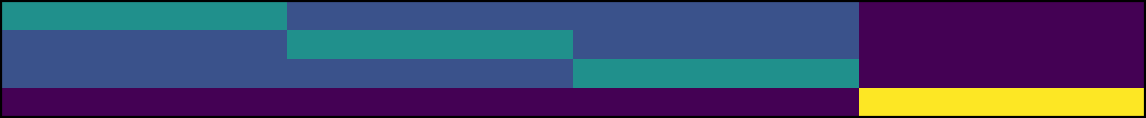}
\end{tabular}
\caption{Two settings of ground truth word-topic distributions $\beta$. In each figure, rows represent topics and columns represent words. Yellow indicates higher probability mass on the corresponding word, purple indicates lower. Top image is when $\beta$ is identifiable, bottom is when $\beta$ is not. } \vspace{-0.1cm}
\end{figure}

\subsection{Dharma Seed Corpus}

Since the Dharma Seed corpus is a collection of audio works, we apply a pre-processing pipeline to transcribe and prepare it for topic modelling which includes breaking documents into small documents of similar length (see Appendix A.2). There are 11047 documents post-processing which have a mean length of 1006 words with a standard deviation of 242 words. For the purposes of exploration and comparison with SAP-sLDA and pf-sLDA which require a certain percentage of the dataset to be labelled, we select a simple random sample of 200 documents from the larger corpus to run experiments on. It is possible that the same contained multiple documents chunked from the same original documents.
We answer the following questions:
\begin{enumerate}
    \item How sensitive is SAP-sLDA to the \textit{type} of label given?
    \item How well does SAP-sLDA satisfy the desired interpretability criteria as compared to LDA and pf-sLDA when 50\% of documents are labelled?
\end{enumerate}
To answer these questions we run SAP-sLDA with 10 topics and with $\lambda_1 = 1$, $\lambda_3 = 0.1$ and $\lambda_2 = \lambda_4 = 4$.
We only run a single iteration of SAP-sLDA providing all known labels at this iteration. We try using labels from each of the three label classes discussed in section 4.3 (labelling documents randomly, labelling by author, labelling by broad theme). For this iteration of SAP-sLDA, we train the objective over 200 iterations.
We compare the performance of SAP-sLDA to regular LDA and pf-sLDA run with 10 topics for 200 iterations. We set $p = 0.25$ for pf-sLDA. All hyperparameters were set to reasonable values for the corpus at hand (ELBO stopped increasing significantly after 200 iterations), but were not tuned. For all experiments we create projections using the TSNE class from \texttt{sklearn.manifold} with perplexity 20.

\subsection{Results}
\begin{figure}[h!]\centering
\subfloat[LDA]{\label{fig.a}
\begin{tabular}[b]{@{}c@{}}
\includegraphics[width=.49\linewidth]{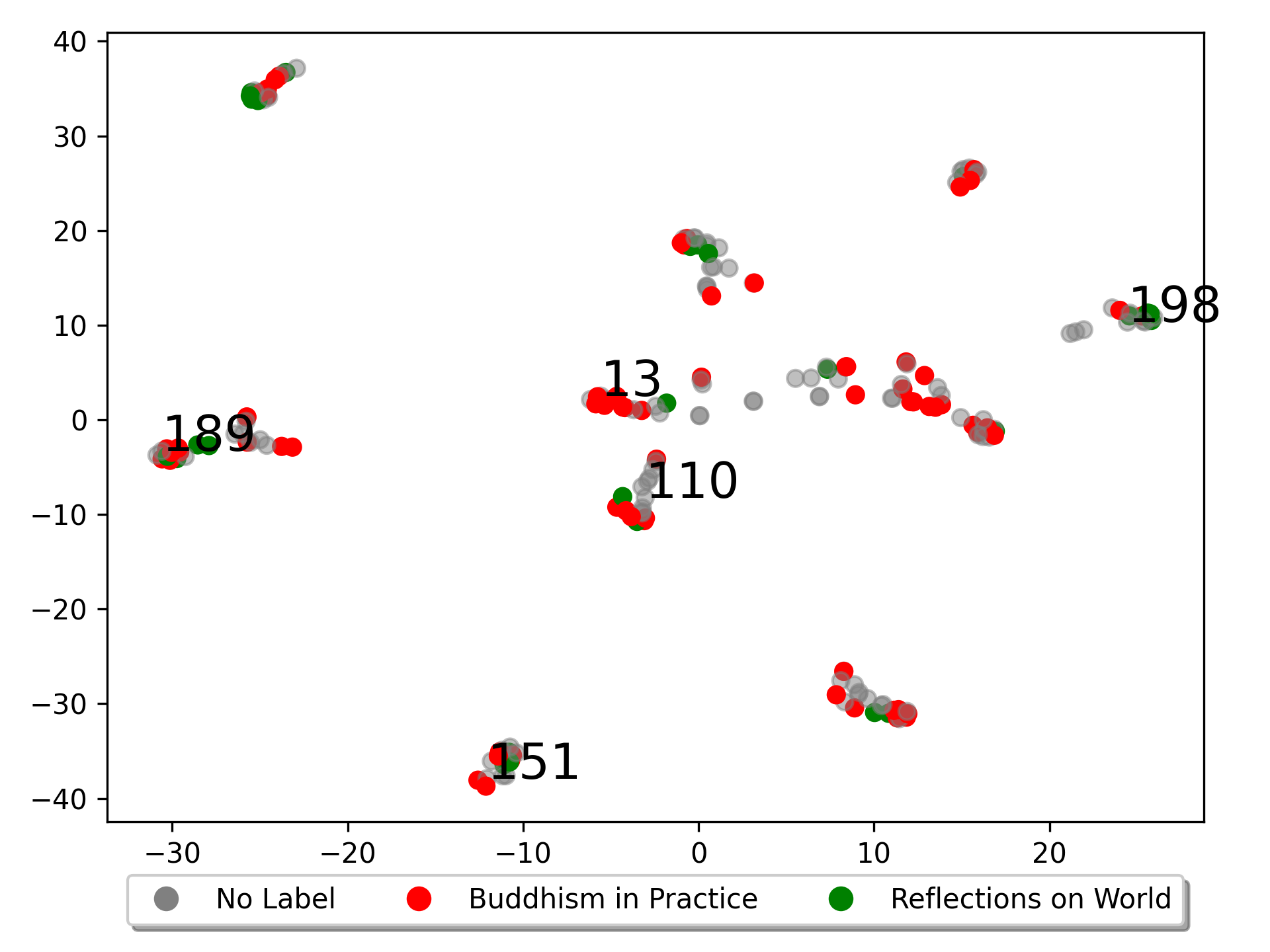}
\includegraphics[width=.49\linewidth]{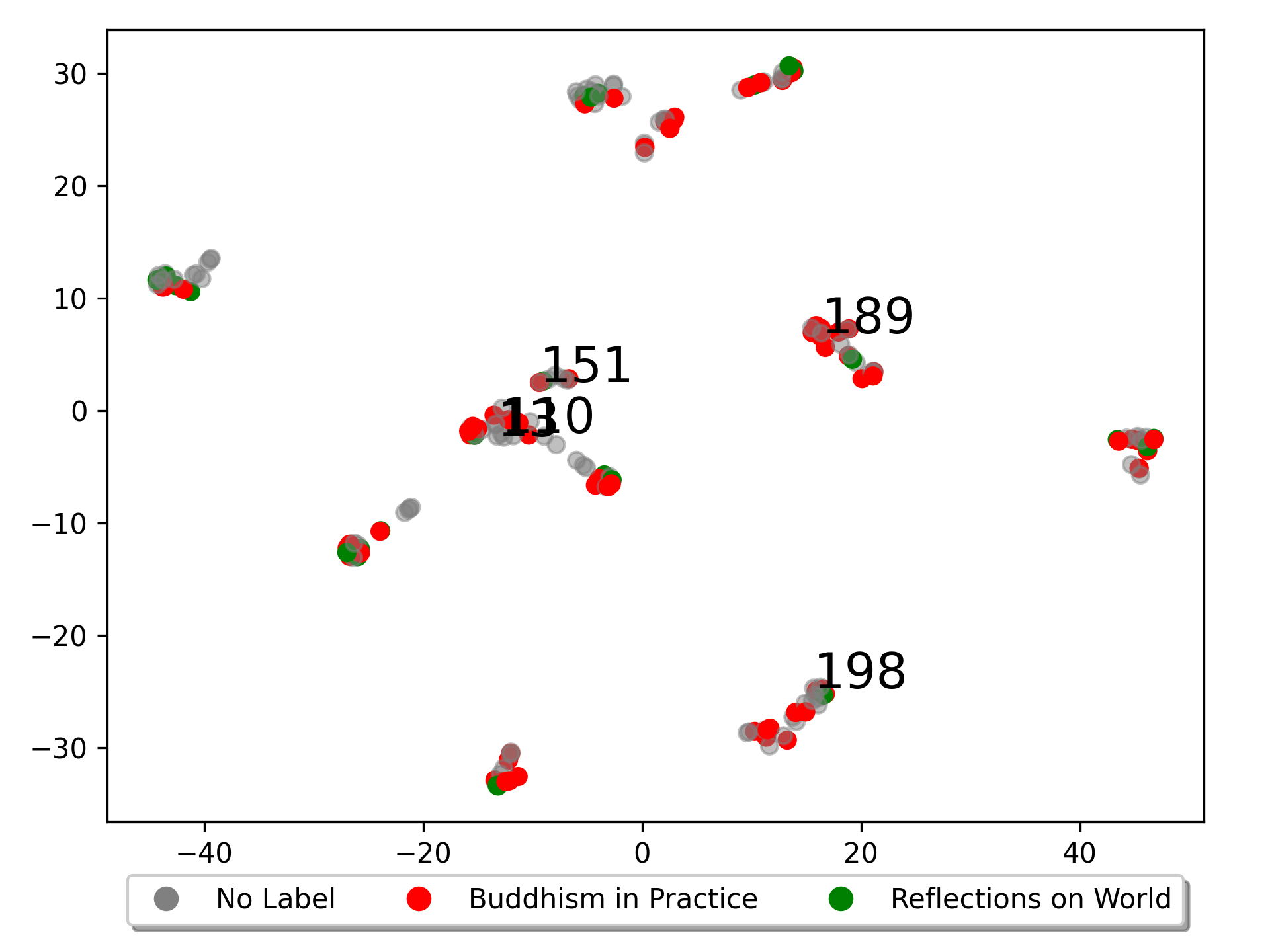}
\end{tabular}
}
\hfill
\subfloat[pf-sLDA]{\label{fig.b}
\begin{tabular}[b]{@{}c@{}}
\includegraphics[width=.49\linewidth]{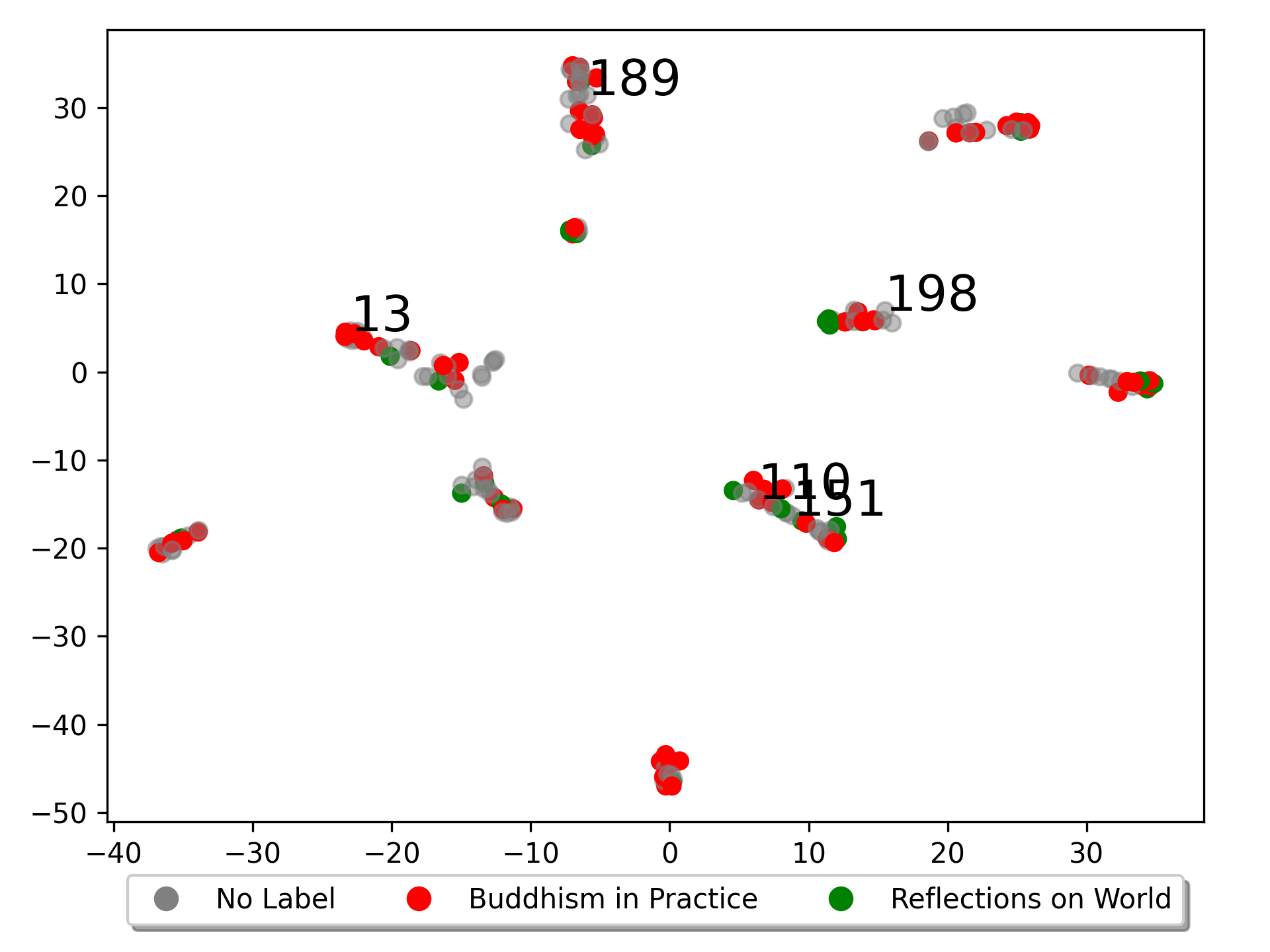}
\includegraphics[width=.49\linewidth]{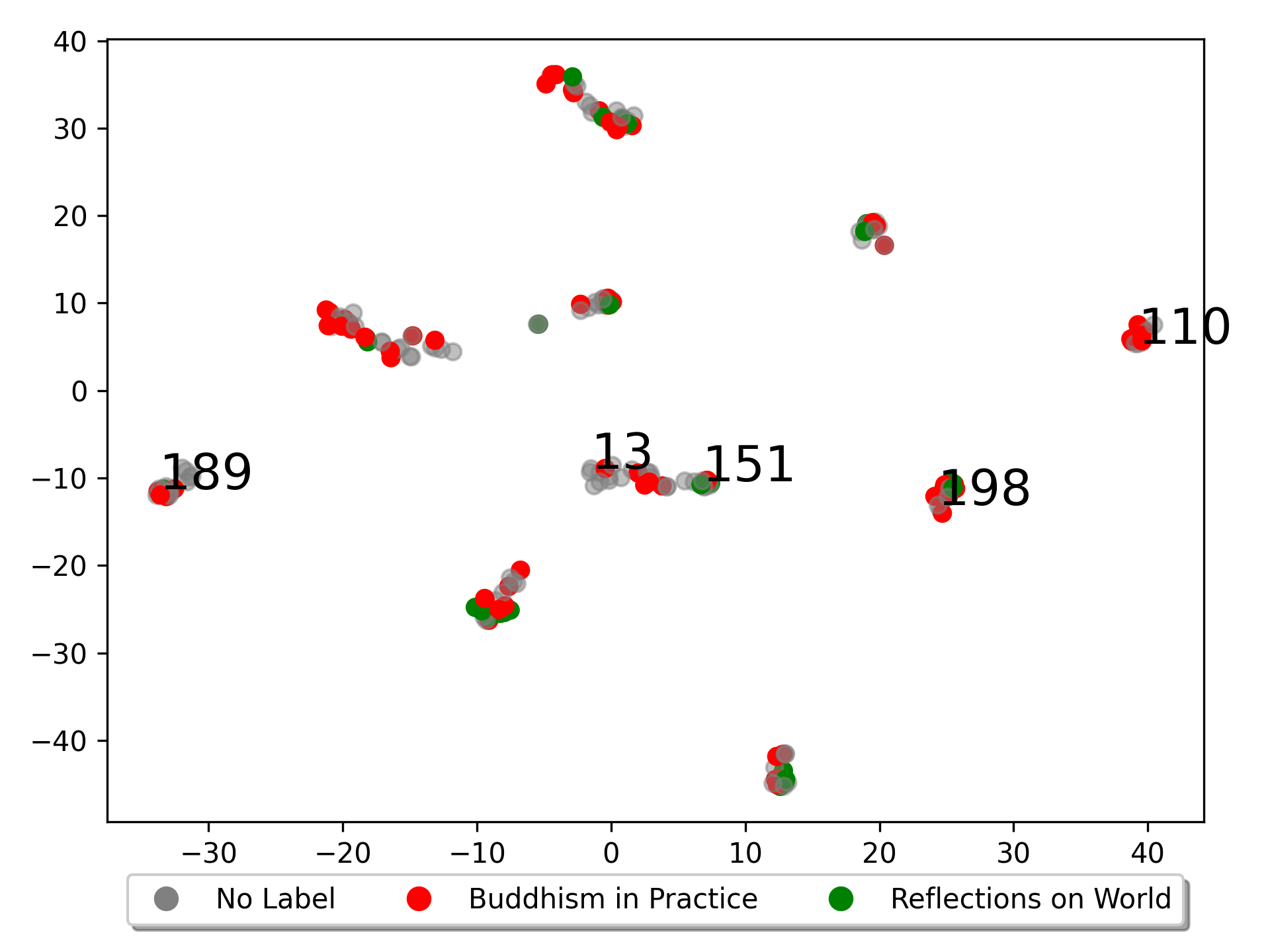}
\end{tabular}
}
\hfill
\subfloat[SAP-sLDA]{\label{fig.c}
\begin{tabular}[b]{@{}c@{}}
\includegraphics[width=.49\linewidth]{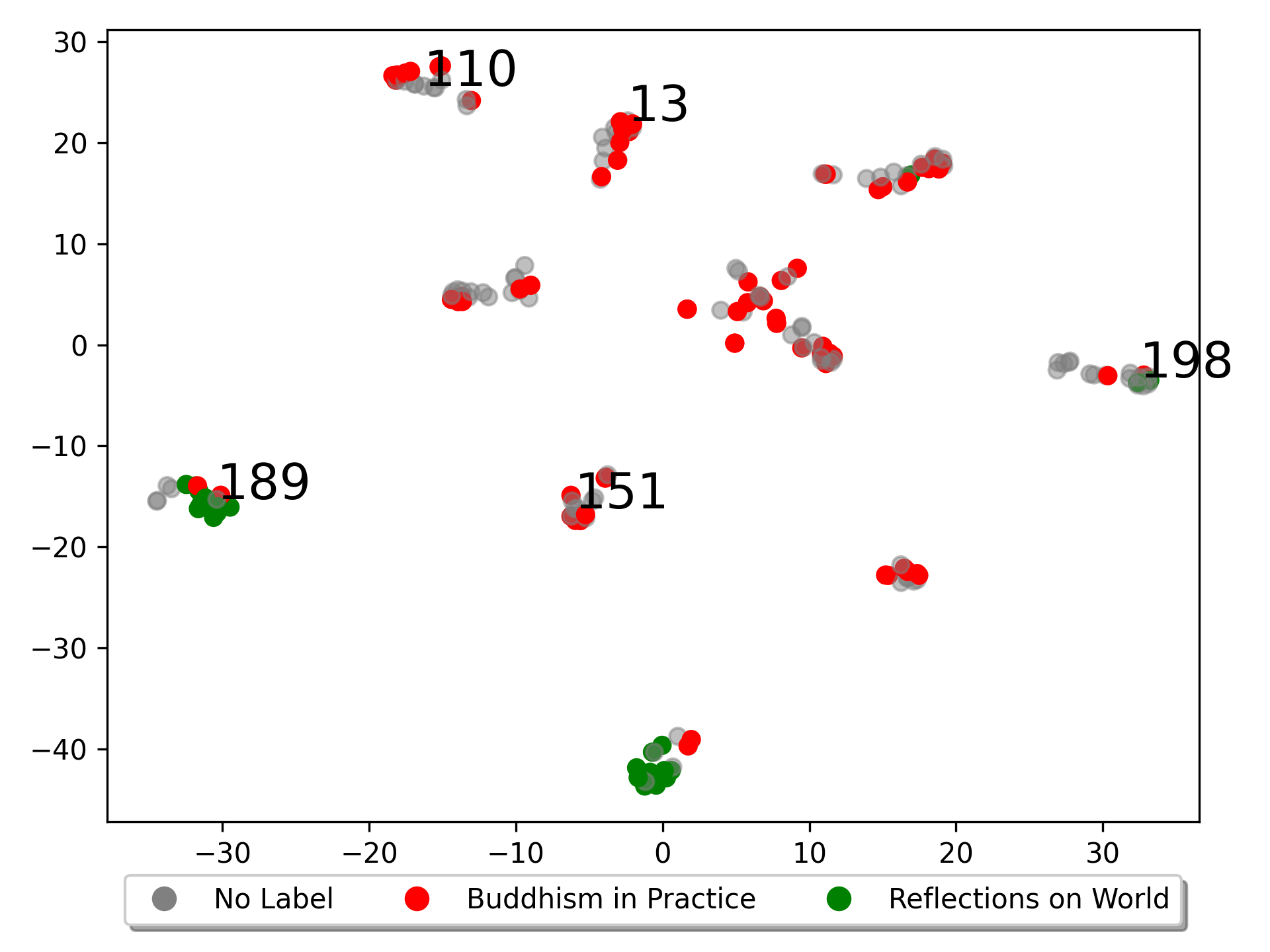}
\includegraphics[width=.49\linewidth]{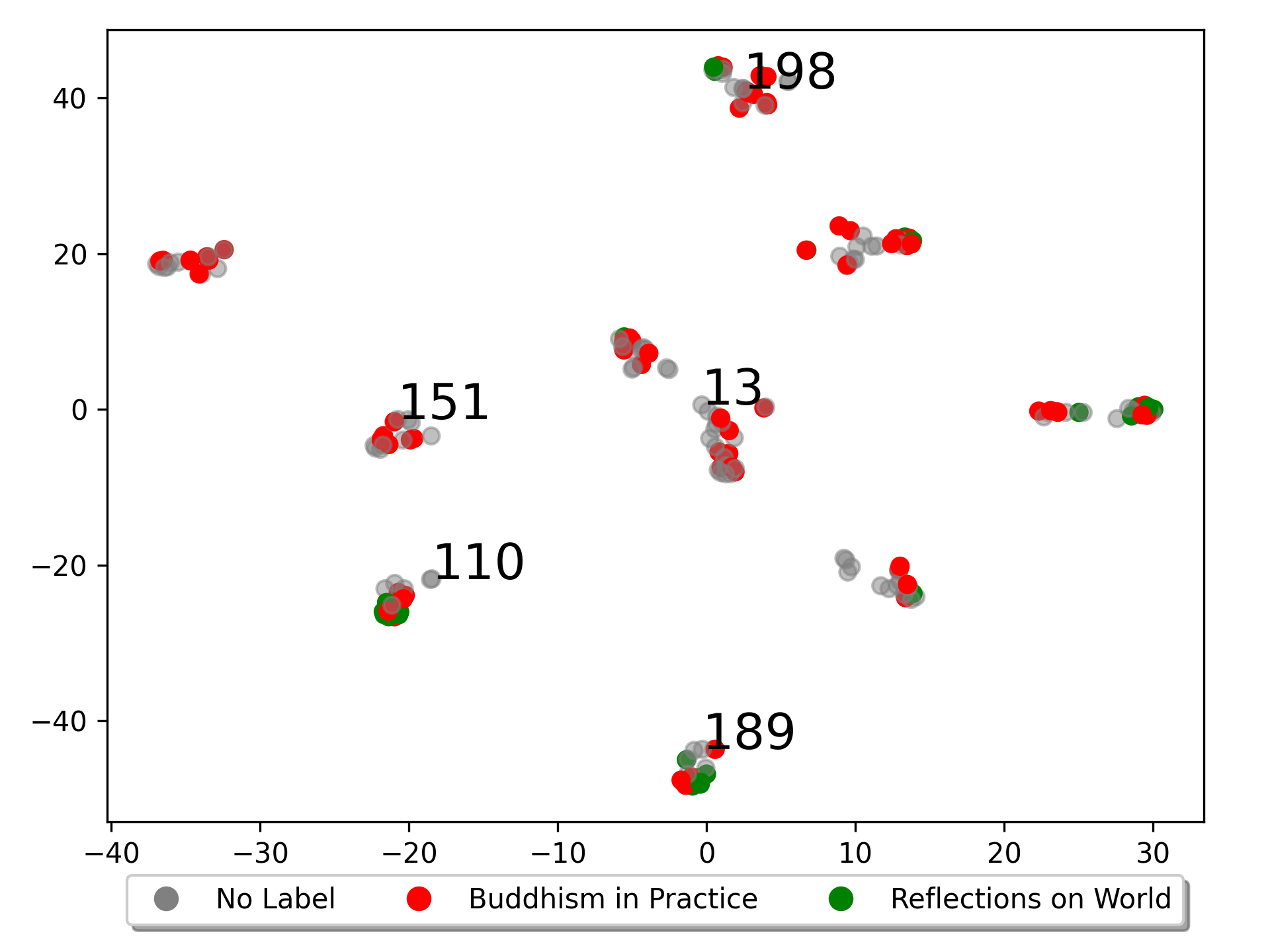}
\end{tabular}%
}
\caption{Projections of $\hat{\theta}$ learned by LDA, pf-sLDA, and SAP-sLDA when 50\% of documents were labelled by theme across two random restarts. Documents are colored by label, with unlabelled documents colored in grey. We see that SAP-sLDA yields projections which are purer and more stable with respect to the relative positions of 5 unlabelled documents (marked in the figures) than LDA and pf-sLDA.}
 \label{fig:theme-labelling}
\end{figure}

On the Dharma Seed corpus, we find that \textbf{clustering quality is highly dependent on signal provided by the labels}. Both random labelling and labelling by author yield three mixed-label clusters that vary from run to run (Figure \ref{fig:naive-label} in Appendix A.6). Labelling 50\% of the corpus by theme, which has slightly higher signal, yields ``purer" clusters (with respect to clusters having predominantly documents of the same color), aligning with human intuition that similarly themed documents should be nearby in the projected space. We note that by setting the regularization strength controlled by $\lambda_{1\cdots 4}$ high enough, it is possible to get pure clusters regardless of label class, however, this yields a clustering that is useless to humans. The fact that pure clusters were achievable using a relatively low-strength regularizer suggests that higher-signal labels are more synergistic with the original LDA ELBO guiding the optimizer towards a better optimum.
Tracking the position of five randomly selected unlabelled documents across two random restarts shows reasonable preservation of their relative positions in the projected space. This does \textit{not} hold for pf-sLDA and LDA both of which yield more mixed-label clusters and less stable projections (Figure \ref{fig:theme-labelling}).

\section{Discussion}
We demonstrate the potential of SAP-sLDA, for creating interpretable 2-dimensional projections of unstructured text data, facilitating exploration. Serving as a proof-of-concept, these results suggest several avenues for future research.

First, the success of SAP-sLDA relies on the labels providing sufficient signal about $\theta$. While the Dharma Seed corpus labels used in this paper were determined naively, future work could involve humans ``tagging" documents with relevant keywords. Furthermore, although our visualizations are semantically meaningful, it is unclear whether they are useful for exploration. This could be addressed through human studies. Investigating different active learning labeling schemes, such as labeling documents in clusters with the fewest labeled points, would also be valuable.

Lastly, we note that our regularizer's hyperparameters are not extensively tuned. Performing a hyperparameter sweep and adding additional components to the regularizer (e.g., enforcing sparsity or orthogonality of topics) could further improve the interpretability of the final projection.

\section{Conclusion}
We introduced Semantically-Aligned-Projection (focused) supervised LDA (SAP-sLDA) whose novel regularizer directly enforces interpretable projections of corpora in which clusters capture human notions of document similarity. On synthetic data, we show that SAP-sLDA is able to recover properties of the ground-truth data generating model and requires only a small number of labels to do so. On the target real corpus (Dharma Seed), we show similarly promising results in preliminary experiments. The flexibility of the various components of SAP-sLDA (e.g. the labelling strategy) makes it \textit{customizeable} to specific corpora based on domain knowledge, and invites future work. 

\section*{Acknowledgements}
This material is based upon work supported by the National Science Foundation under Grant No. IIS-1750358.  Any opinions, findings, and conclusions or recommendations expressed in this material are those of the author(s) and do not necessarily reflect the views of the National Science Foundation. A grant from Harvard College URAF's conference funding program is supporting CB's travel costs for attending ICML. We would like to acknowledge the Cambridge Insight Meditation Center and Leandra Tejedor for conducting initial work on the project including dataset collection and exploring data preprocessing techniques.



\nocite{dharmaseed}
\bibliography{main}
\bibliographystyle{icml2023}

\newpage
\appendix
\onecolumn
\section{Appendix}
\subsection{LDA ELBO}
The evidence lower bound (ELBO) that LDA aims to maximize is as follows:
\begin{multline}
    \sum_{k=1}^K\mathbb{E}[\log p(\beta_k)] + \sum_{d = 1}^D \mathbb{E}[\log p(\theta_d|\alpha)] + 
    \sum_{d=1}^D\sum_{n=1}^N \mathbb{E}[\log p(z_{d, n} | \theta_d)] +
    \sum_{d=1}^D\sum_{n=1}^N \mathbb{E}[\log p(w_{d, n}|z_{d, n}, \beta_{1\cdots K})] + H(q)
\end{multline}

where $K$ is the number of topics, $D$ is the number of documents, and $N$ is the number of words in a document. $\beta_k$ is the probability distribution over words for topic $k$ and $\theta_d$ is the probability distribution over topics for document $d$. $z_{d, n}$ is the latent topic of word $w_{d, n}$. $H$ denotes the entropy and $q$ is the variational distribution. All expectations are taken over $q$.

\subsection{Pre-processing Dharma Seed Corpus}
Recall that the Dharma Seed corpus is originally a collection of \textit{audio} talks. After downloading the talks, we use Welder to transcribe them into text files. Before applying LDA, we convert documents to bag of words (BoW) format using the \texttt{CountVectorizer} class from \texttt{sklearn.feature\_extraction.text} and remove all English stop-words.

\subsection{Labelling by Theme in Dharma Seed Corpus}
One of the labelling schemes considered for experimentation on the Dharma Seed corpus is labelling by broad theme. After inspecting several documents by hand, we decided that most documents tended to fall under one of two main themes: \textit{Buddhism in Practice} and \textit{Reflections on the World}. In order to efficiently classify documents as falling under one of these categories, we queried ChatGPT-3.5 with the prompt below followed by the document text.

\textit{I have passages that are about one of two main themes:\\
1. Buddhism in Practice\\
2. Reflections on the World\\
What theme does the following passage fall under? If it doesn't nicely fall under one of the two themes listed above, please suggest another theme that fits it better.\\ 
...
}

A sample response reads as follows:

\textit{The passage falls under the theme of "Buddhism in Practice." It explores the concept of shaping the mind through intentional attitudes and cultivating qualities such as mindfulness, compassion, and metta (loving-kindness). The example of Sita, the eleven-year-old girl who overcame her fear of dogs by choosing not to be afraid anymore, serves as a teaching on the pliability of the mind and the way our experiences are shaped by our relationship with the world. The passage also mentions the practice of reframing, reflecting, and intentionally looking again with kindness and interest, emphasizing the transformative power of such practices in changing our experiences and how we relate to the world.\\
If you have any more questions or need further clarification, feel free to ask!}

We note that ChatGPT-3.5 never suggested an alternate theme. Labels were assigned according to the model's output.

\subsection{LDA, pf-sLDA, and SAP-sLDA Variance Across Runs}
We run LDA, pf-sLDA, and SAP-sLDA thrice for each set of synthetic data to visually assess how much the projections vary across random restarts.

\begin{figure}[H]\centering
\begin{subfigure}[b]{0.49\textwidth}
\begin{tabular}[b]{@{}c@{}}
\includegraphics[width=.32\linewidth]{figs/toy/setting_1/separable/tsne_LDA_setting_1_separable_True_run_0.png}%
\includegraphics[width=.32\linewidth]{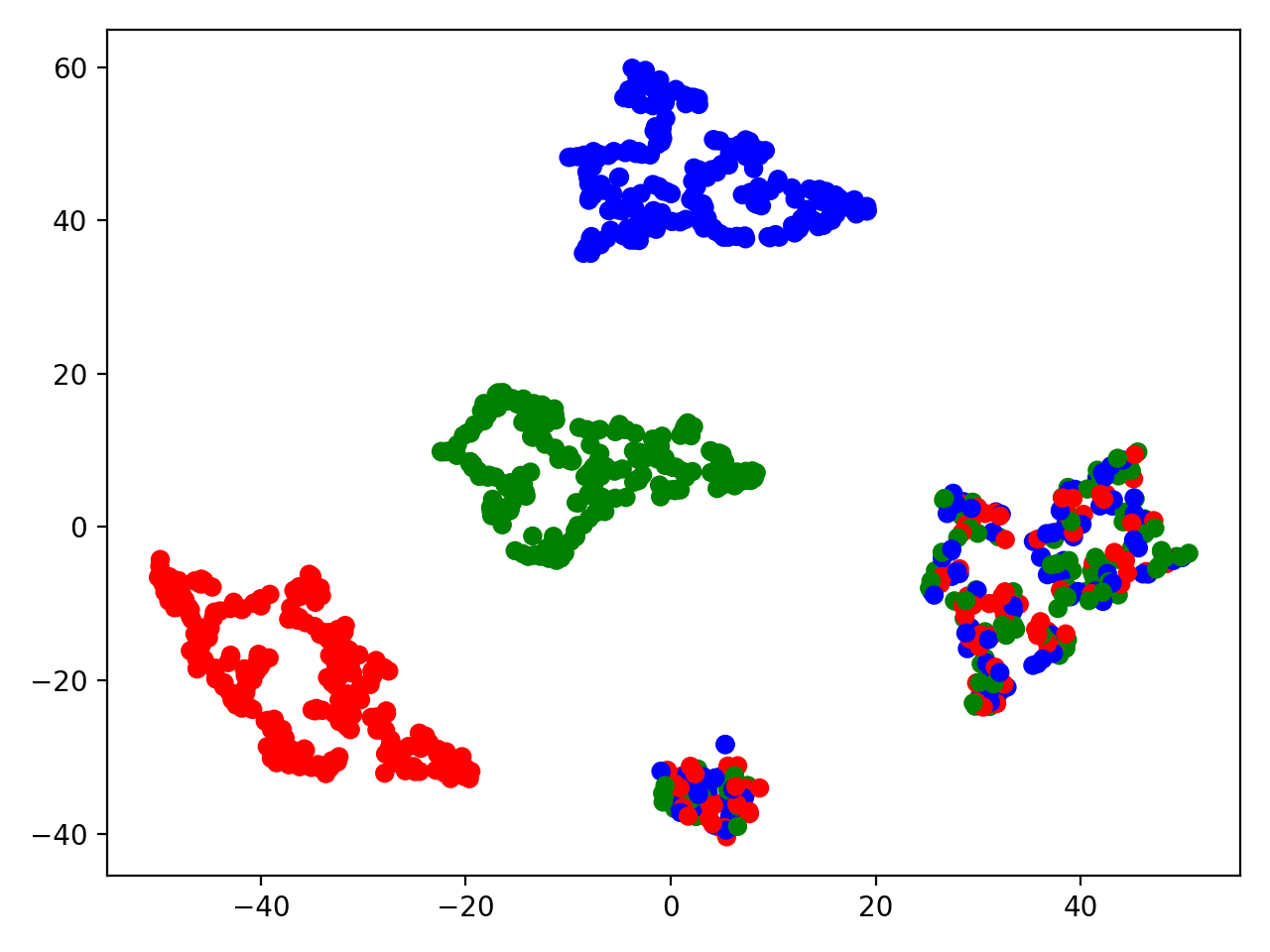}
\includegraphics[width=.32\linewidth]{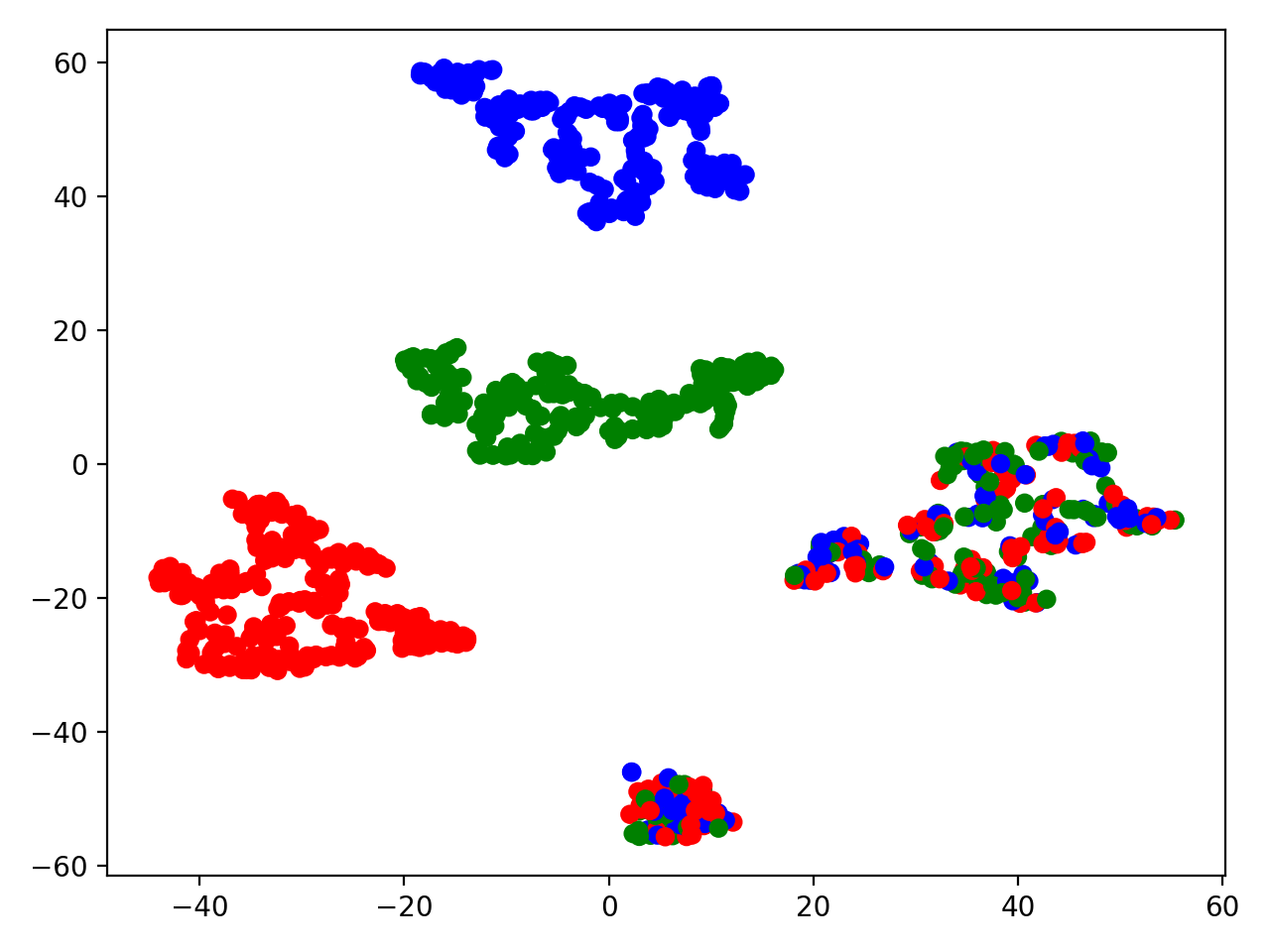}\\
\includegraphics[width=.32\linewidth]{figs/toy/setting_2/separable/tsne_LDA_setting_2_separable_True_run_0.png}%
\includegraphics[width=.32\linewidth]{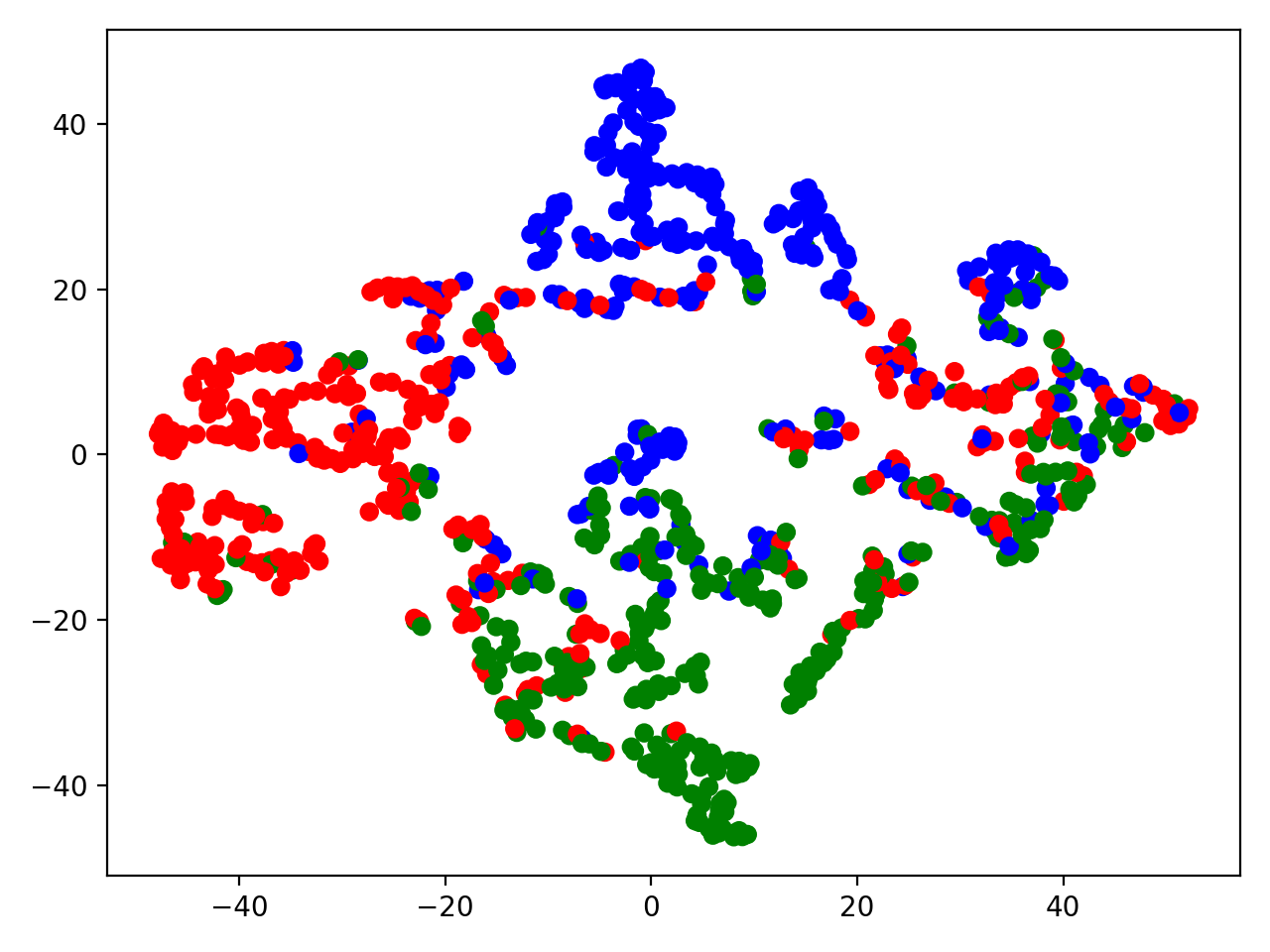}
\includegraphics[width=.32\linewidth]{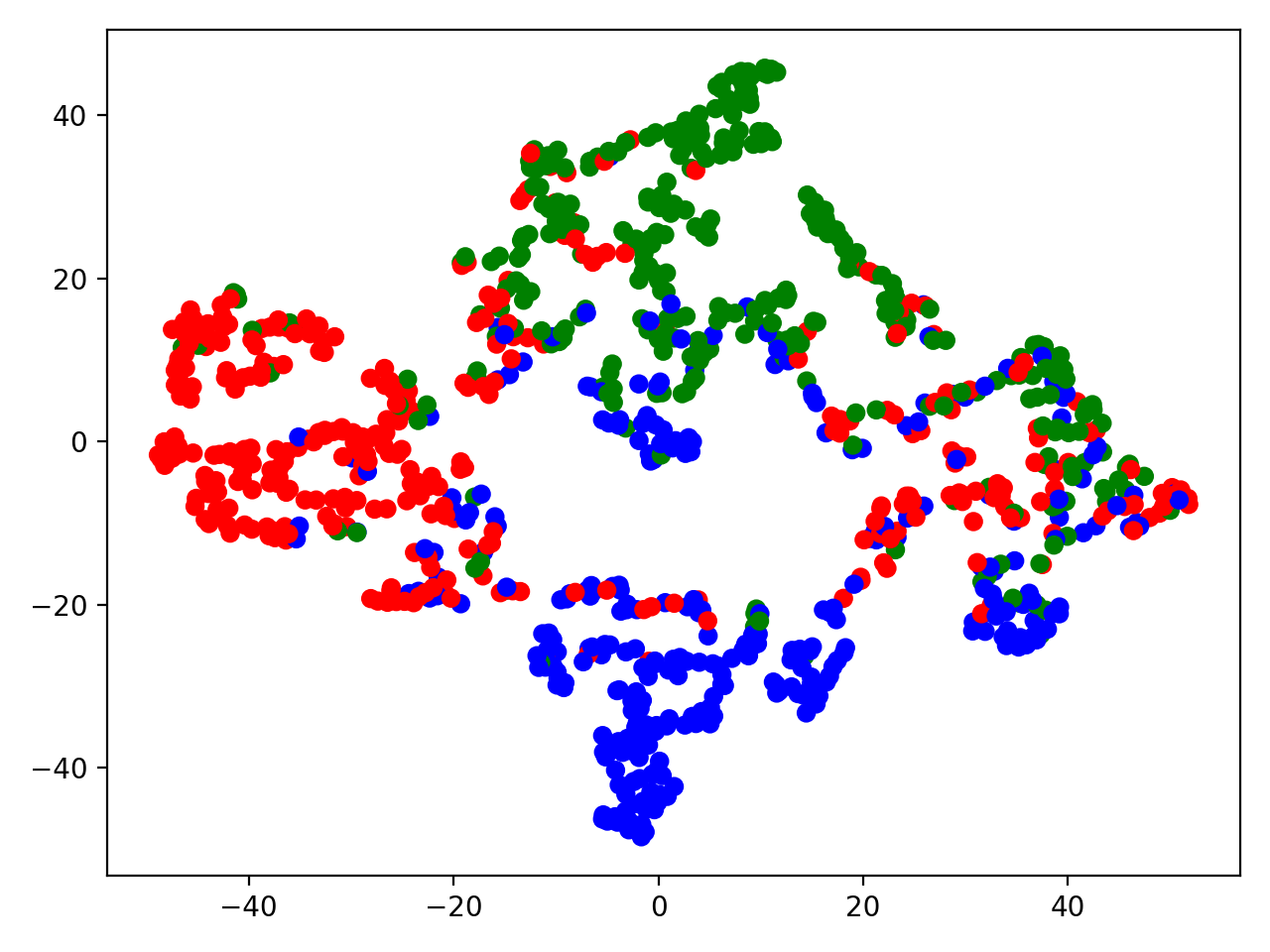}\\
\includegraphics[width=.32\linewidth]{figs/toy/setting_3/separable/tsne_LDA_setting_3_separable_True_run_0.png}%
\includegraphics[width=.32\linewidth]{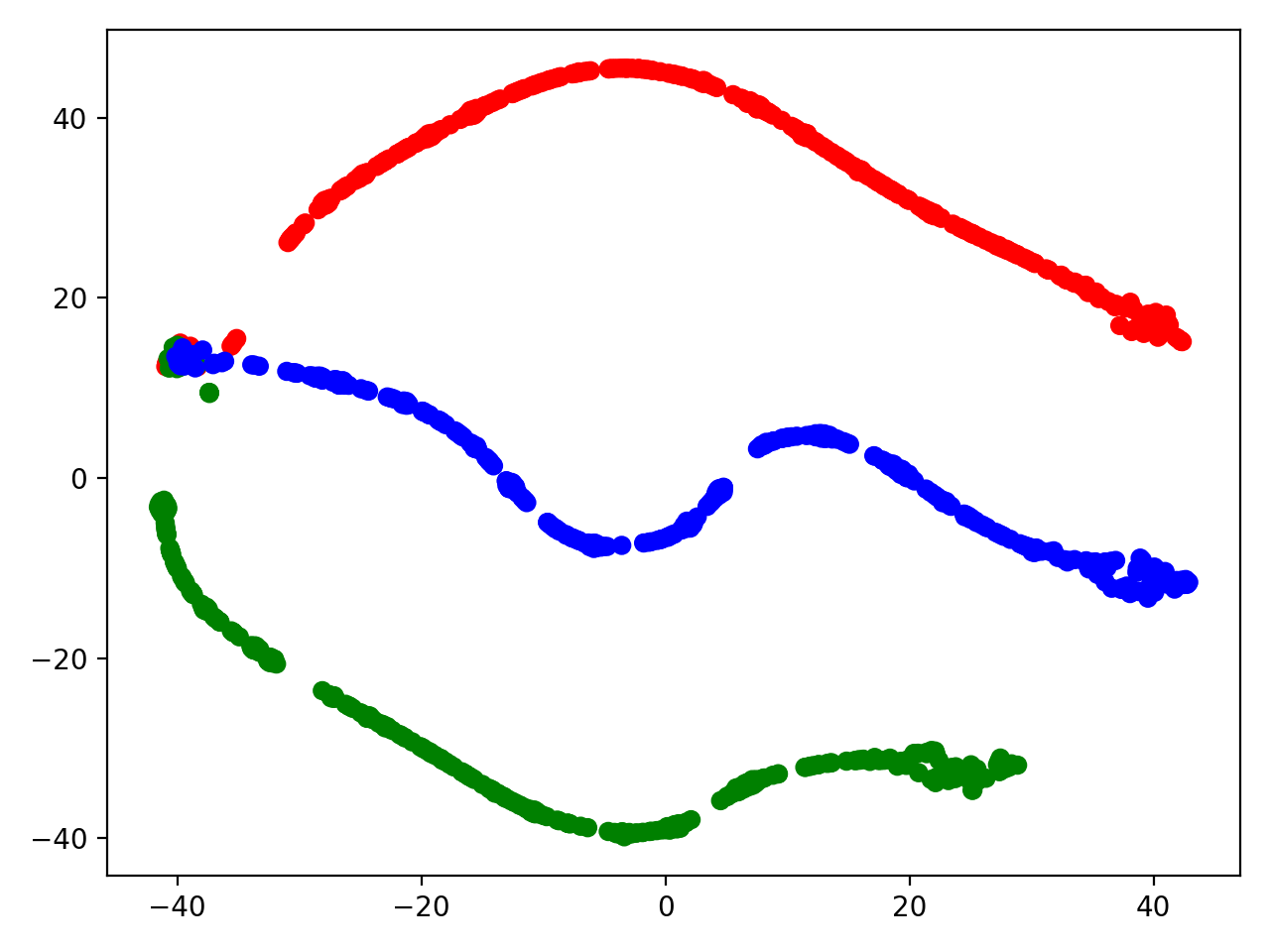}
\includegraphics[width=.32\linewidth]{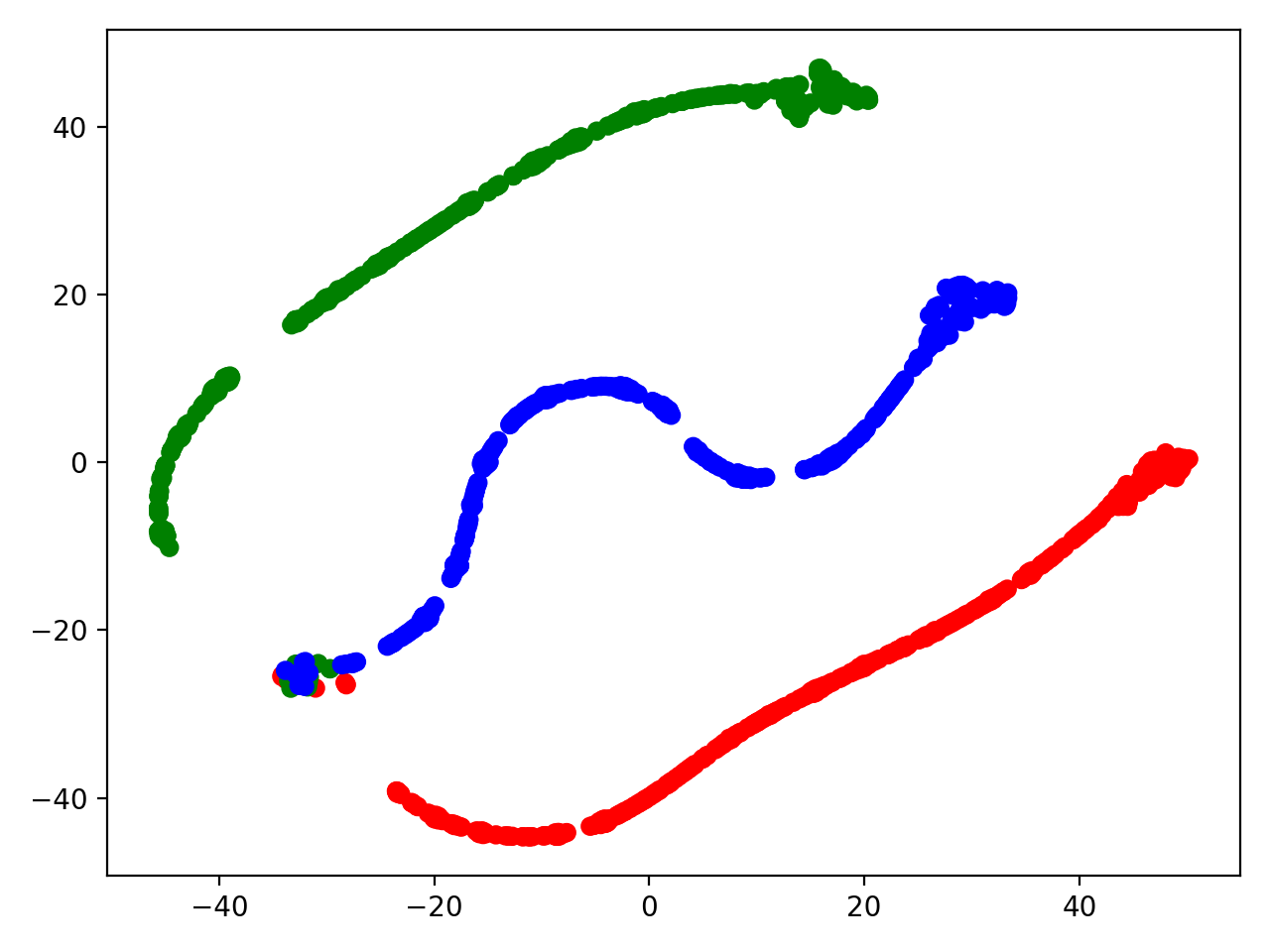}%
\end{tabular}
\end{subfigure}
\hfill
\begin{subfigure}[b]{0.49\textwidth}
\begin{tabular}[b]{@{}c@{}}
\includegraphics[width=.32\linewidth]{figs/toy/setting_1/not_separable/tsne_LDA_setting_1_separable_False_run_0.png}%
\includegraphics[width=.32\linewidth]{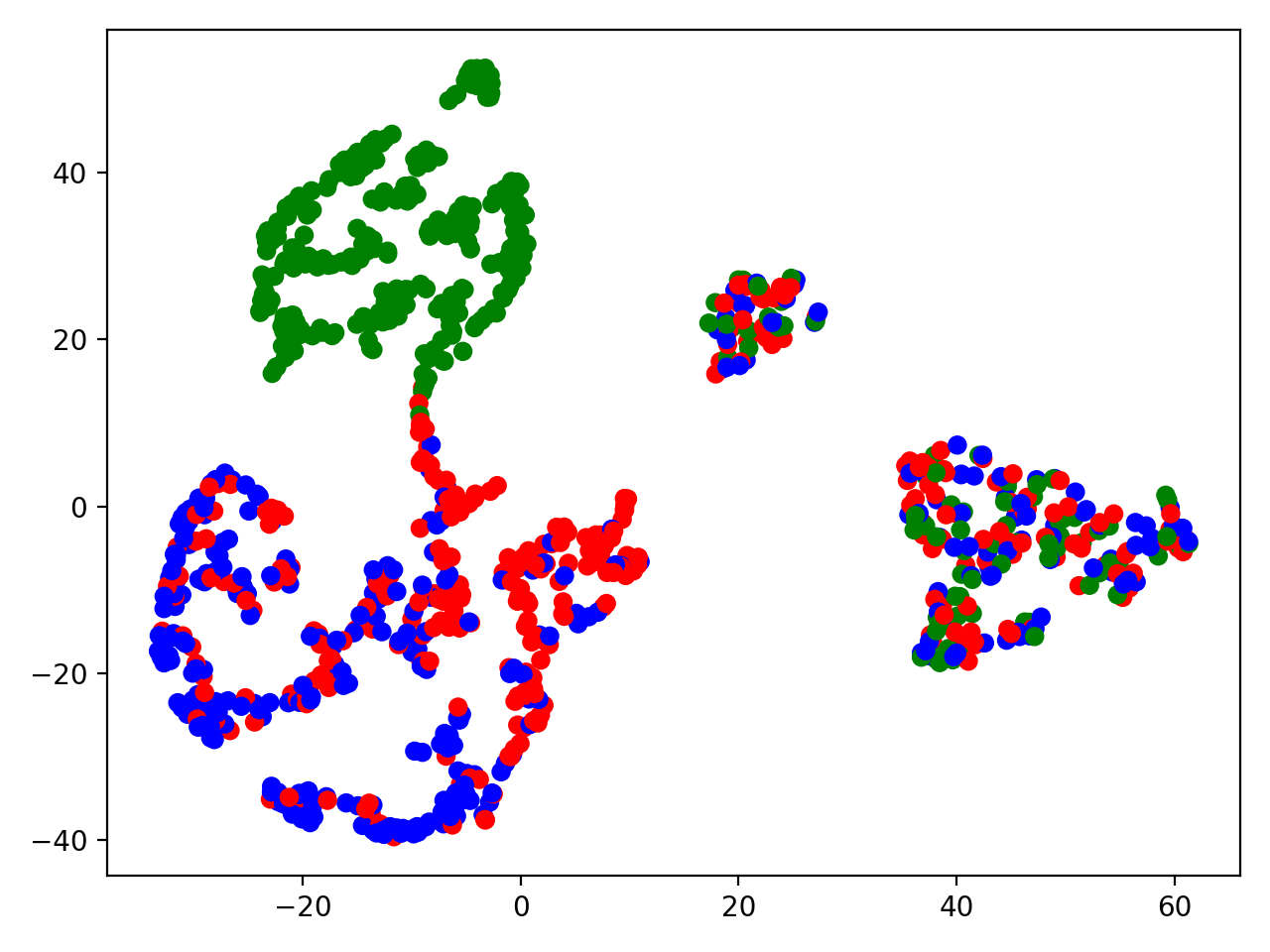}
\includegraphics[width=.32\linewidth]{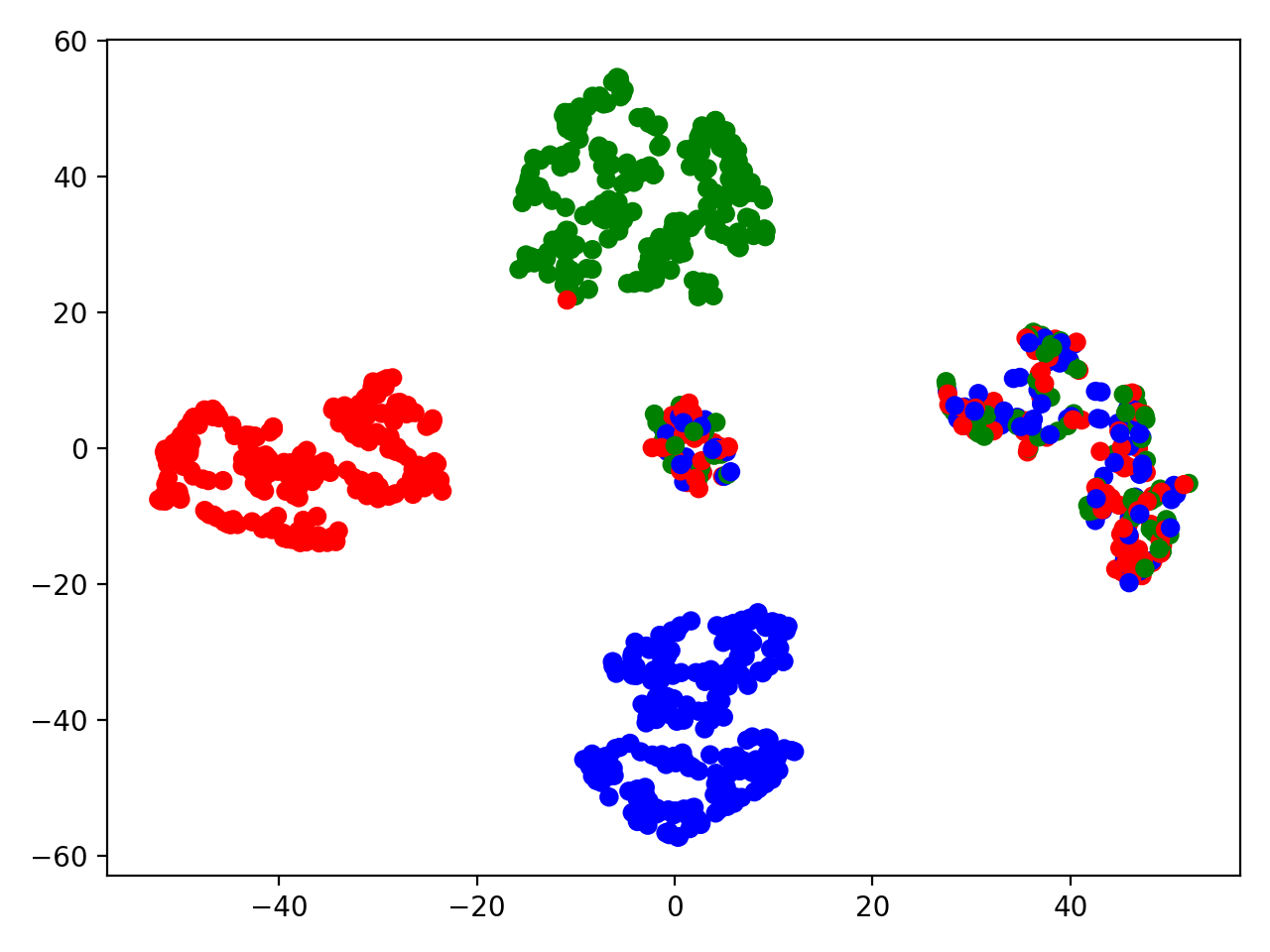}\\
\includegraphics[width=.32\linewidth]{figs/toy/setting_2/not_separable/tsne_LDA_setting_2_separable_False_run_0.png}%
\includegraphics[width=.32\linewidth]{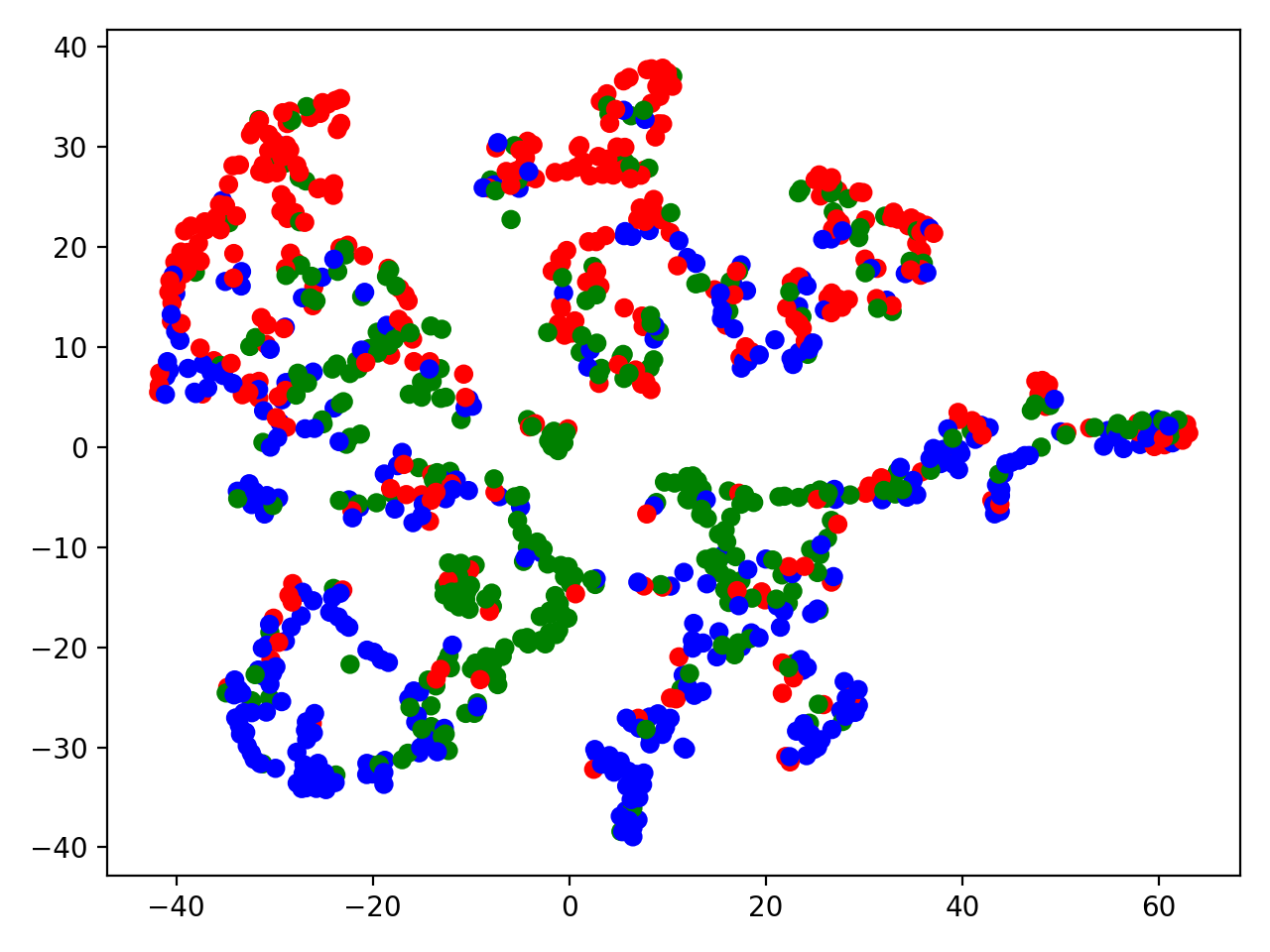}
\includegraphics[width=.32\linewidth]{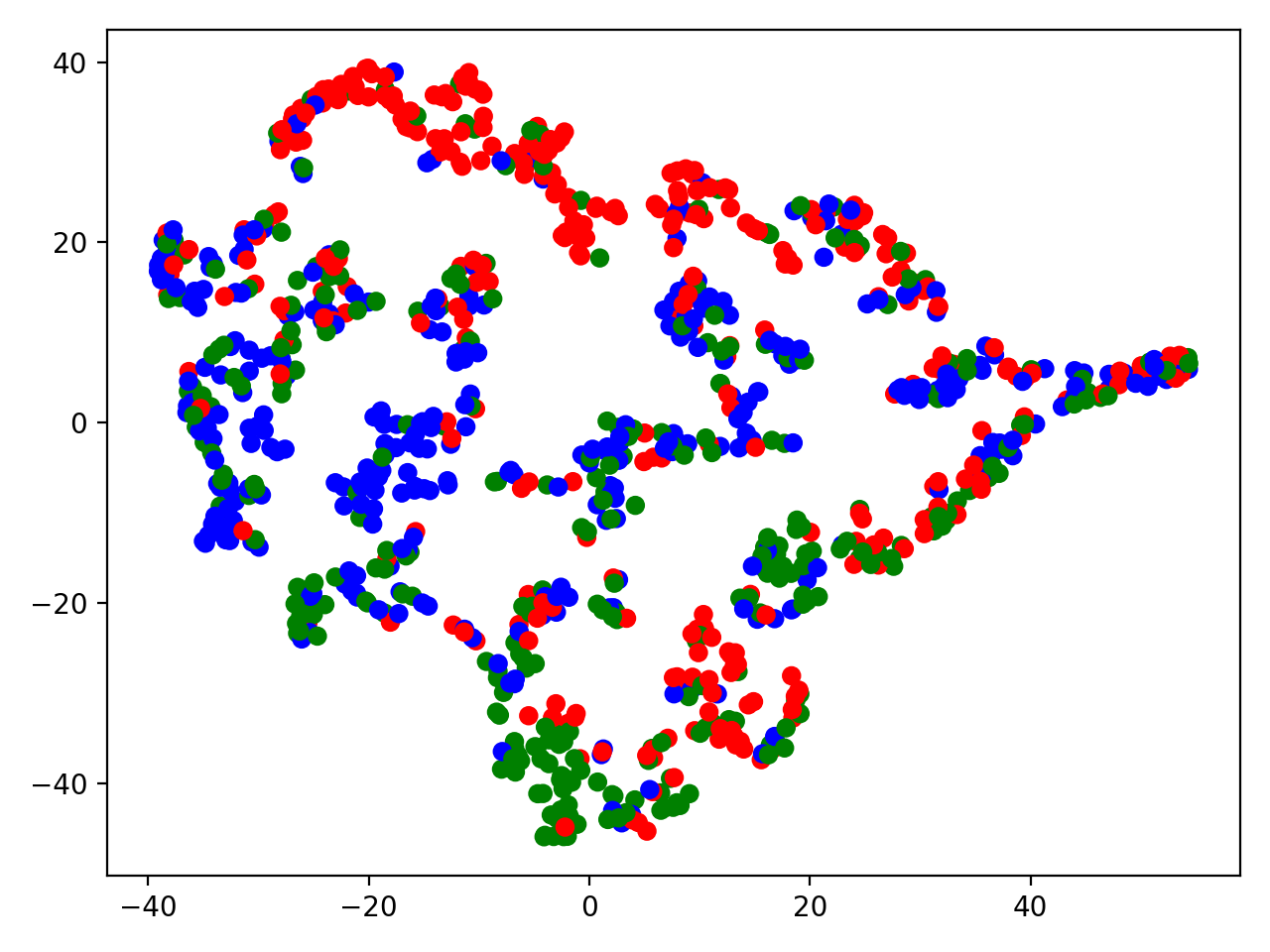}\\
\includegraphics[width=.32\linewidth]{figs/toy/setting_3/not_separable/tsne_LDA_setting_3_separable_False_run_0.png}%
\includegraphics[width=.32\linewidth]{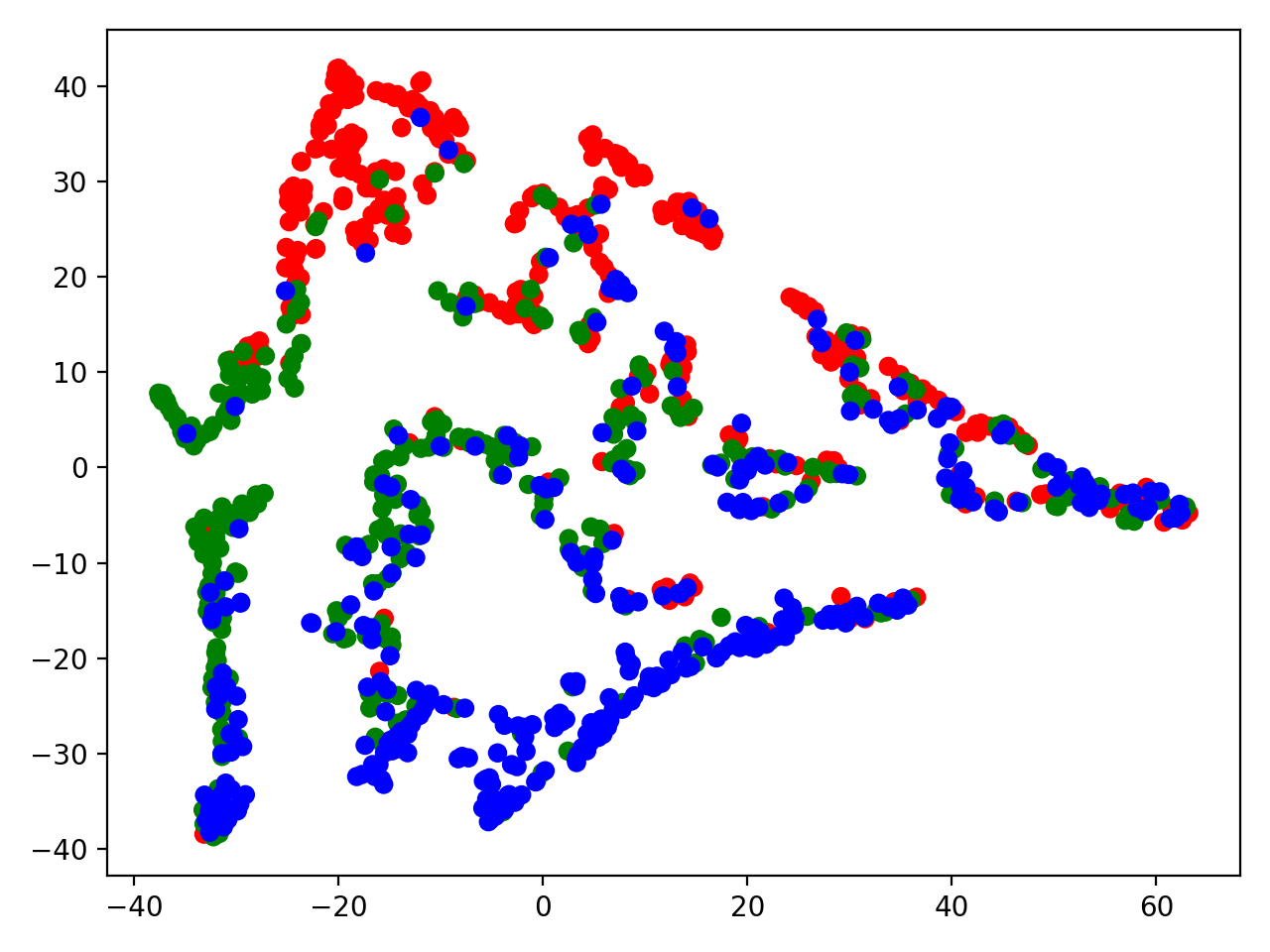}
\includegraphics[width=.32\linewidth]{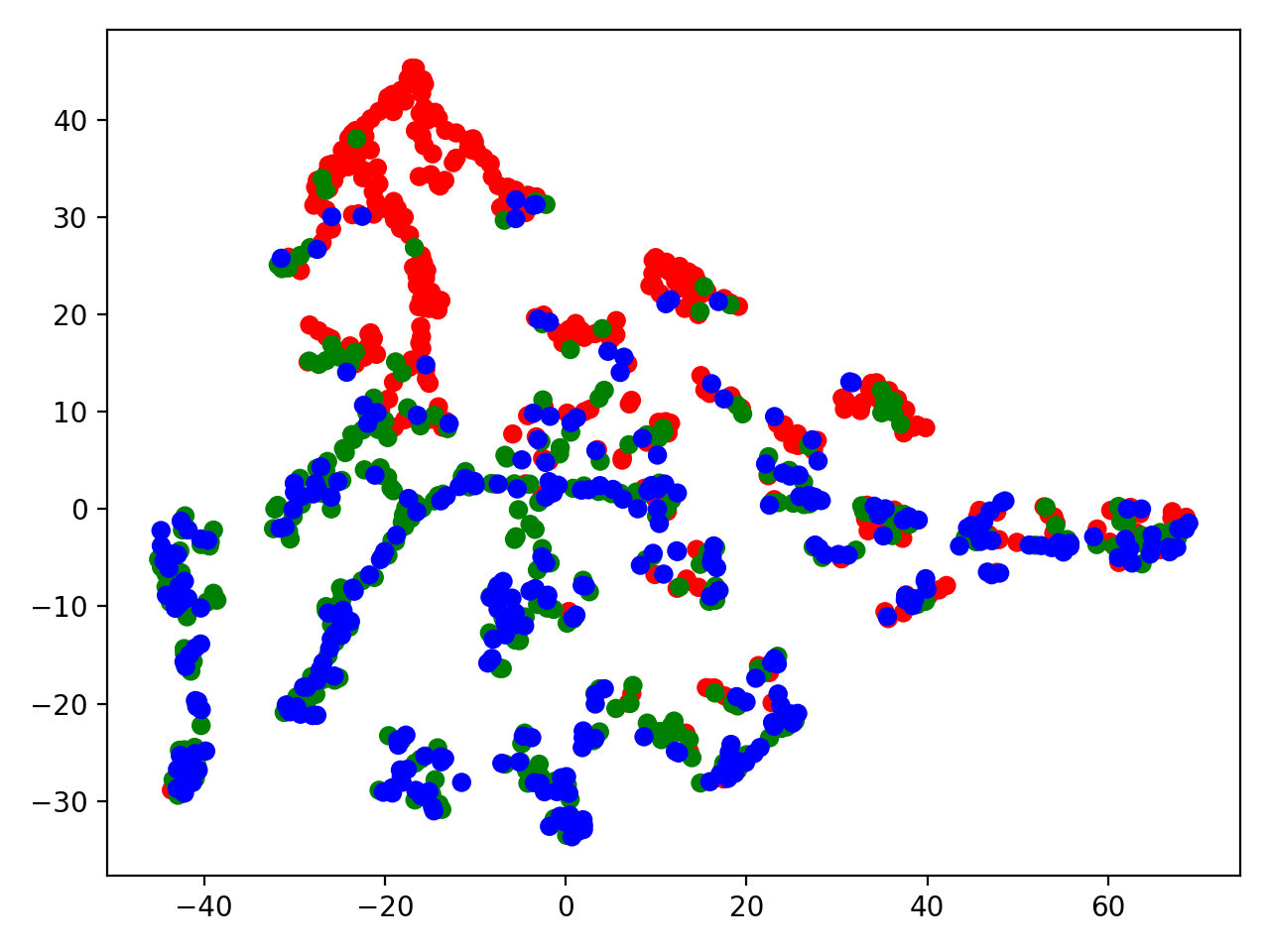}%
\end{tabular}
\end{subfigure}
\caption{Projections of $\hat{\theta}$ learned by LDA across three random restarts. Rows correspond to different settings of $\theta$ while columns correspond to different restarts. The grid of 9 figures on the left correspond to experiments where $\beta$ is identifiable and the grid of 9 figures on the right correspond to experiments where $\beta$ is non-identifiable. We see that while LDA tends to produce stable projections for identifiable $\beta$, it is highly unstable for non-identifiable $\beta$.}
\label{fig:lda_restarts}
\end{figure}

\begin{figure}[H]\centering
\begin{subfigure}[b]{0.49\textwidth}
\begin{tabular}[b]{@{}c@{}}
\includegraphics[width=.32\linewidth]{figs/toy/setting_1/separable/tsne_pfLDA_setting_1_separable_True_run_0.png}%
\includegraphics[width=.32\linewidth]{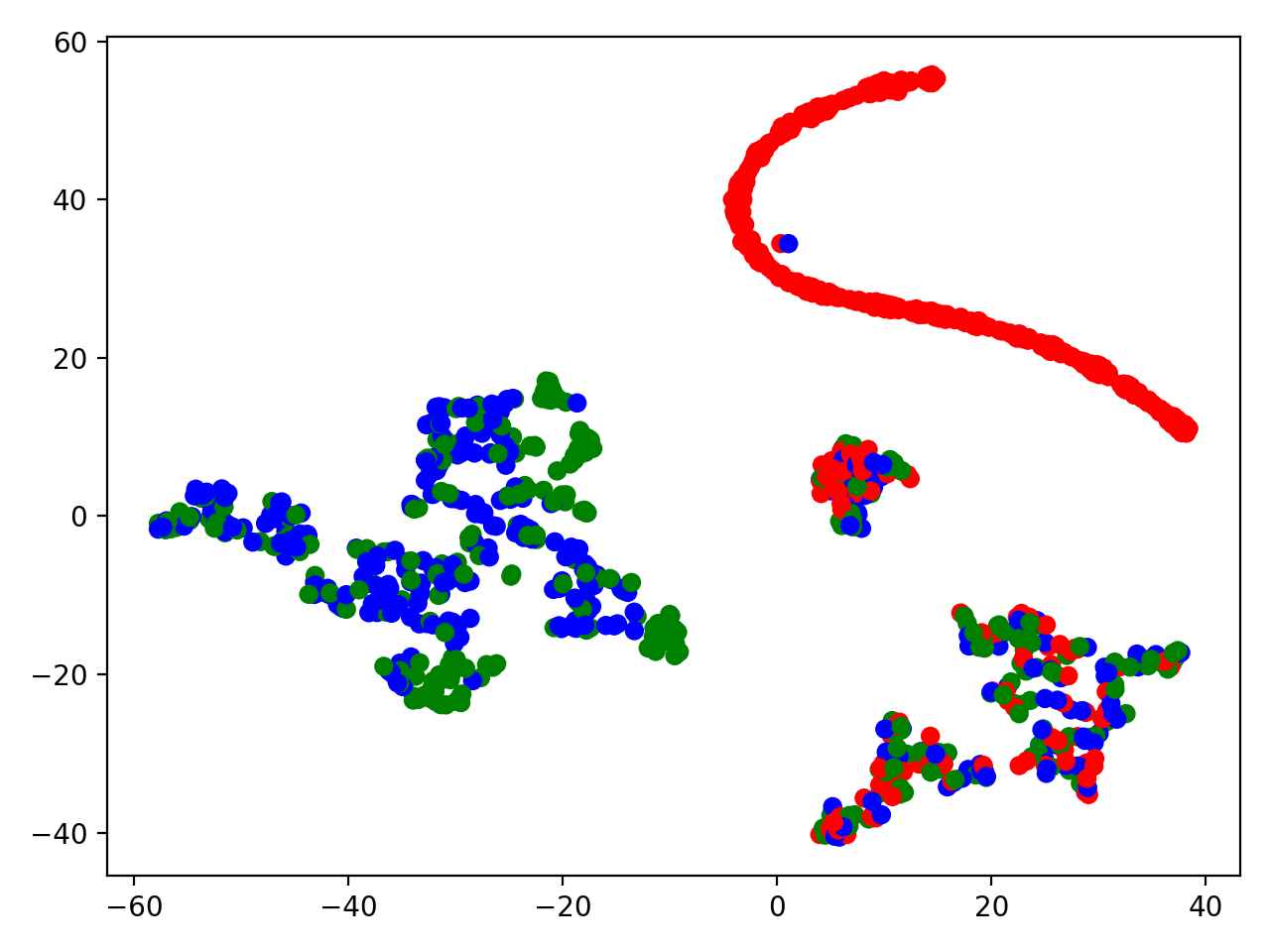}
\includegraphics[width=.32\linewidth]{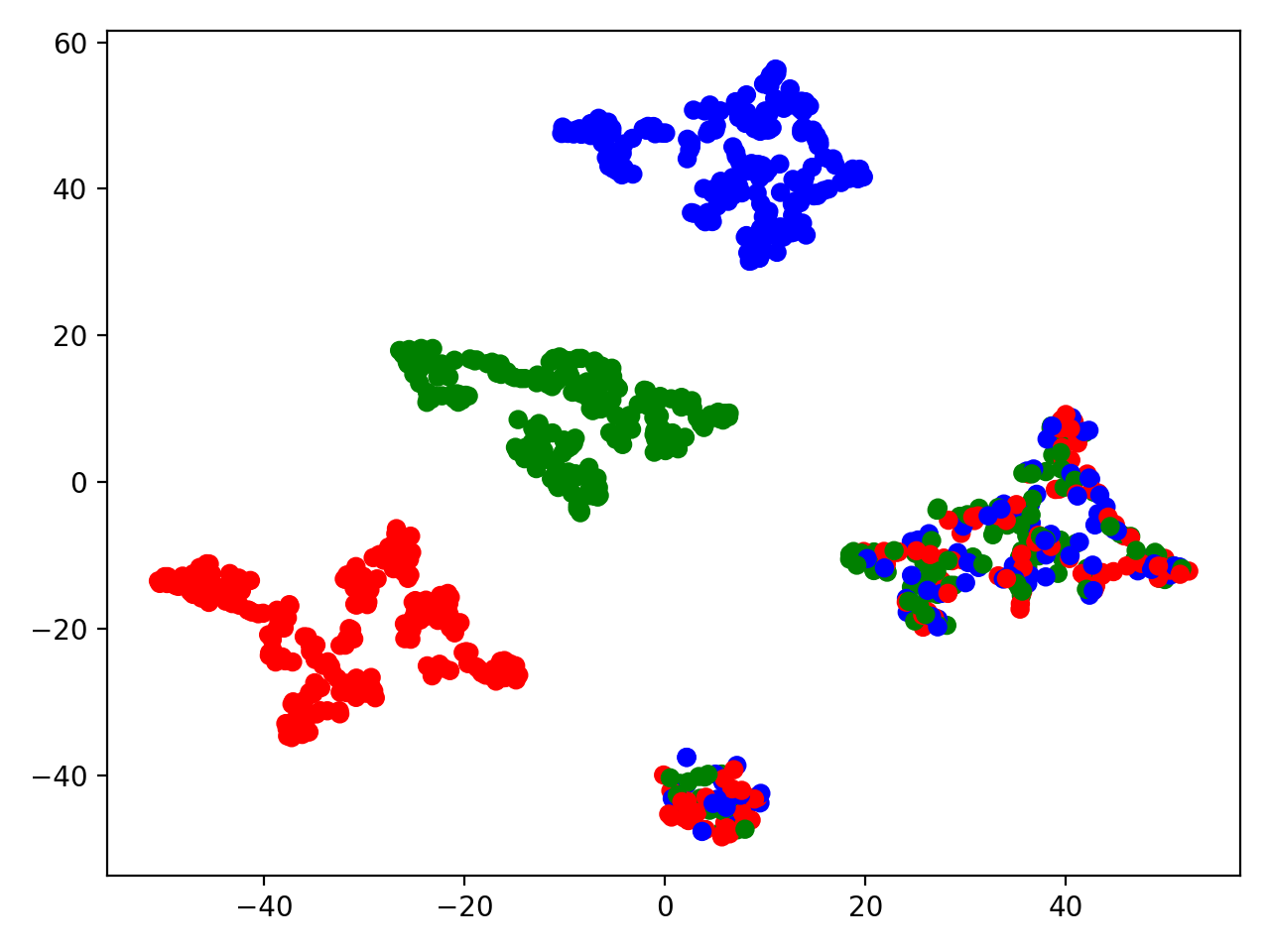}\\
\includegraphics[width=.32\linewidth]{figs/toy/setting_2/separable/tsne_pfLDA_setting_2_separable_True_run_0.png}%
\includegraphics[width=.32\linewidth]{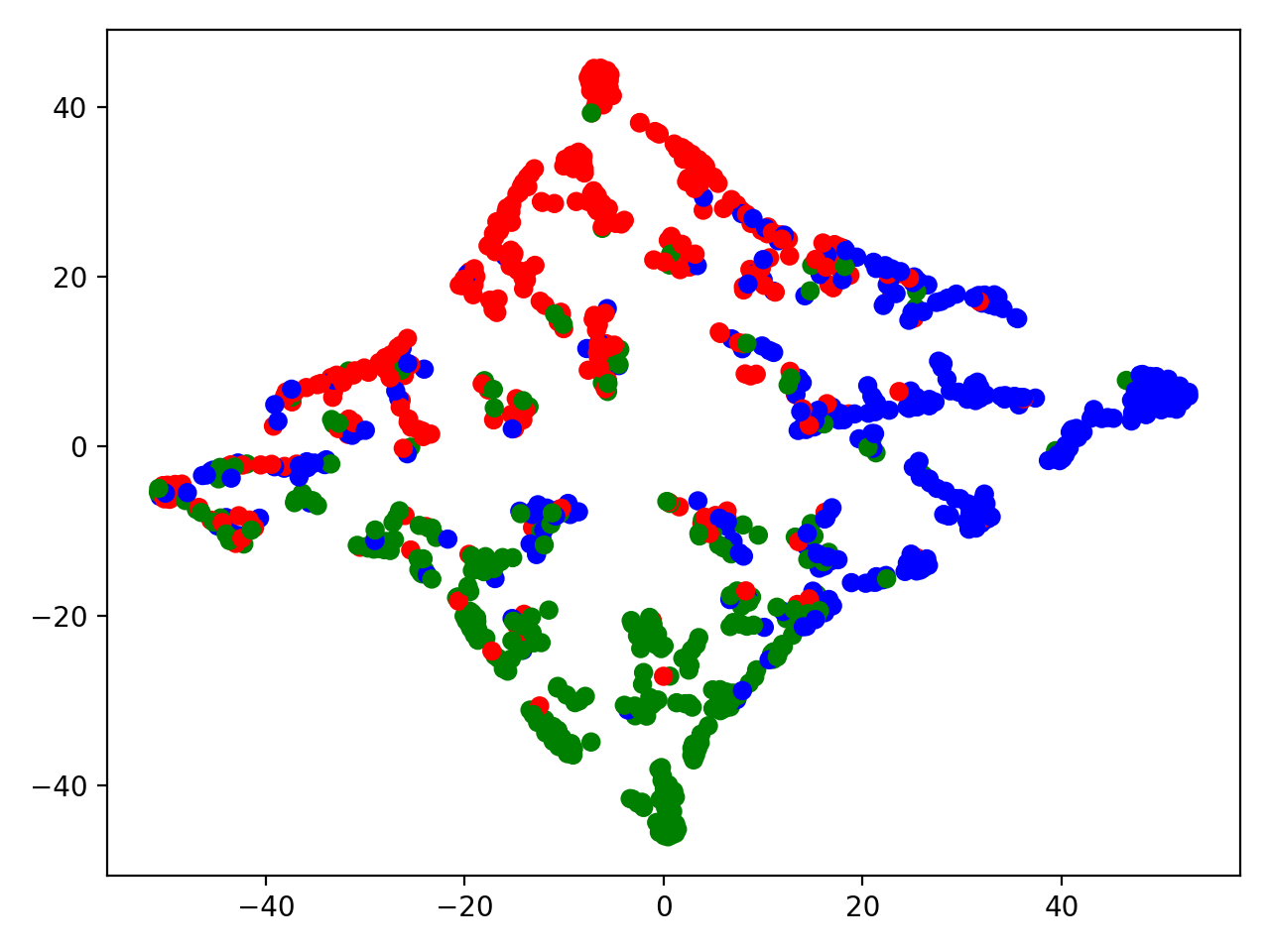}
\includegraphics[width=.32\linewidth]{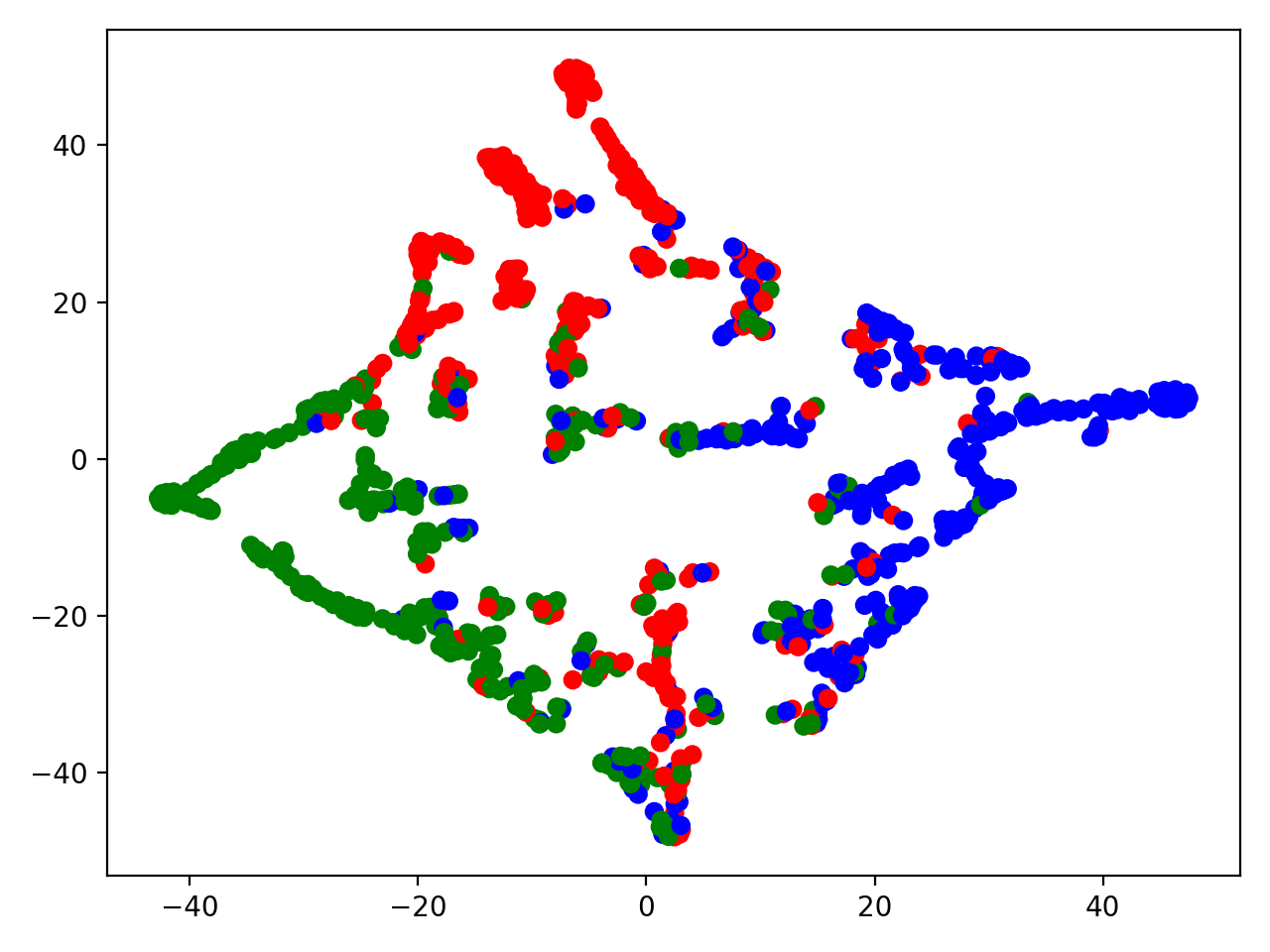}\\
\includegraphics[width=.32\linewidth]{figs/toy/setting_3/separable/tsne_pfLDA_setting_3_separable_True_run_0.png}%
\includegraphics[width=.32\linewidth]{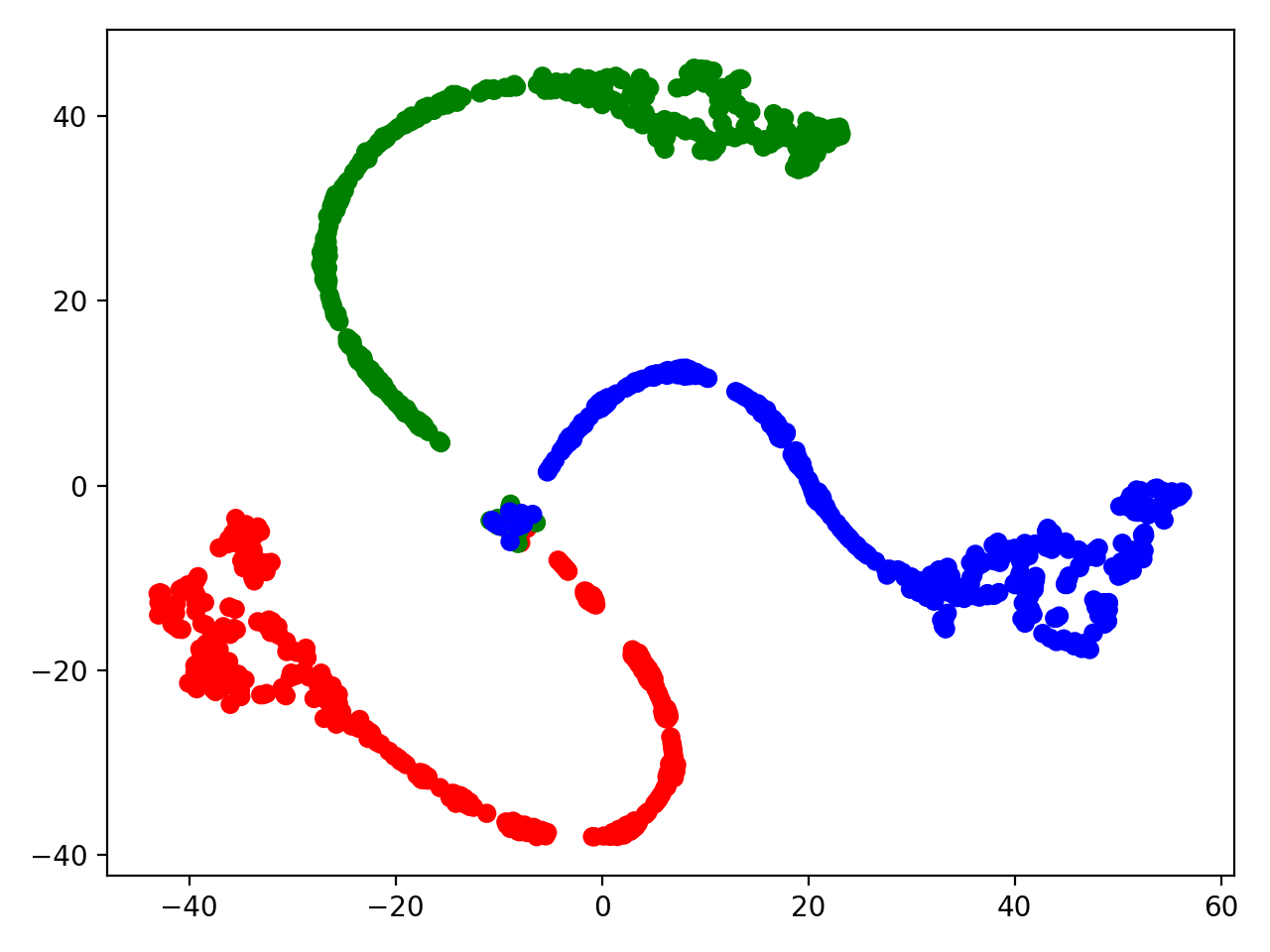}
\includegraphics[width=.32\linewidth]{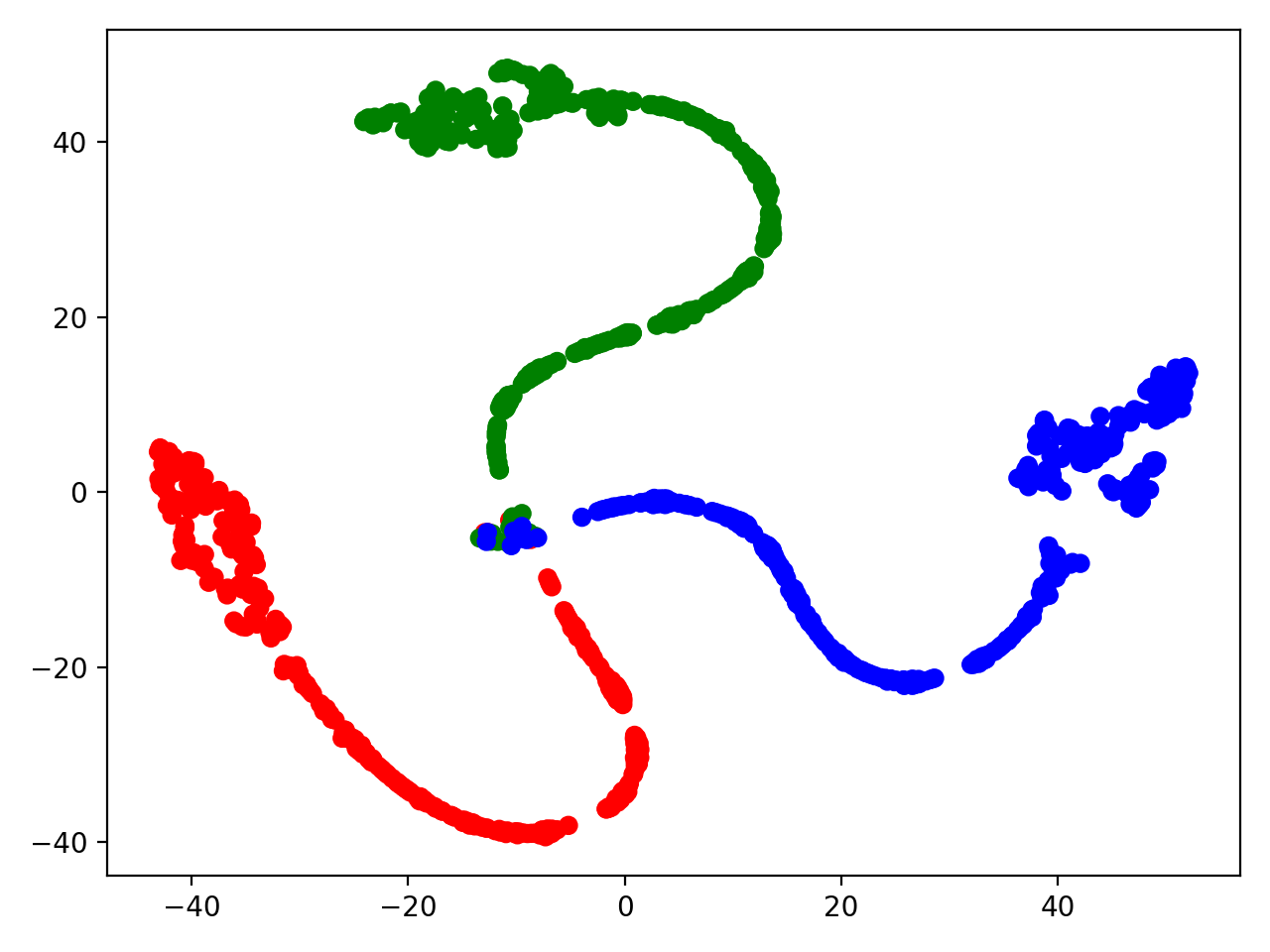}%
\end{tabular}
\end{subfigure}
\hfill
\begin{subfigure}[b]{0.49\textwidth}
\begin{tabular}[b]{@{}c@{}}
\includegraphics[width=.32\linewidth]{figs/toy/setting_1/not_separable/tsne_pfLDA_setting_1_separable_False_run_0.png}%
\includegraphics[width=.32\linewidth]{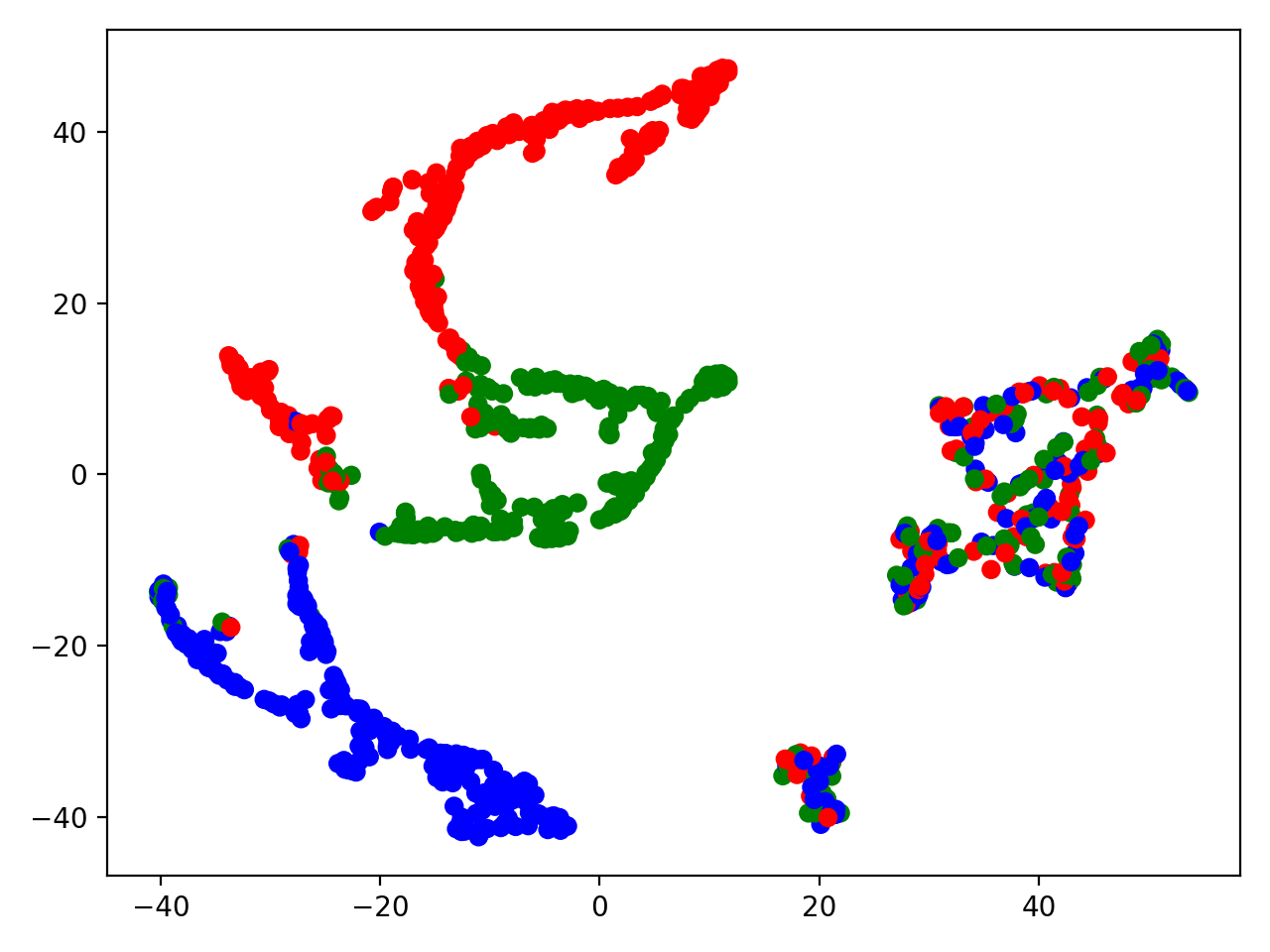}
\includegraphics[width=.32\linewidth]{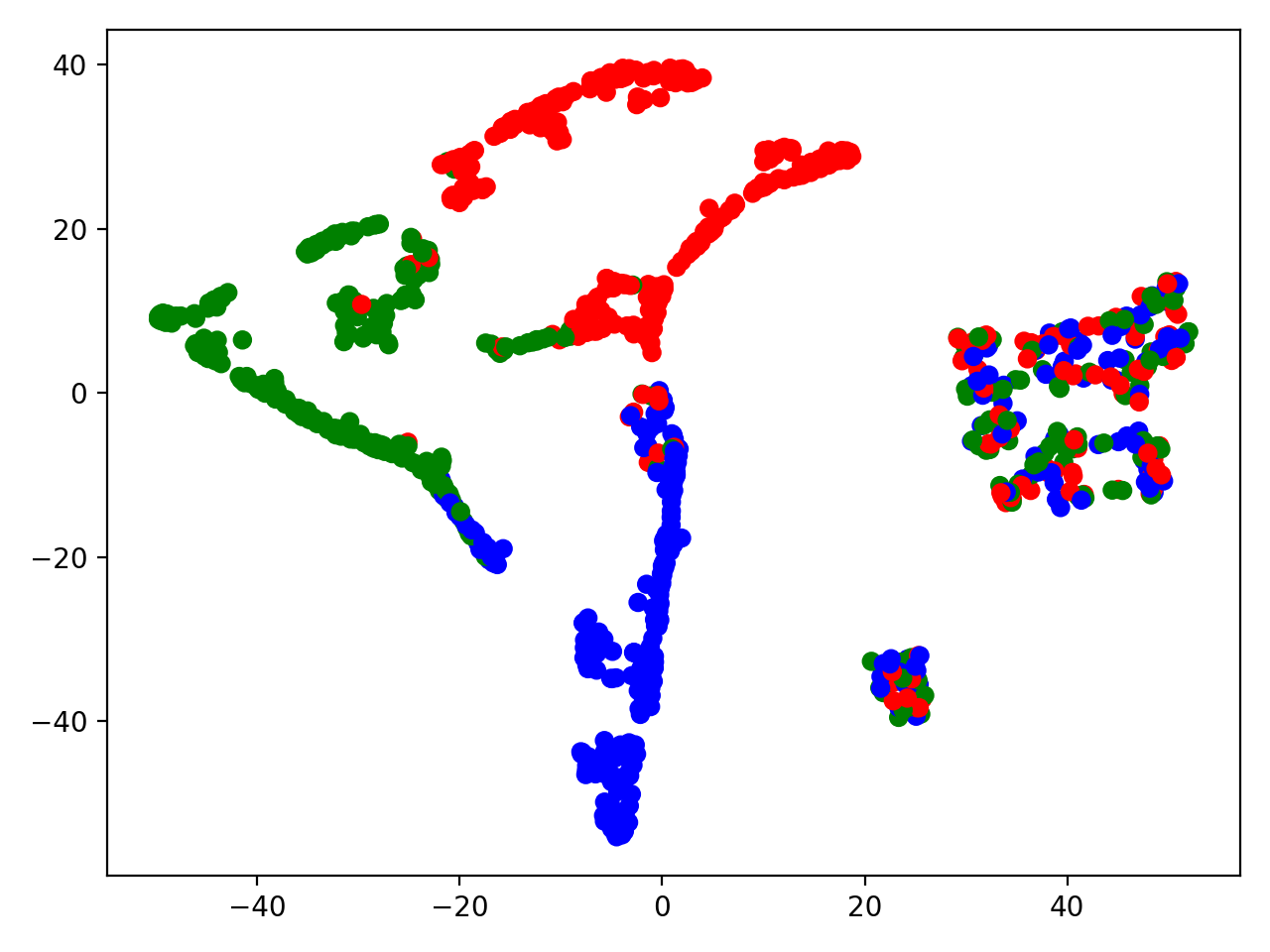}\\
\includegraphics[width=.32\linewidth]{figs/toy/setting_2/not_separable/tsne_pfLDA_setting_2_separable_False_run_0.png}%
\includegraphics[width=.32\linewidth]{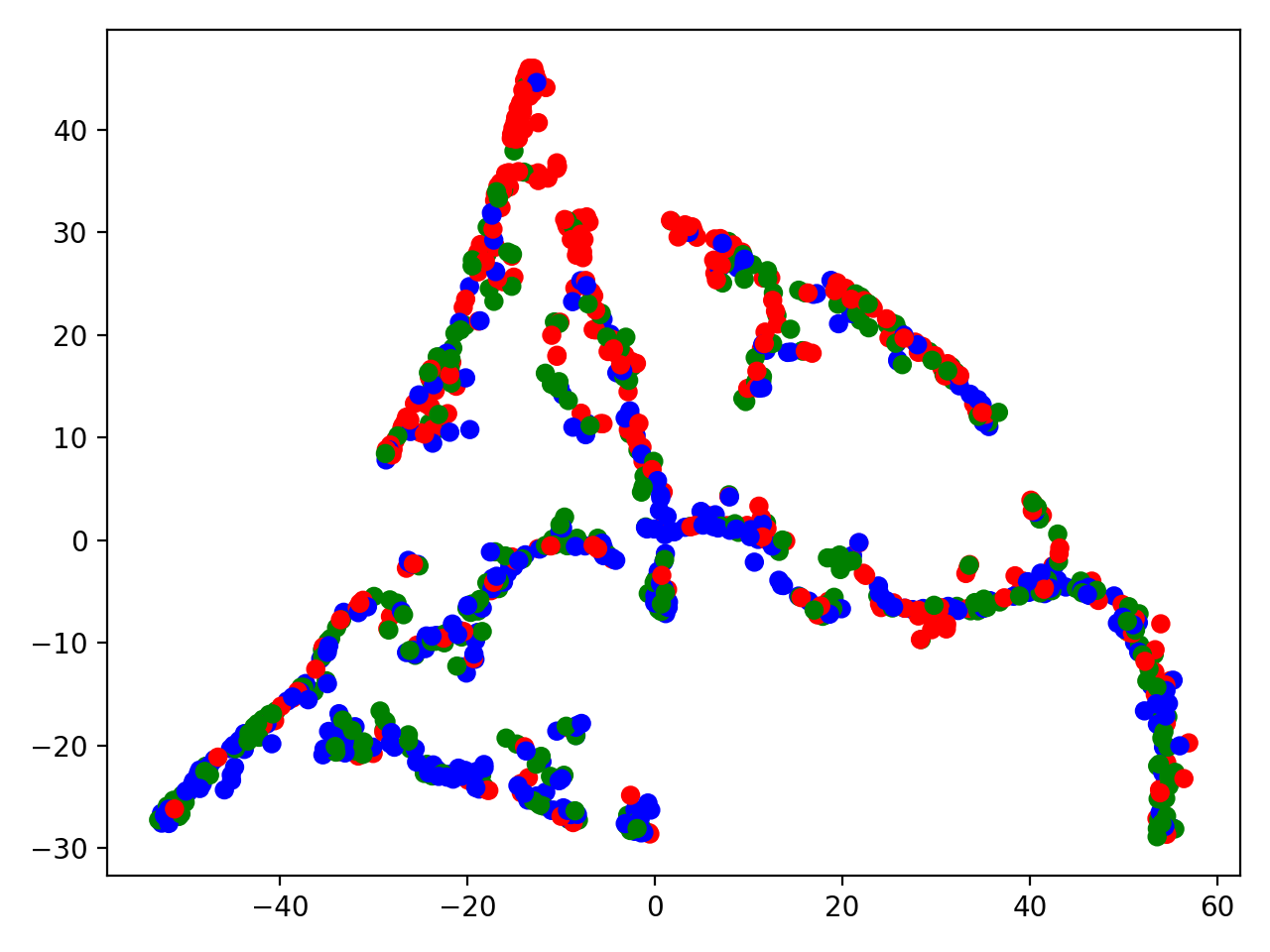}
\includegraphics[width=.32\linewidth]{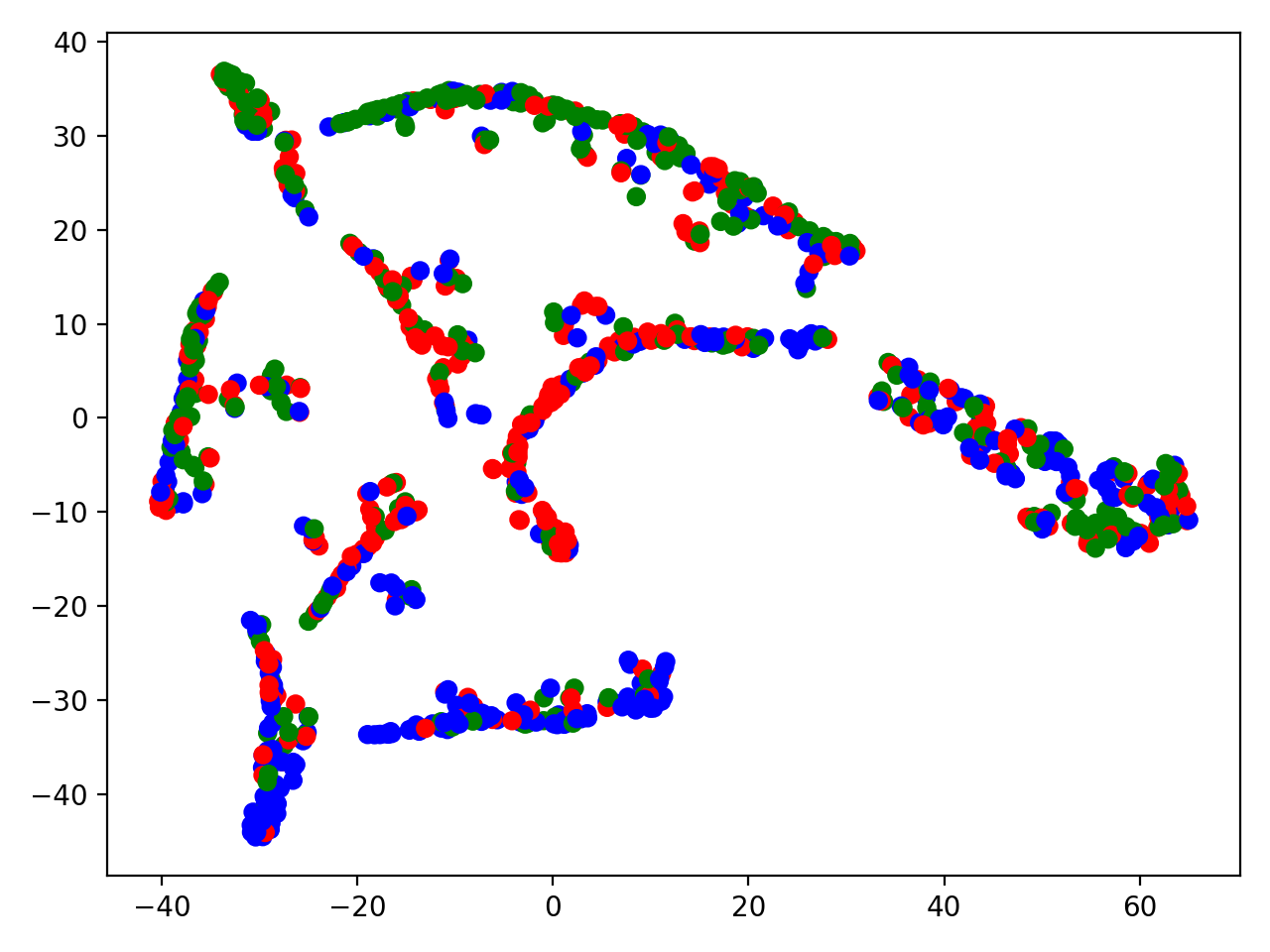}\\
\includegraphics[width=.32\linewidth]{figs/toy/setting_3/not_separable/tsne_pfLDA_setting_3_separable_False_run_0.png}%
\includegraphics[width=.32\linewidth]{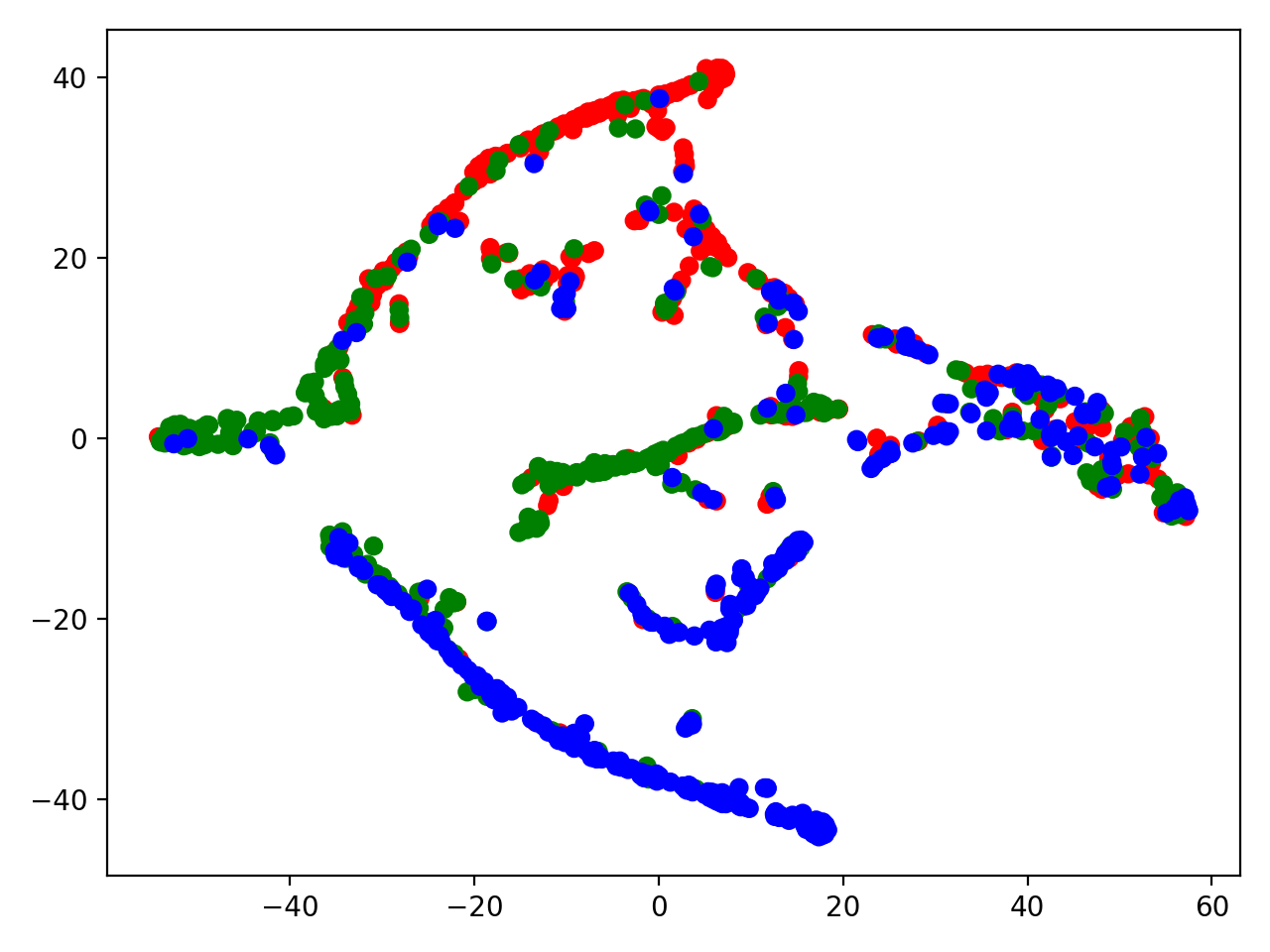}
\includegraphics[width=.32\linewidth]{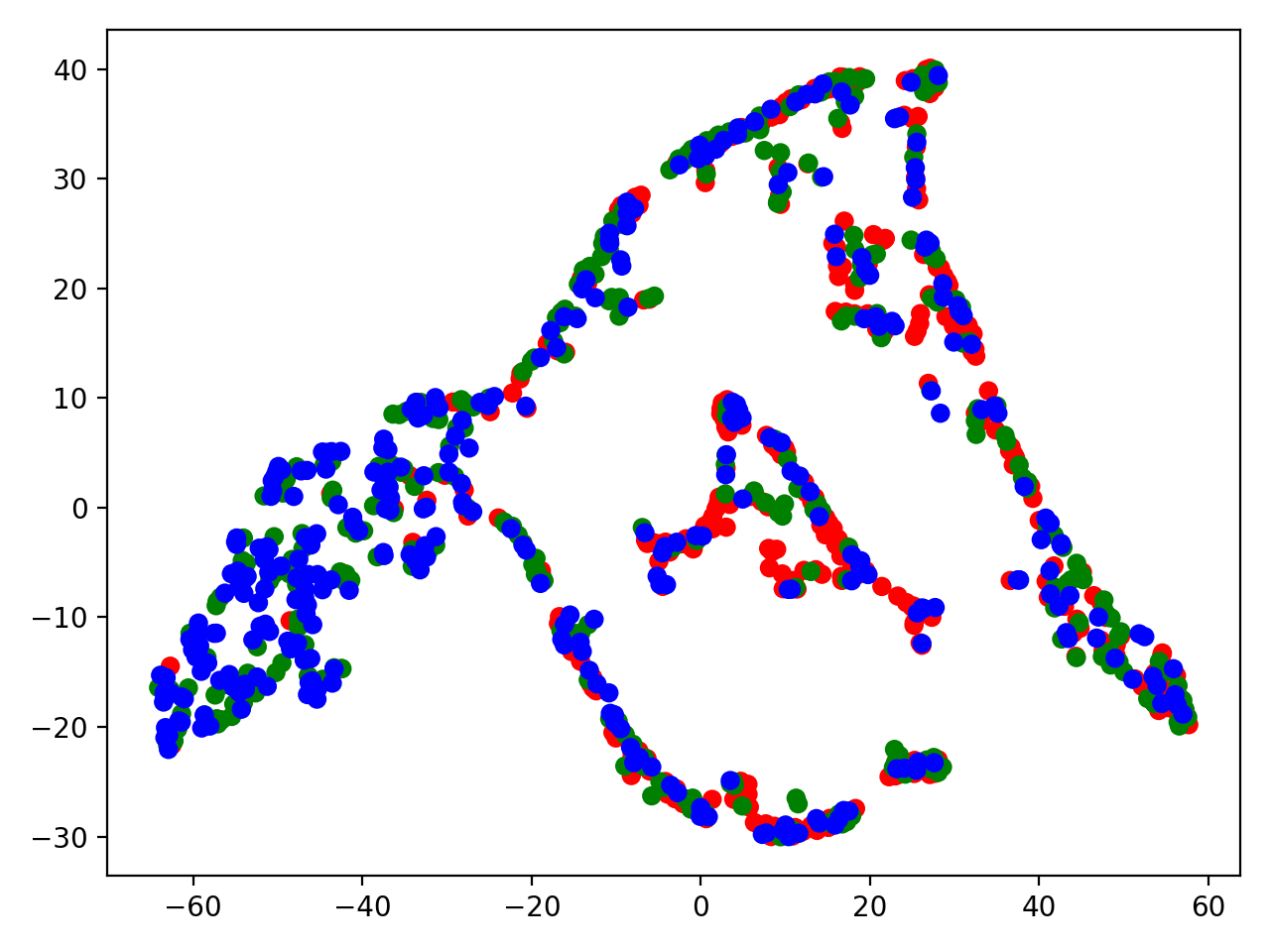}%
\end{tabular}
\end{subfigure}
\caption{Projections of $\hat{\theta}$ learned by pf-sLDA across three random restarts (see caption of Figure \ref{fig:lda_restarts} for interpretation of figures). We see that while pf-sLDA also produces relatively stable projections for identifiable $\beta$ (similar to LDA), it continues to be unstable for non-identifiable $\beta$ for settings 2 and 3.}
\label{fig:pf-slda-restarts}
\end{figure}

\begin{figure}[H]\centering
\begin{subfigure}[b]{0.49\textwidth}
\begin{tabular}[b]{@{}c@{}}
\includegraphics[width=.32\linewidth]{figs/toy/setting_1/separable/tsne_ours_setting_1_separable_True_run_0.png}%
\includegraphics[width=.32\linewidth]{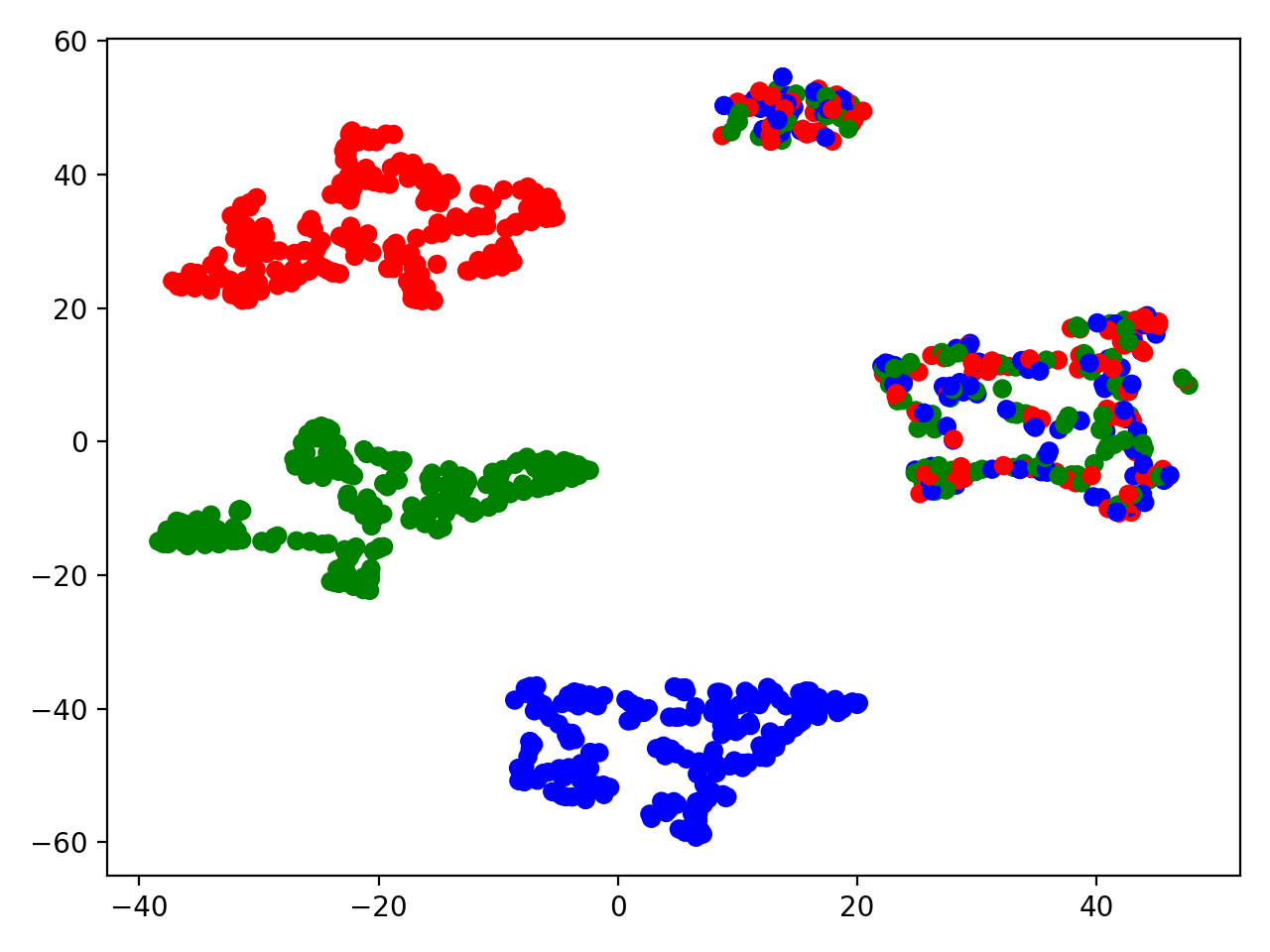}
\includegraphics[width=.32\linewidth]{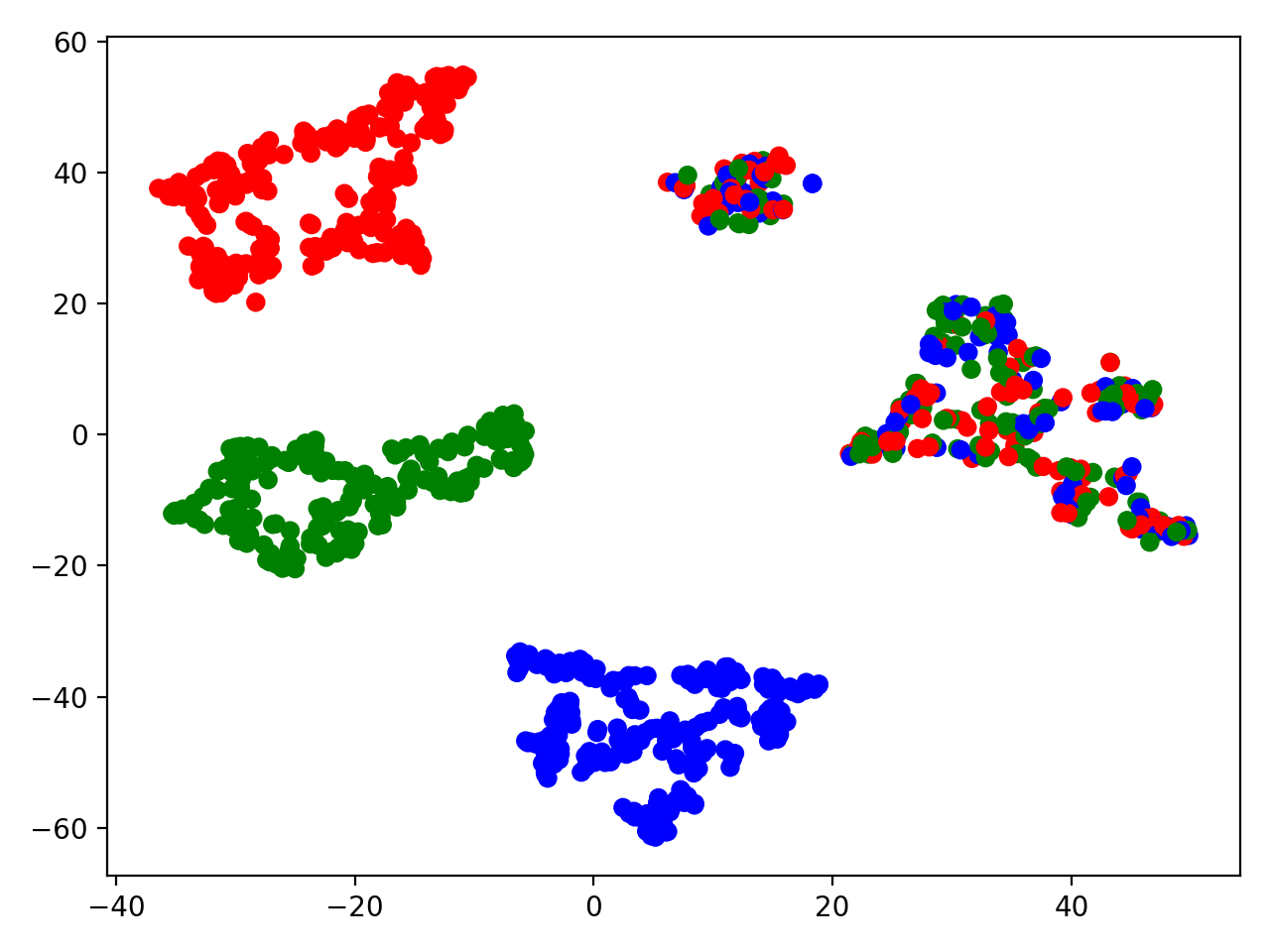}\\
\includegraphics[width=.32\linewidth]{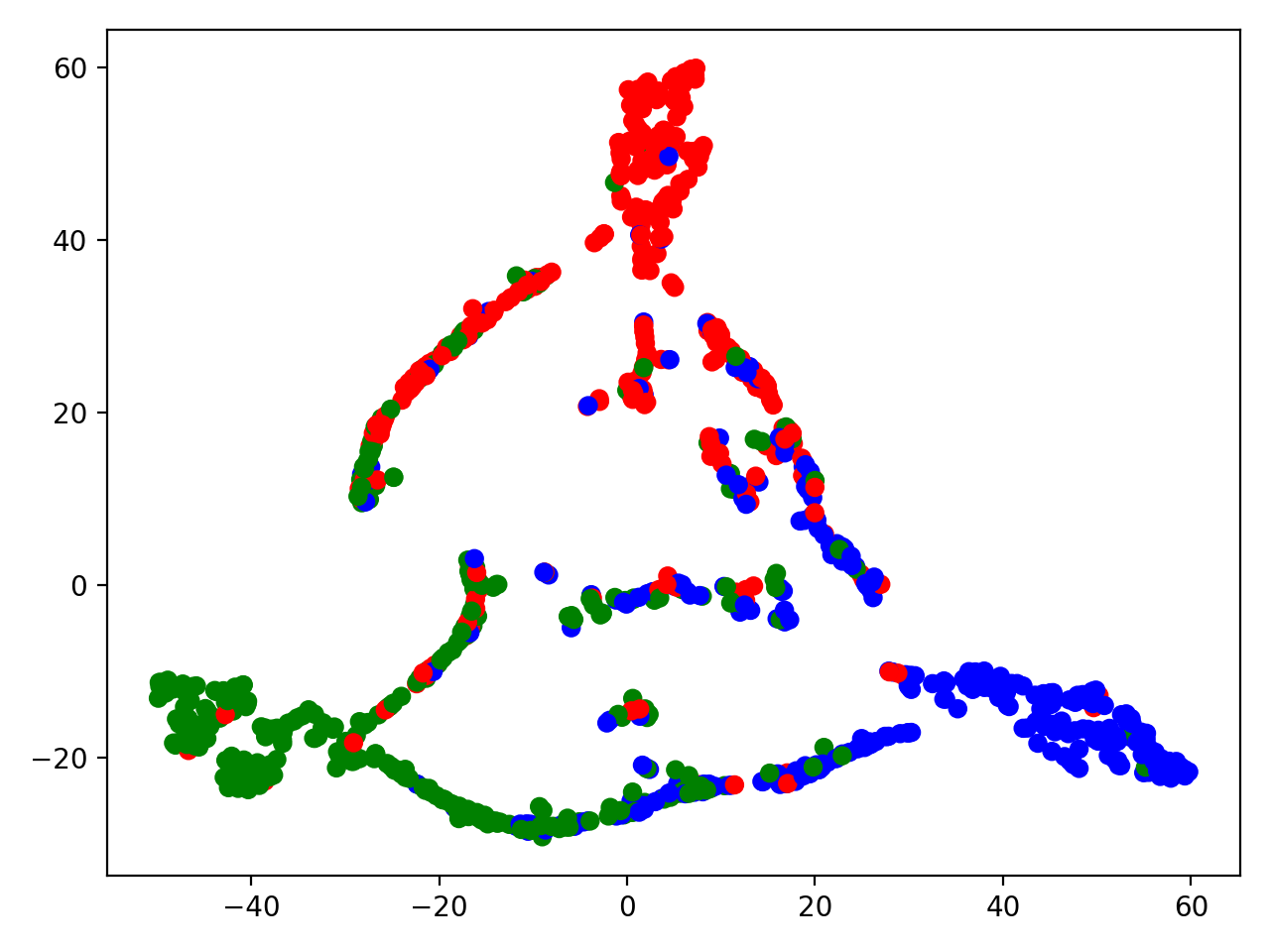}%
\includegraphics[width=.32\linewidth]{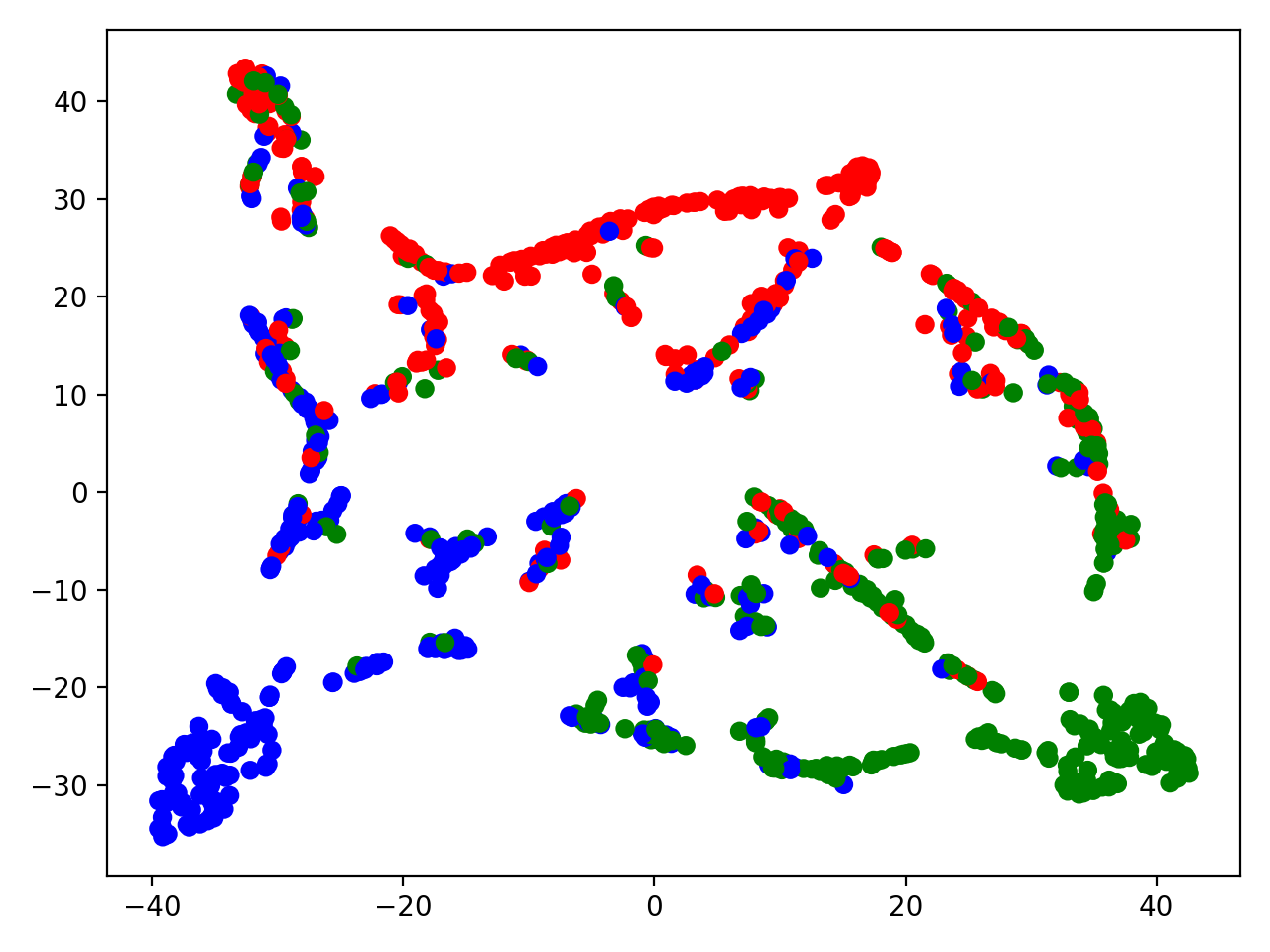}
\includegraphics[width=.32\linewidth]{figs/toy/setting_2/separable/tsne_ours_setting_2_separable_True_run_2.png}\\
\includegraphics[width=.32\linewidth]{figs/toy/setting_3/separable/tsne_ours_setting_3_separable_True_run_0.png}%
\includegraphics[width=.32\linewidth]{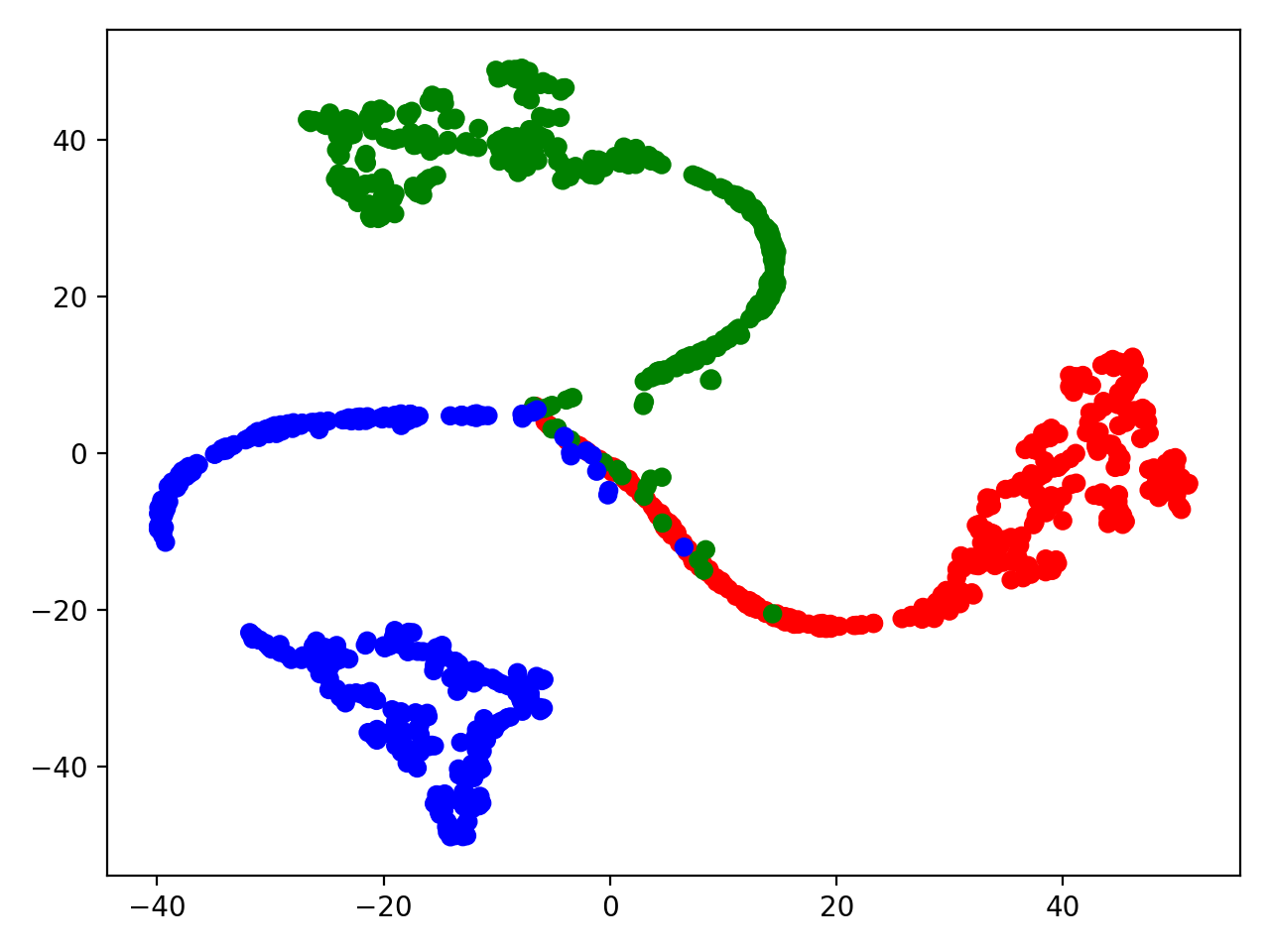}
\includegraphics[width=.32\linewidth]{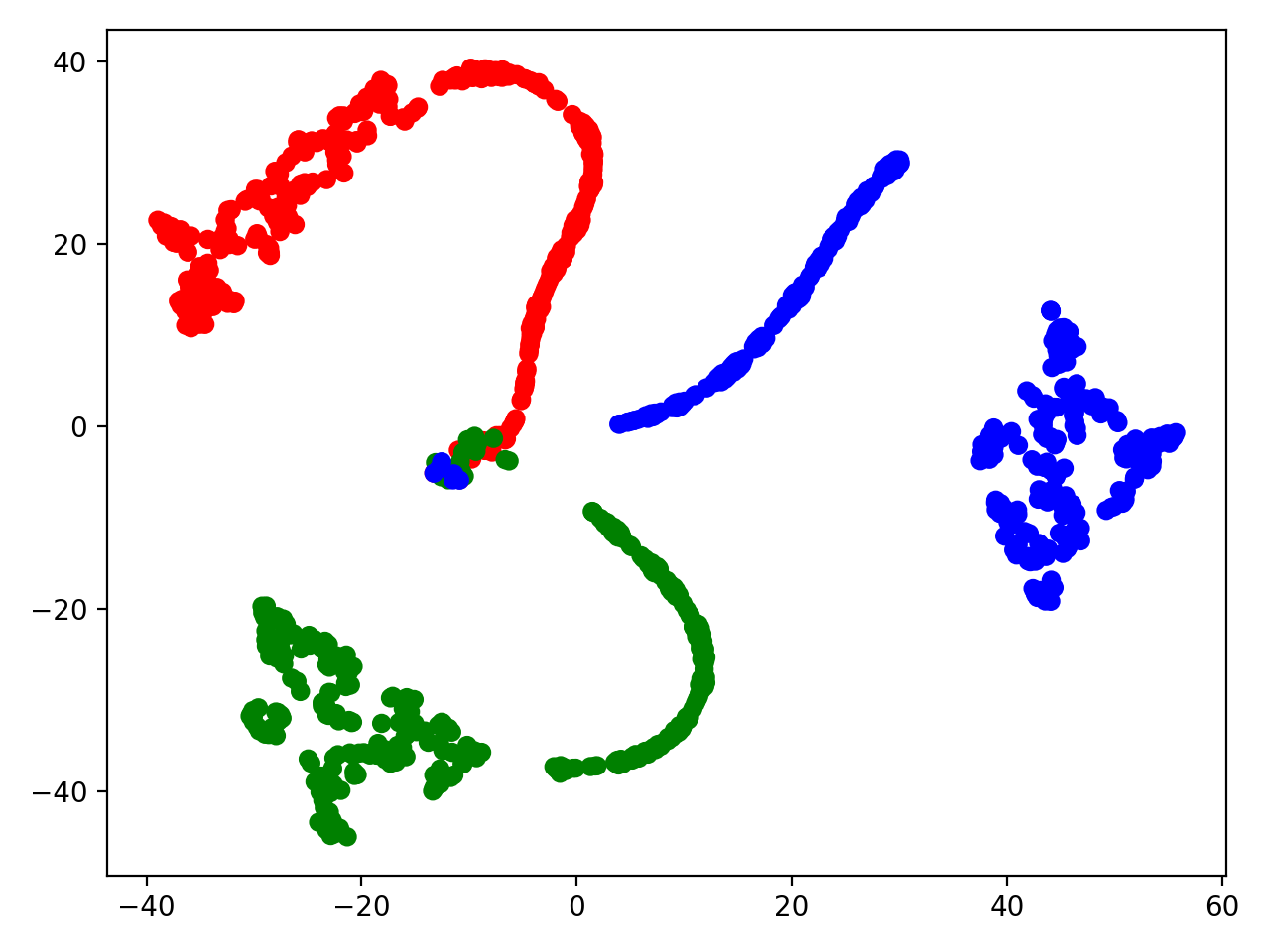}%
\end{tabular}
\end{subfigure}
\hfill
\begin{subfigure}[b]{0.49\textwidth}
\begin{tabular}[b]{@{}c@{}}
\includegraphics[width=.32\linewidth]{figs/toy/setting_1/not_separable/tsne_ours_setting_1_separable_False_run_0.png}%
\includegraphics[width=.32\linewidth]{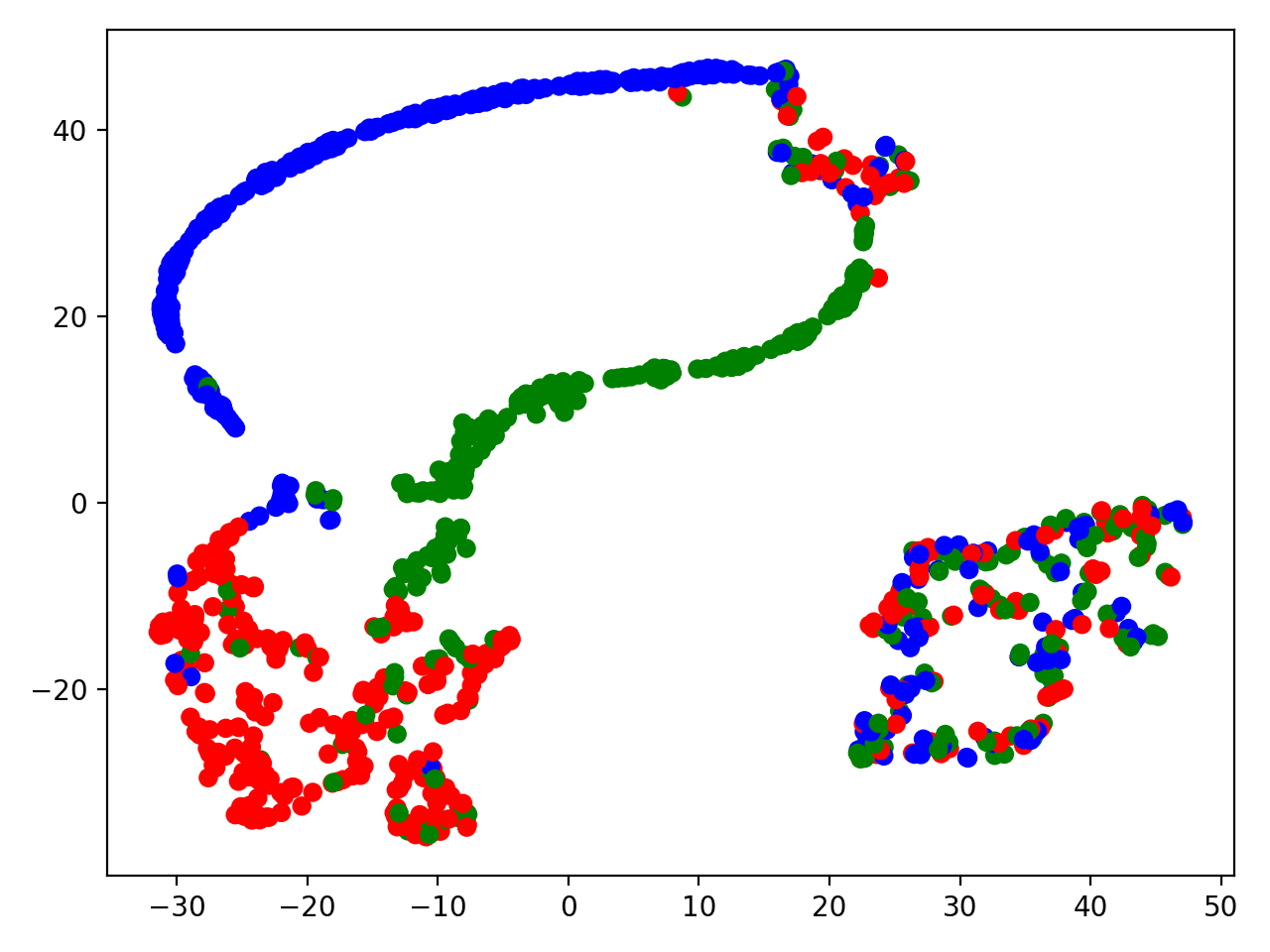}
\includegraphics[width=.32\linewidth]{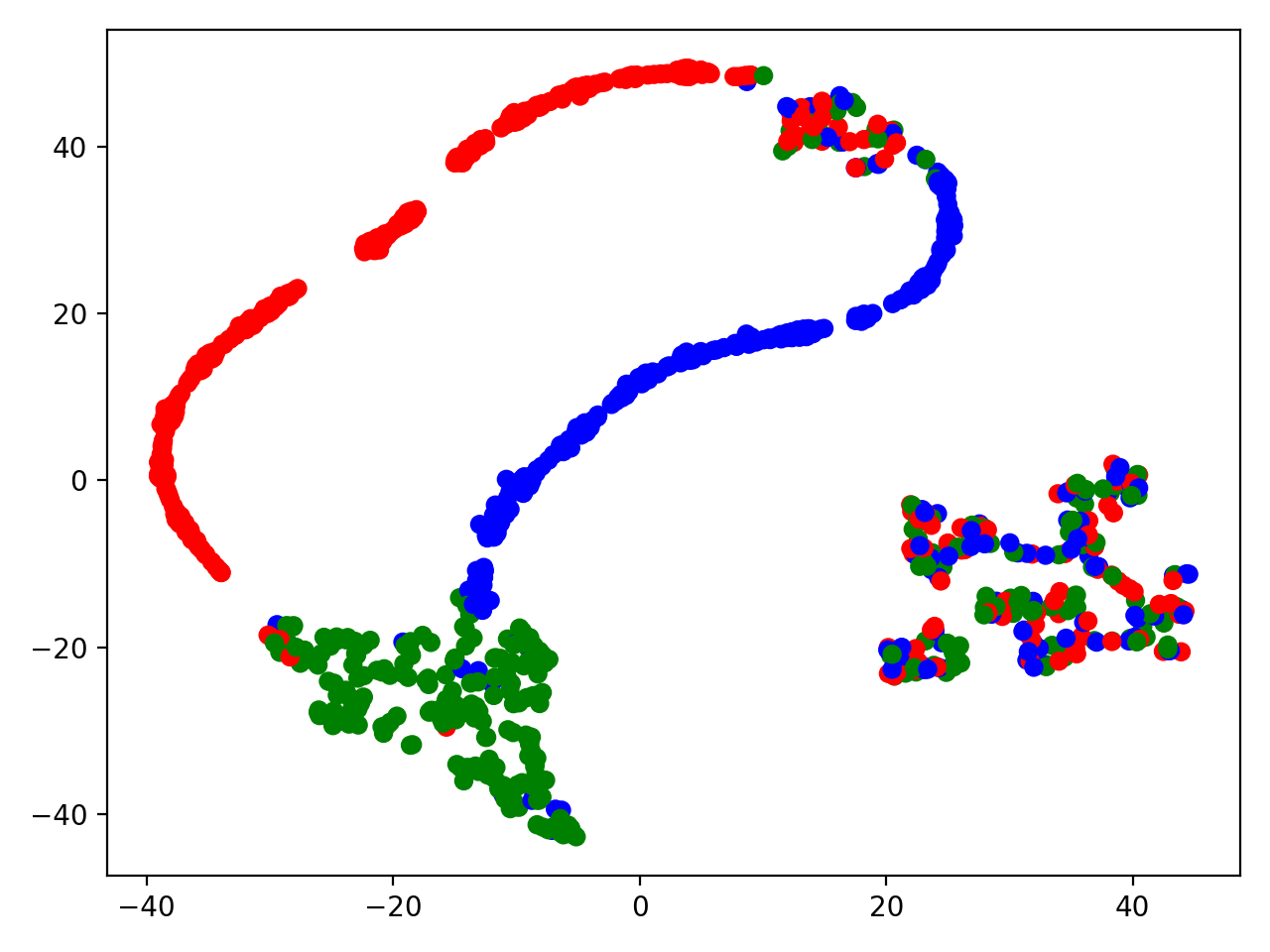}\\
\includegraphics[width=.32\linewidth]{figs/toy/setting_2/not_separable/tsne_ours_setting_2_separable_False_run_0.png}%
\includegraphics[width=.32\linewidth]{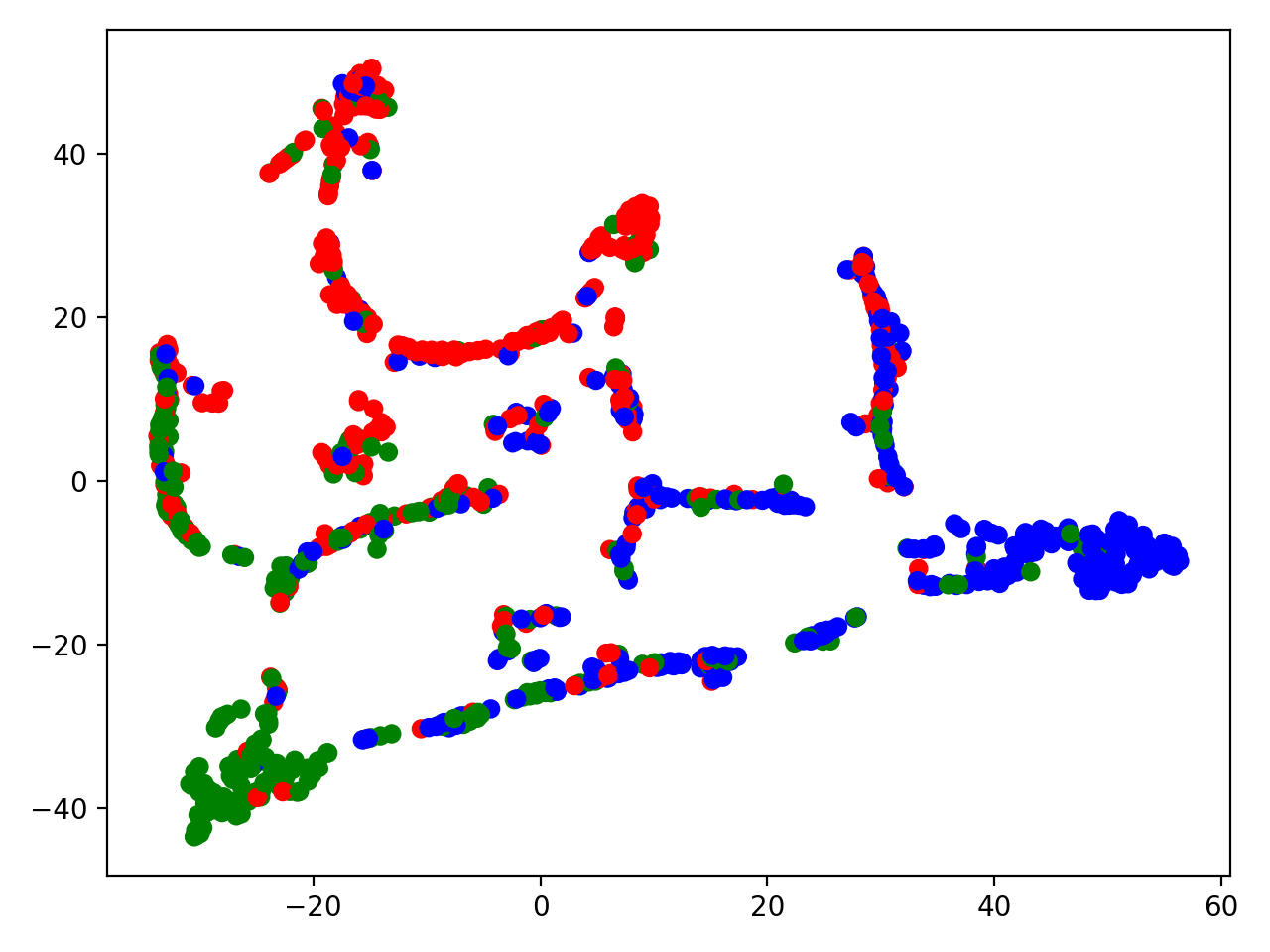}
\includegraphics[width=.32\linewidth]{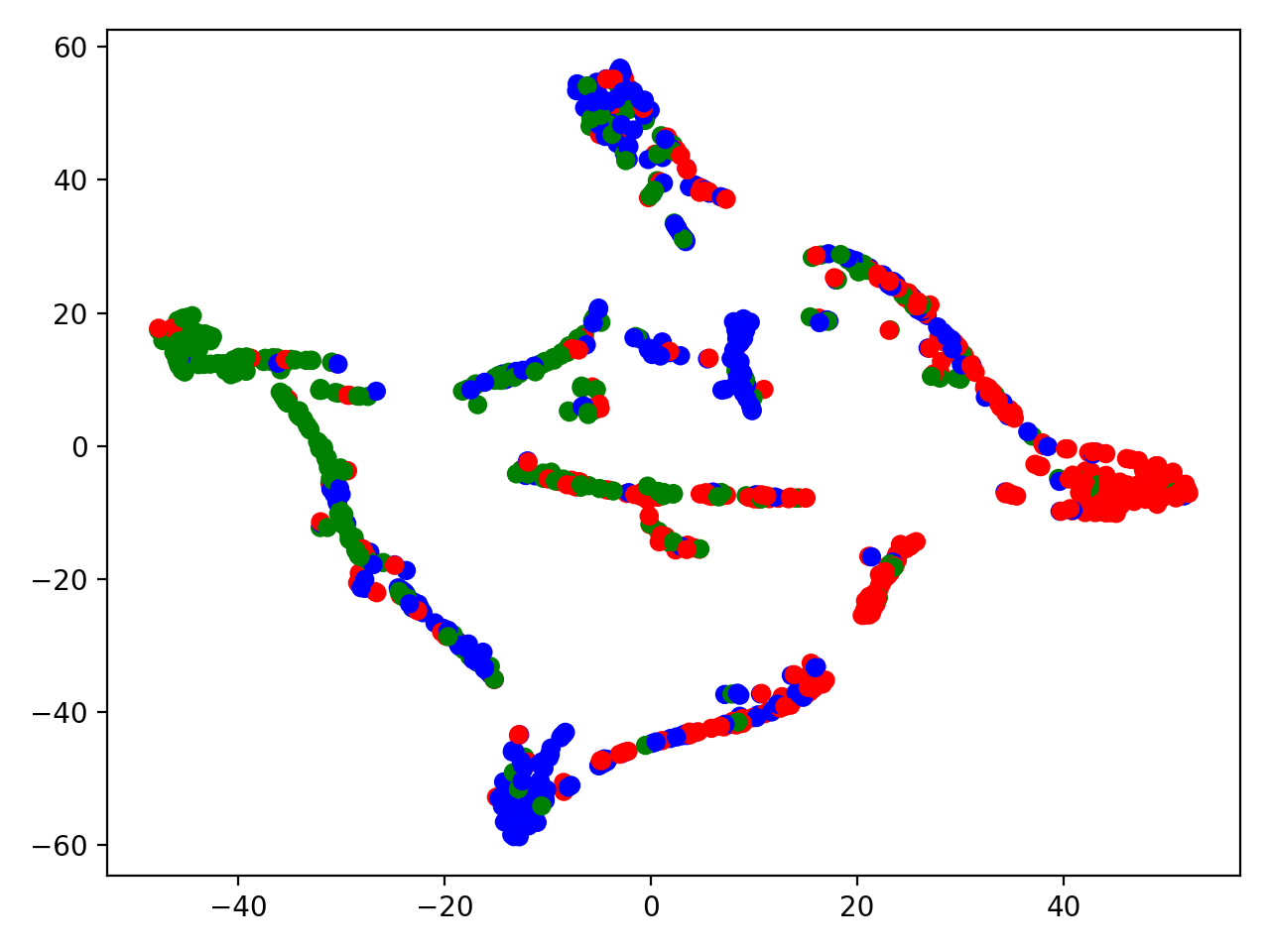}\\
\includegraphics[width=.32\linewidth]{figs/toy/setting_3/not_separable/tsne_ours_setting_3_separable_False_run_0.png}%
\includegraphics[width=.32\linewidth]{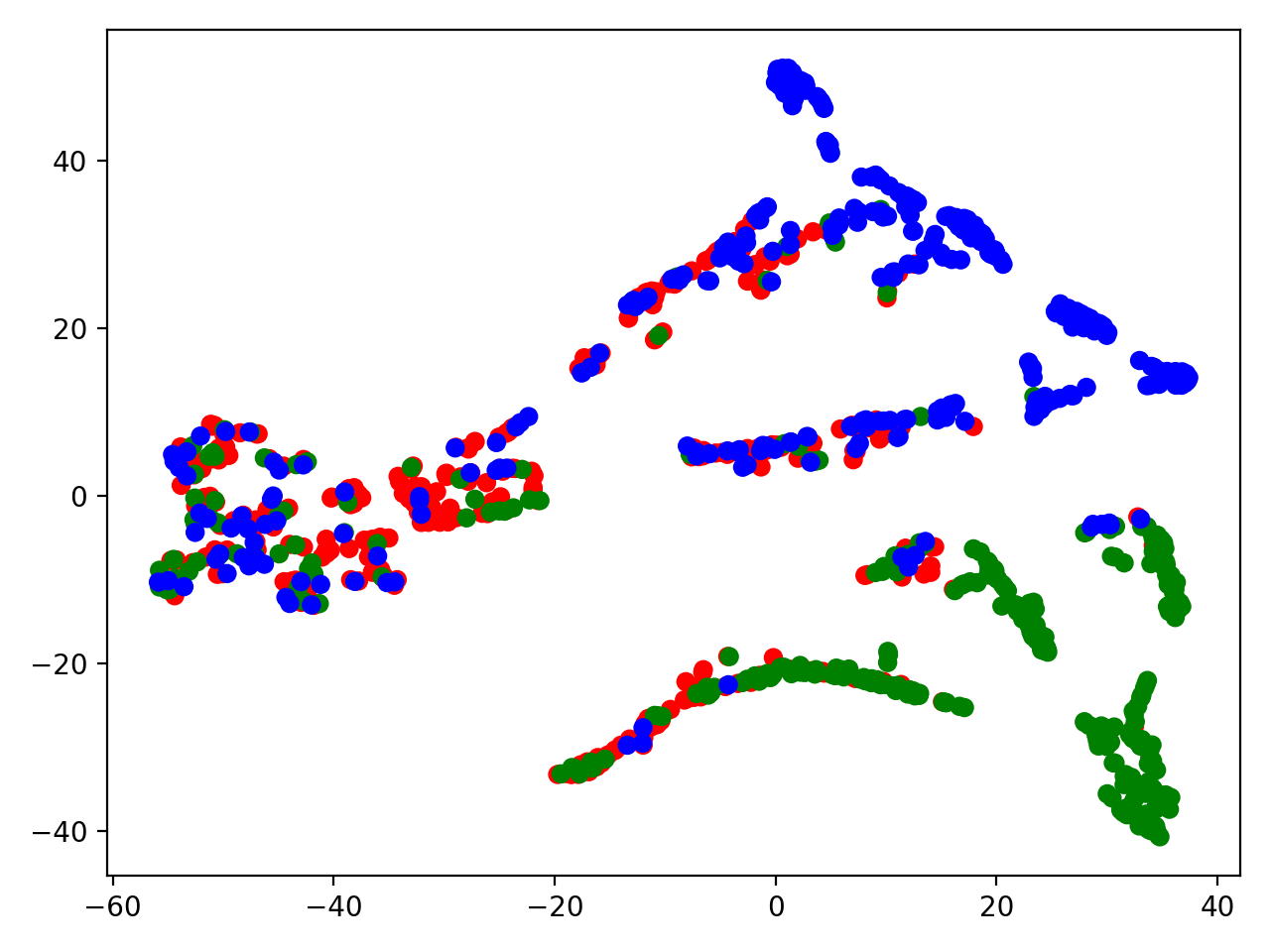}
\includegraphics[width=.32\linewidth]{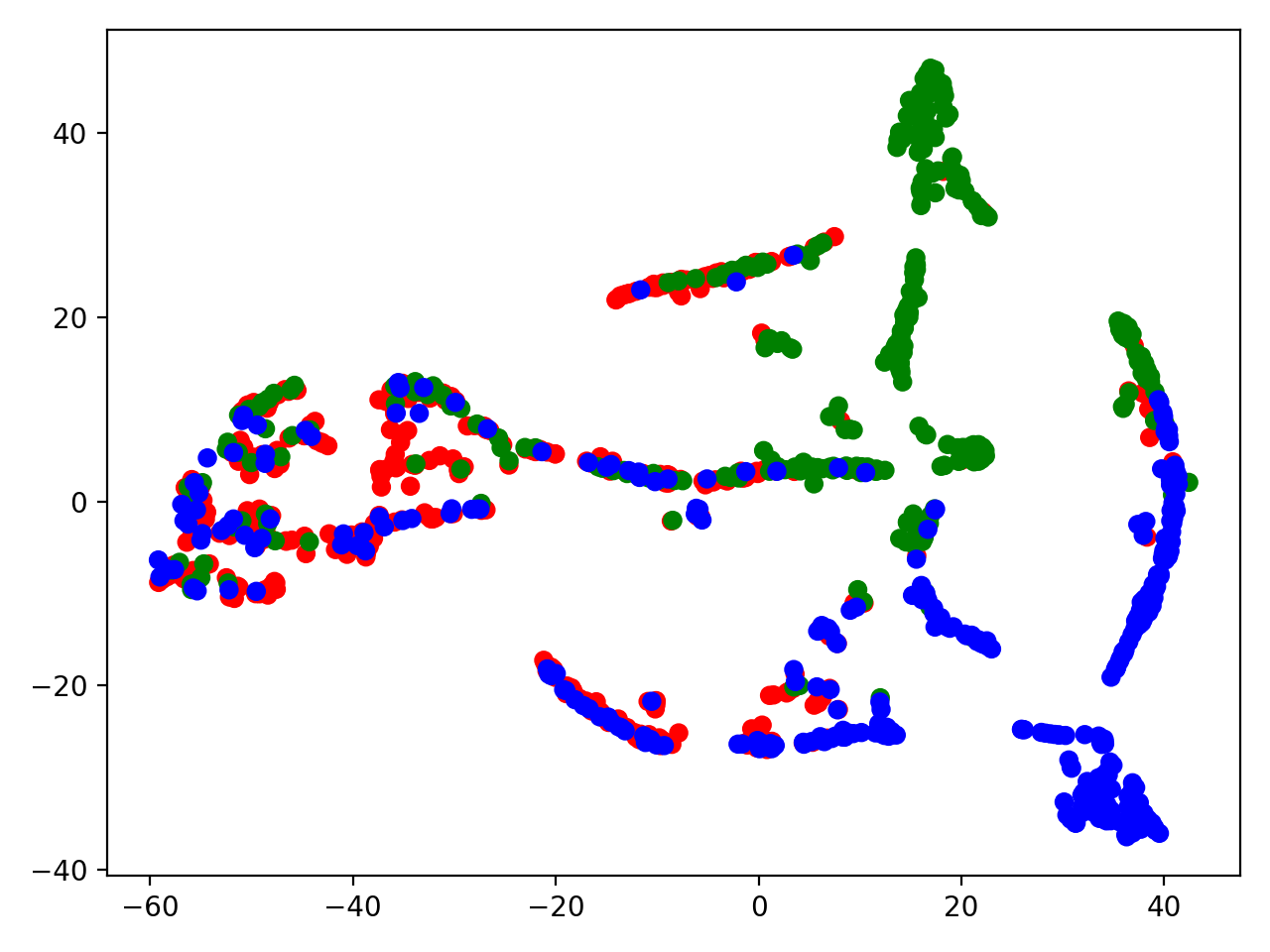}%
\end{tabular}
\end{subfigure}
\caption{Projections of $\hat{\theta}$ learned by SAP-sLDA across three random restarts (see caption of Figure \ref{fig:lda_restarts} for interpretation of figures). We see that SAP-sLDA produces relatively stable projections for both identifiable and non-identifiable $\beta$.}
\label{fig:sap-slda-restarts}
\end{figure}

\subsection{Active Learning for Document Labelling}
We test two types of active-learning for document labelling on one synthetic corpus. 

\begin{figure}[H]
 \centering
 \begin{subfigure}[b]{0.32\linewidth}
     \centering
     \includegraphics[width=\linewidth]{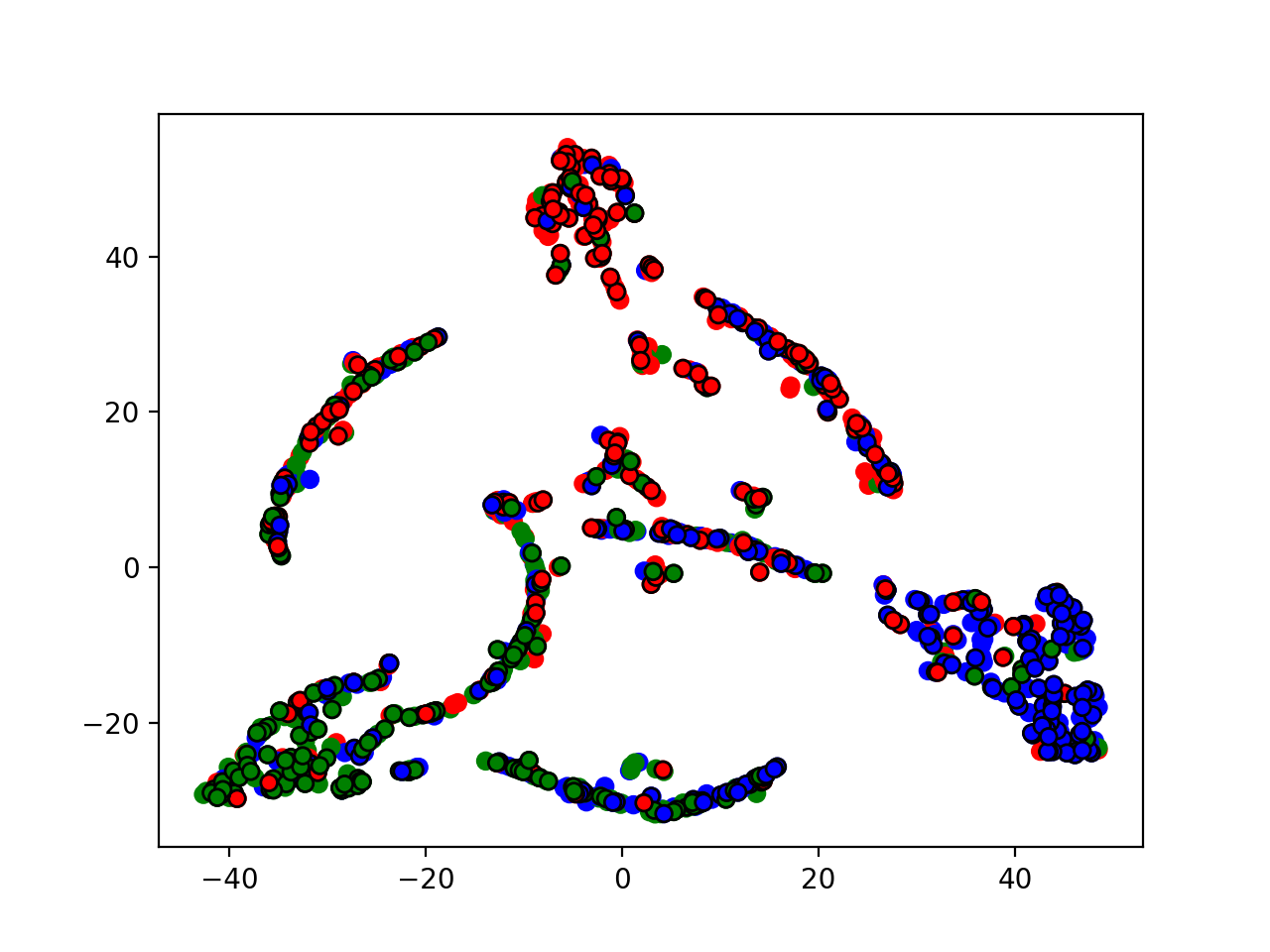}
 \end{subfigure}
 \hfill
 \begin{subfigure}[b]{0.32\linewidth}
     \centering
     \includegraphics[width=\linewidth]{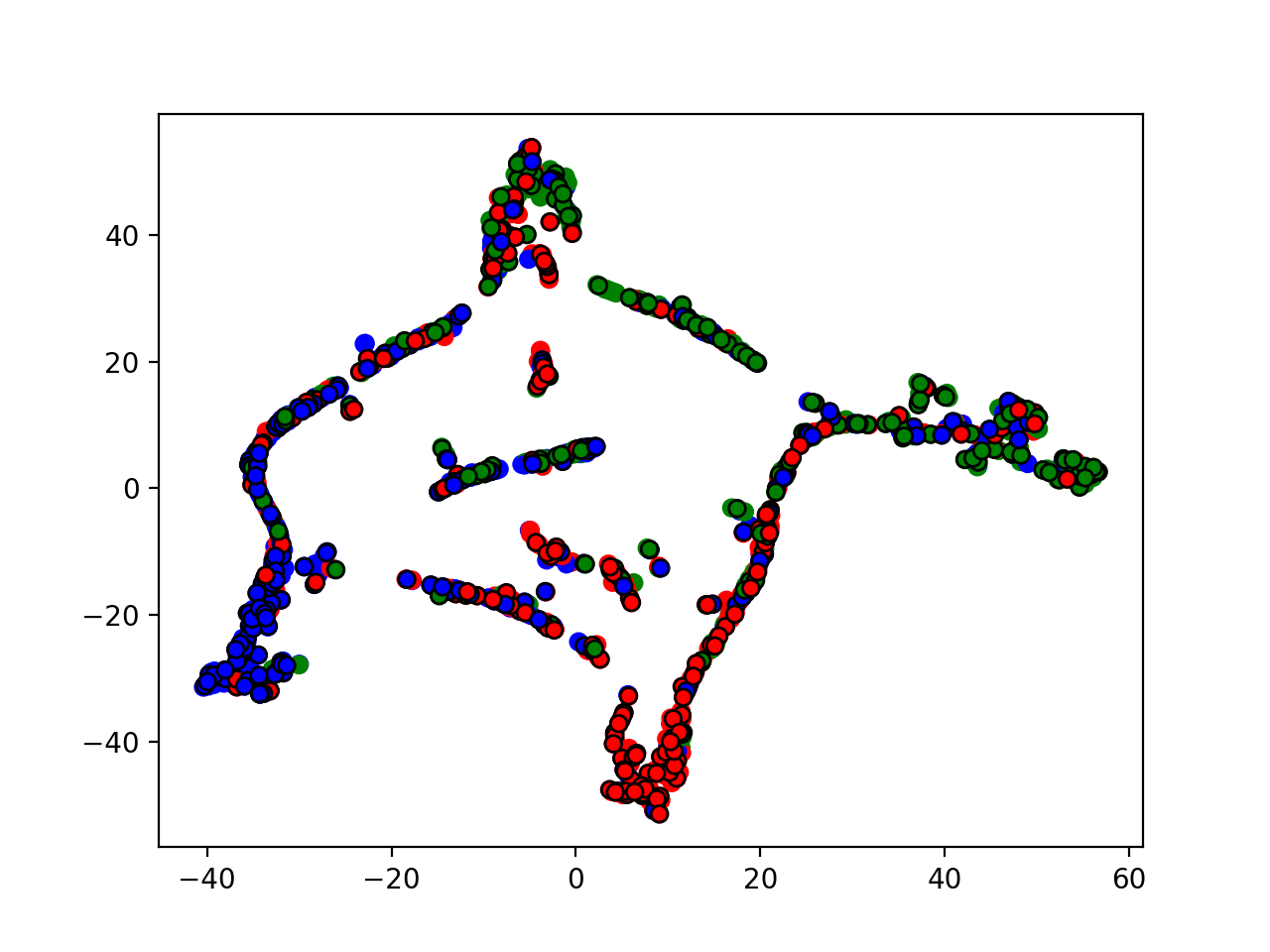}
 \end{subfigure}
\hfill
 \begin{subfigure}[b]{0.32\linewidth}
     \centering
     \includegraphics[width=\linewidth]{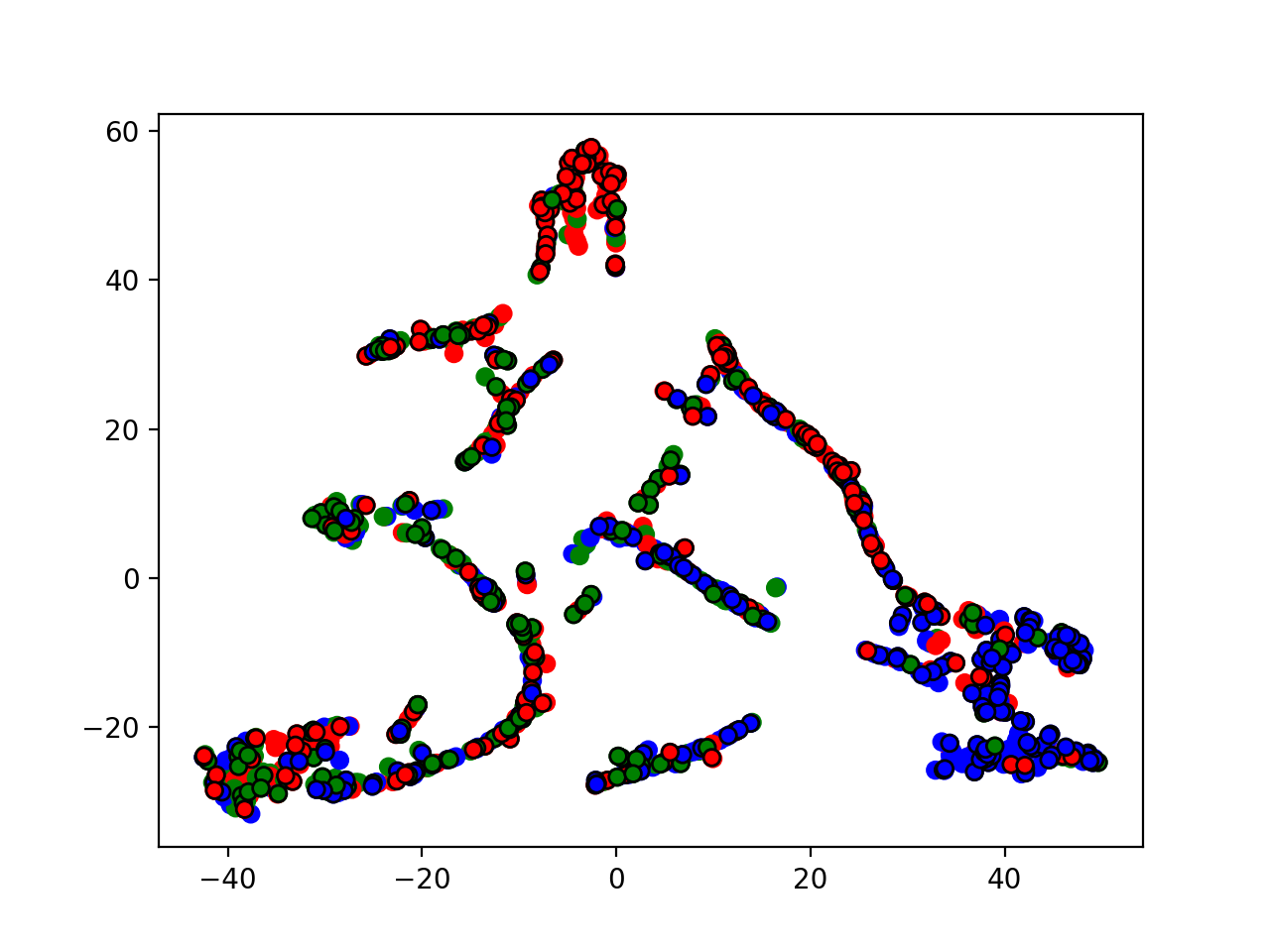}
 \end{subfigure}
    \caption{SAP-sLDA for setting 2 (mixed-topic documents) for nonidentifiable $\beta$ with 50\% of the corpus labelled randomly. We see that the projections are relatively stable across restarts.}
    \label{fig:random-active}
\end{figure}

\begin{figure}[H]
 \centering
 \begin{subfigure}[b]{0.32\linewidth}
     \centering
     \includegraphics[width=\linewidth]{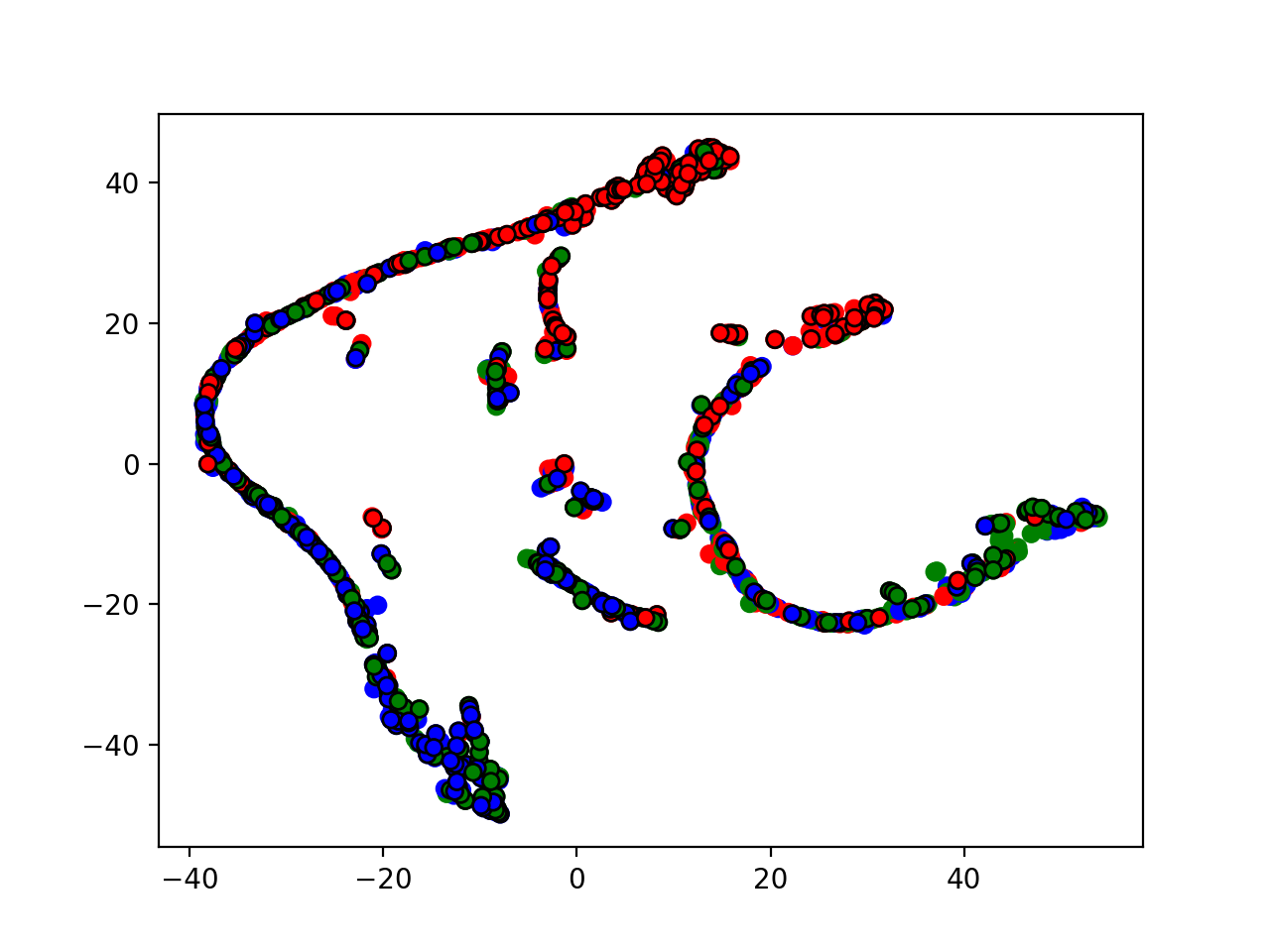}
 \end{subfigure}
 \hfill
 \begin{subfigure}[b]{0.32\linewidth}
     \centering
     \includegraphics[width=\linewidth]{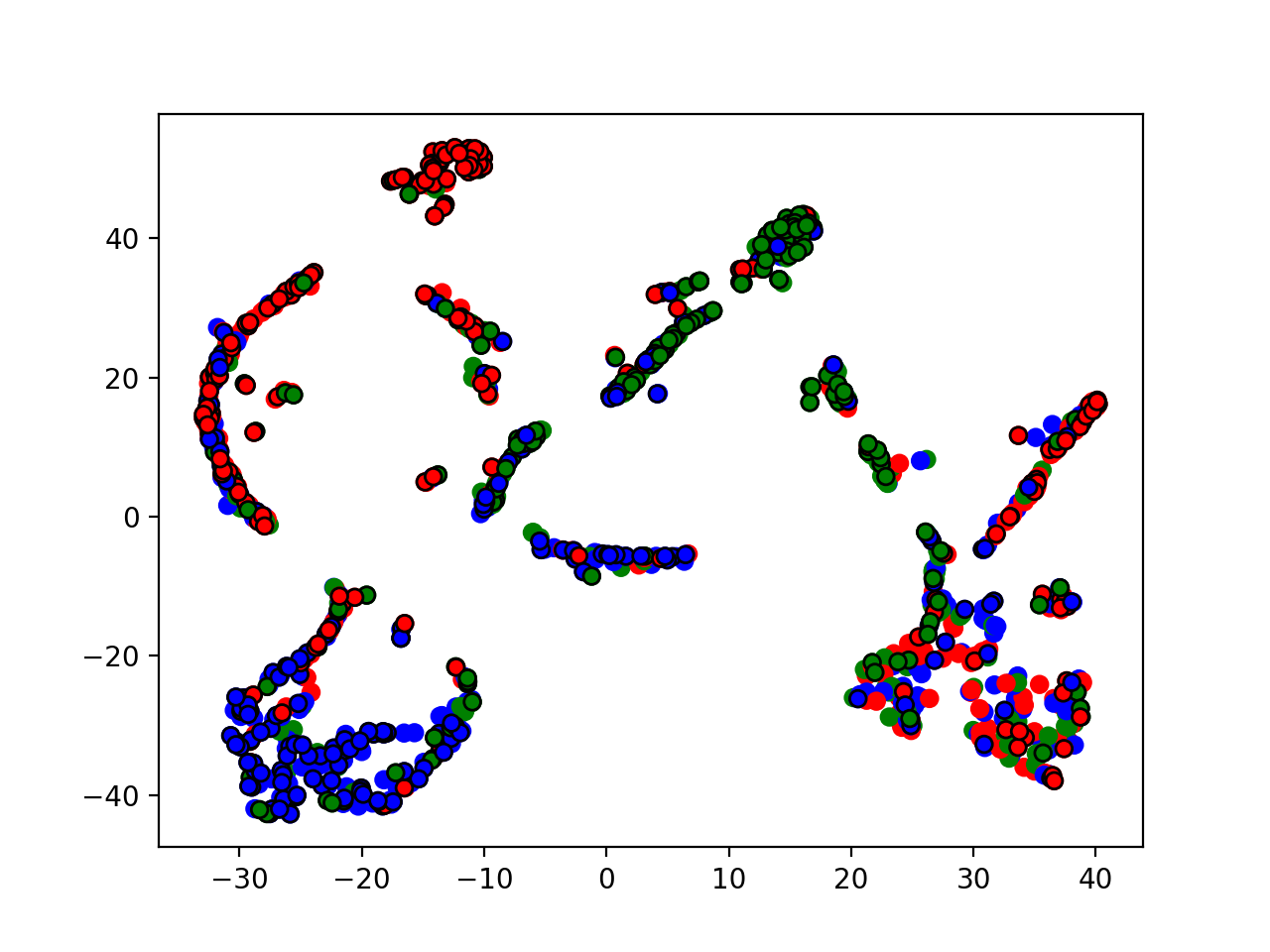}
 \end{subfigure}
\hfill
 \begin{subfigure}[b]{0.32\linewidth}
     \centering
     \includegraphics[width=\linewidth]{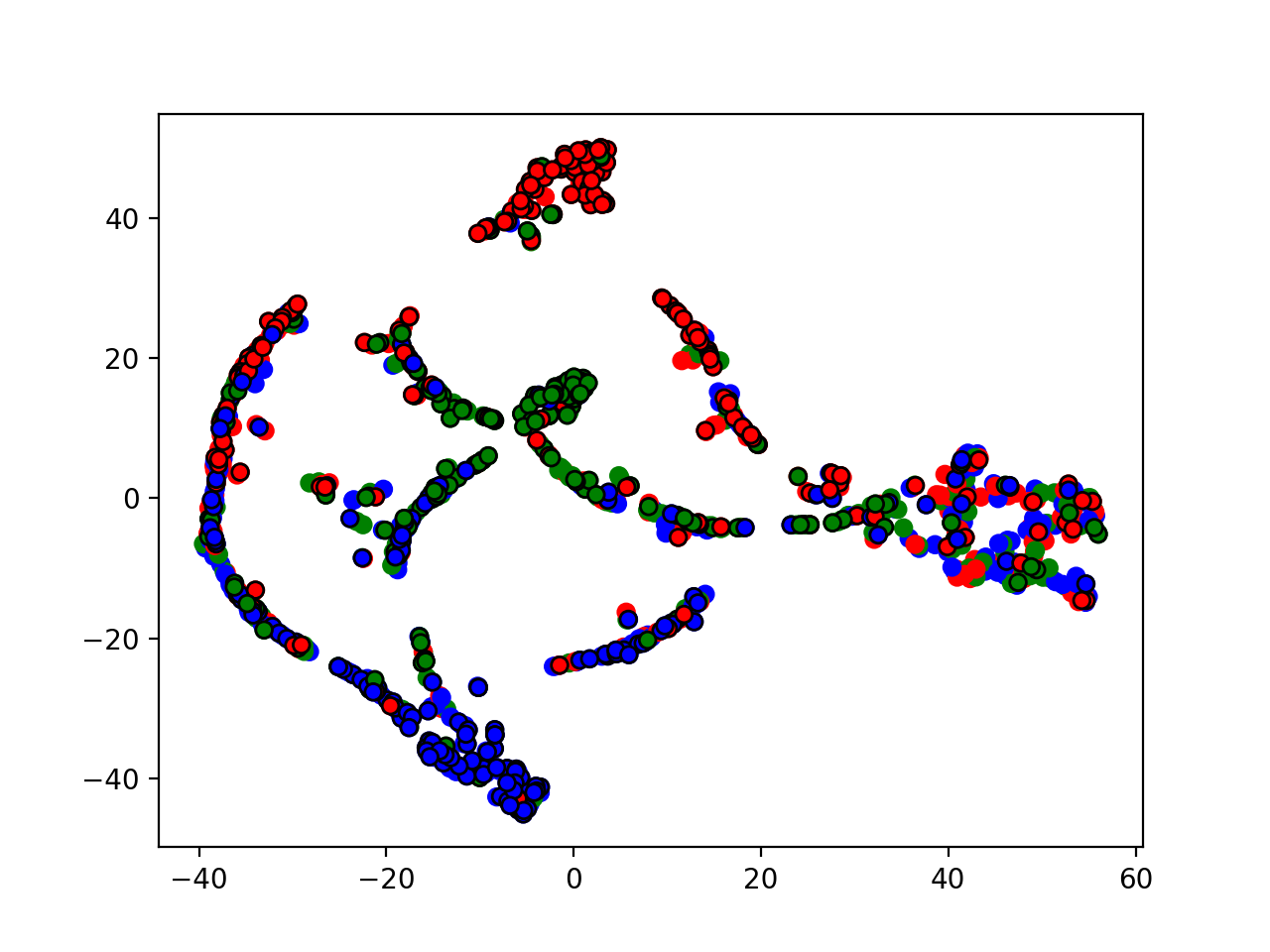}
 \end{subfigure}
    \caption{For the same setup as Figure \ref{fig:random-active} using variance-based active-learning, we see that projections are \textit{not} stable even when 50\% of the corpus is labelled.}
    \label{fig:variance-active}
\end{figure}

\subsection{SAP-sLDA Projections when Labelling Documents Randomly and by Theme on Dharma Seed Corpus}
\begin{figure}[H]
 \centering
\begin{subfigure}[b]{0.49\textwidth}
     \centering
     \includegraphics[width=0.49\linewidth]{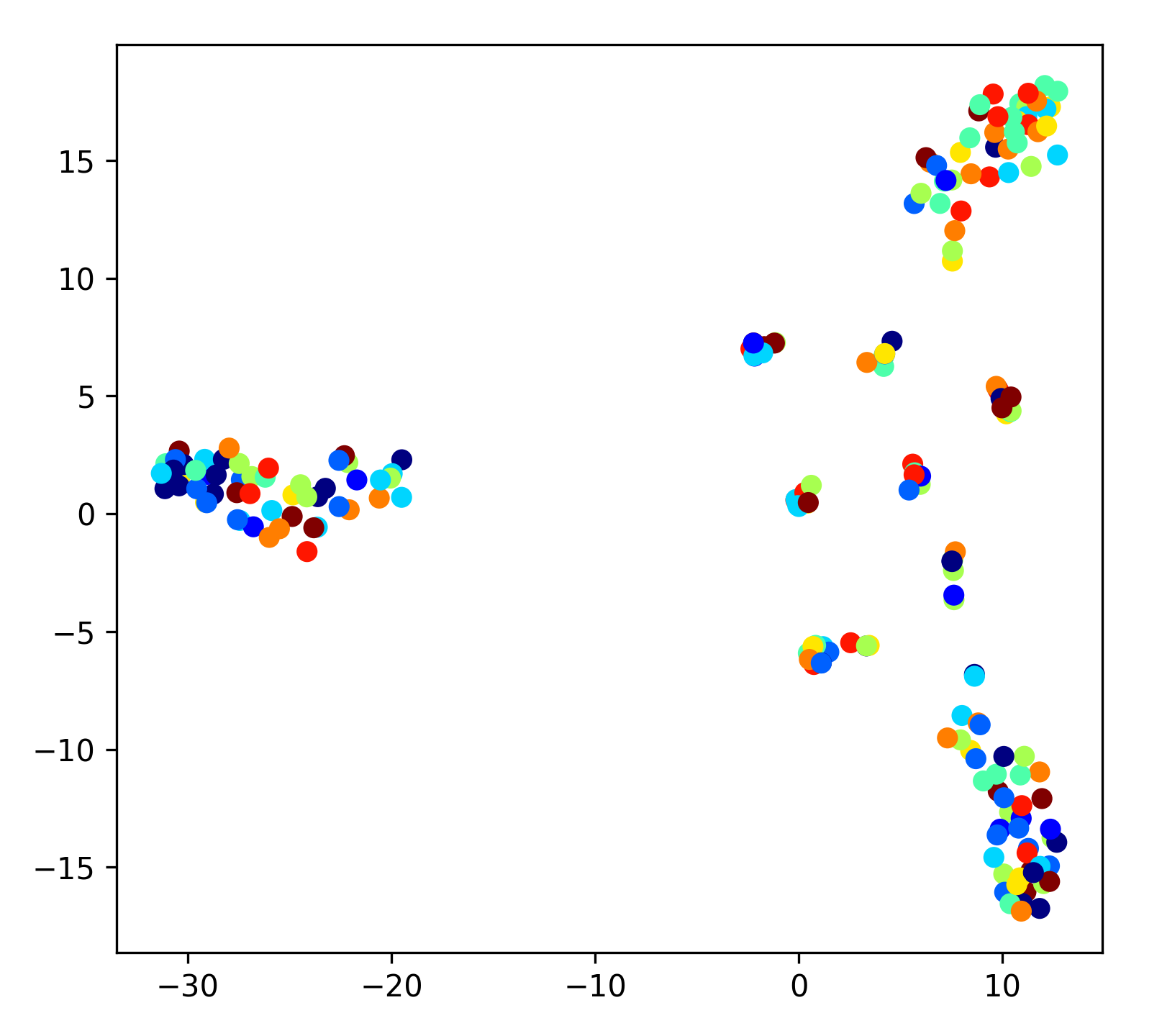}
    \includegraphics[width=0.49\linewidth]{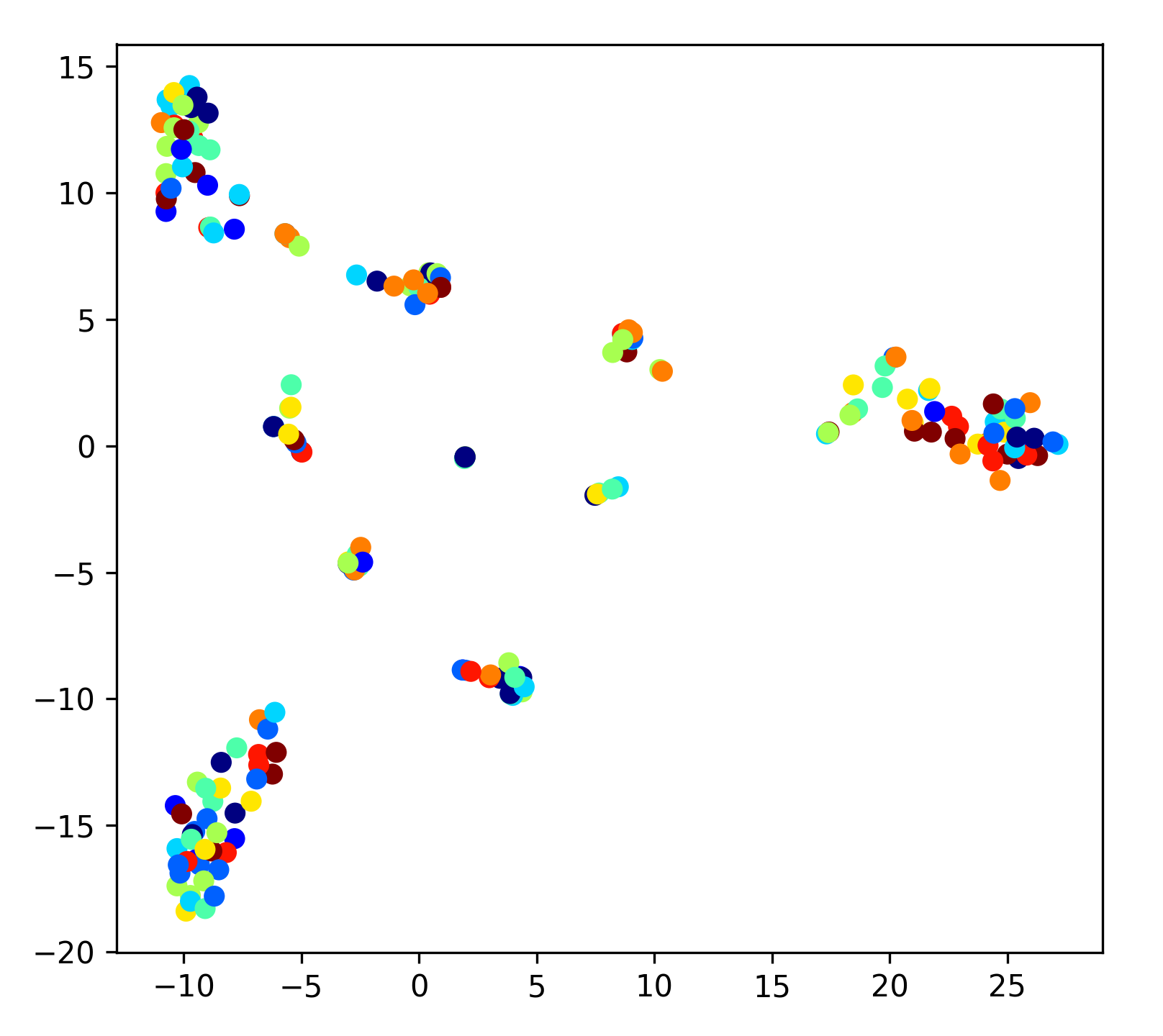}
\end{subfigure}
\begin{subfigure}[b]{0.49\linewidth}
     \centering
     \includegraphics[width=0.49\linewidth]{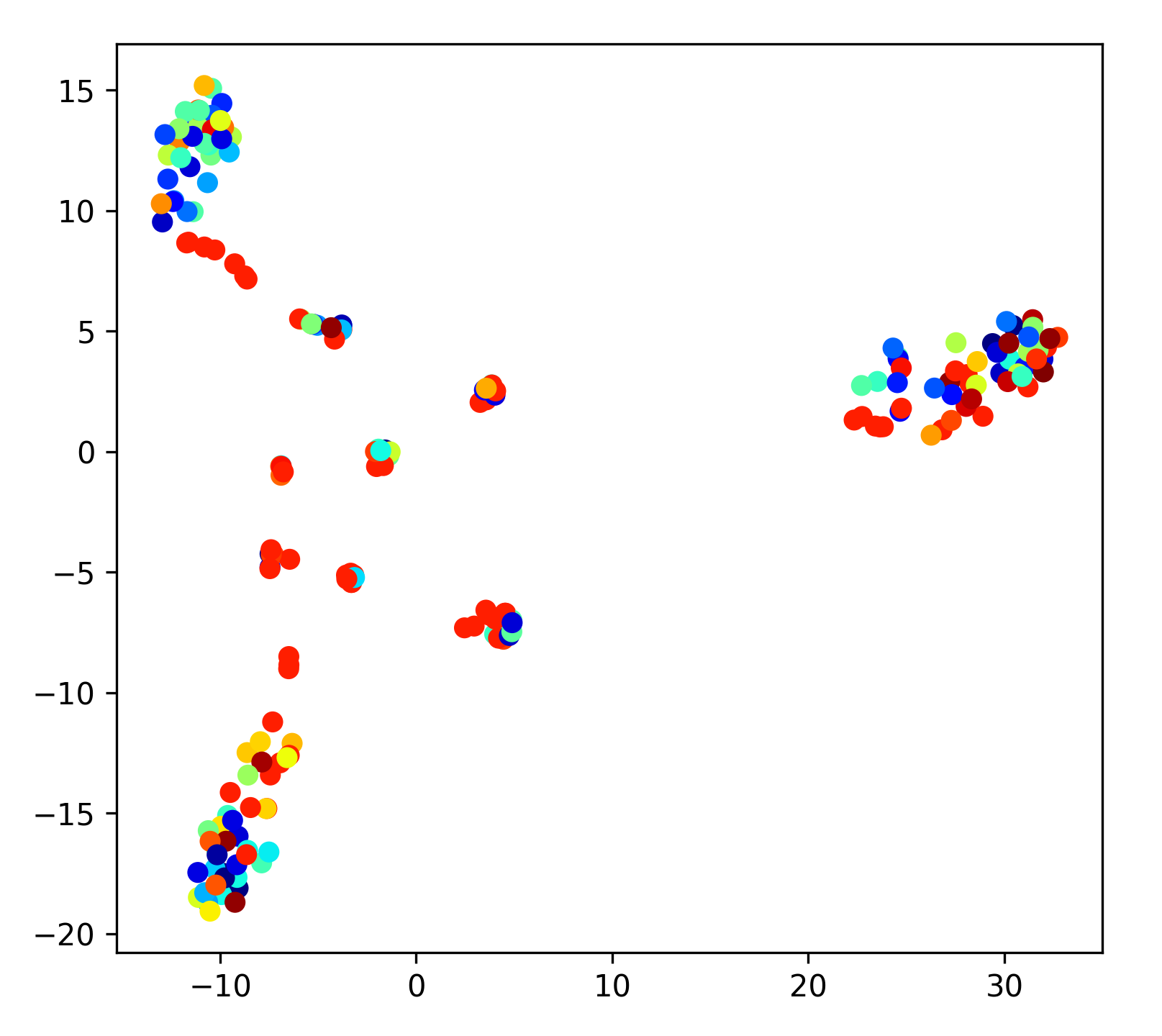}
      \includegraphics[width=0.49\linewidth]{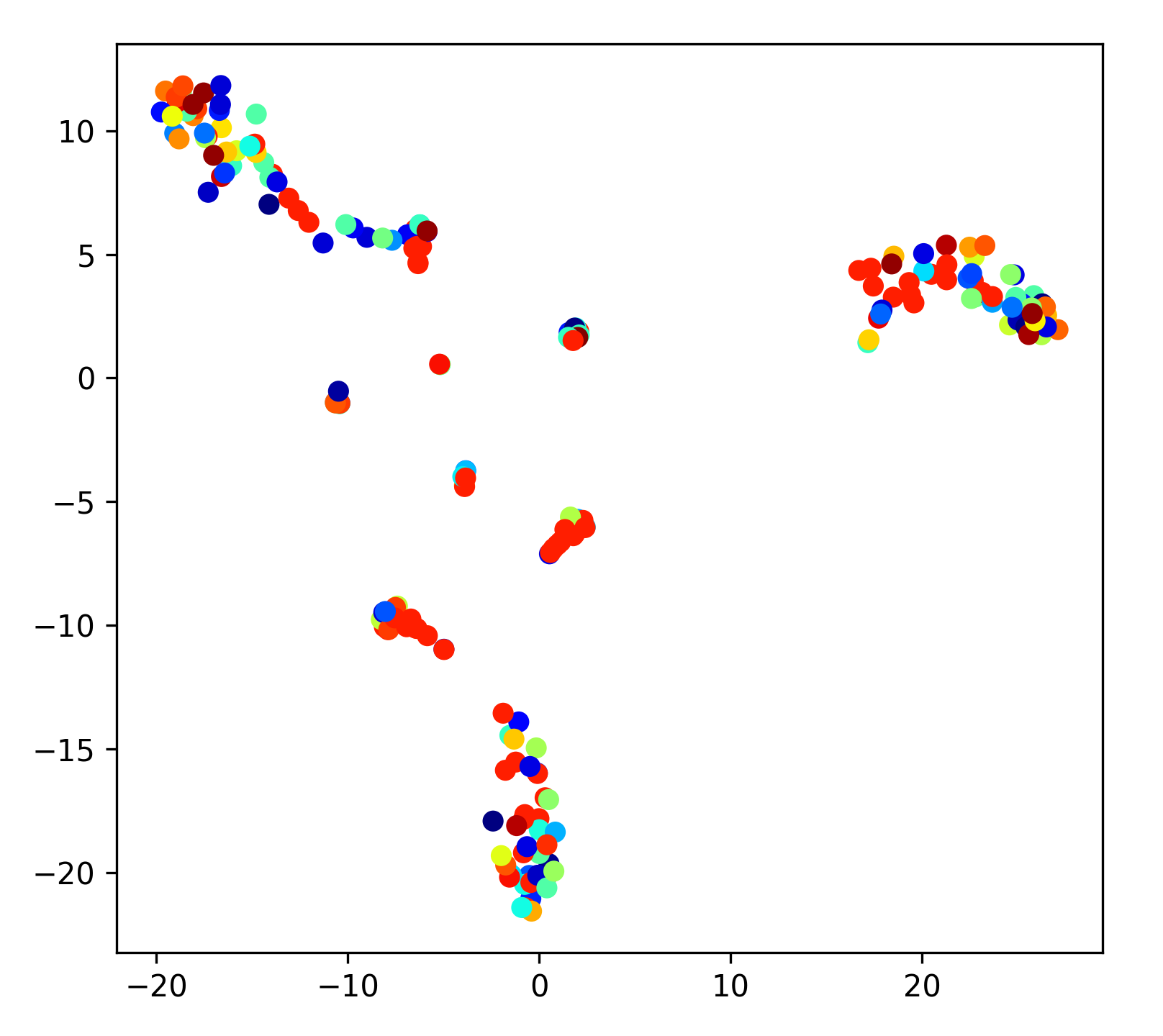}
 \end{subfigure}
\caption{Projections of $\hat{\theta}$ learned by SAP-sLDA when labelling randomly (left two figures) and labelling by author (right two figures). Documents are colored by their label. We see that clusters are mixed-label and are unstable.}
\label{fig:naive-label}
\end{figure}

\end{document}